\documentclass[sigconf]{aamas} 

\newcommand\blfootnote[1]{%
  \begingroup
  \renewcommand\thefootnote{}\footnote{#1}%
  \addtocounter{footnote}{-1}%
  \endgroup
}
\usepackage{balance} 
\usepackage{algorithm}
\usepackage{algorithmic}

\setcopyright{ifaamas}
\acmConference[AAMAS '22]{Proc.\@ of the 21st International Conference
on Autonomous Agents and Multiagent Systems (AAMAS 2022)}{May 9--13, 2022}
{Online}{P.~Faliszewski, V.~Mascardi, C.~Pelachaud,
M.E.~Taylor (eds.)}
\copyrightyear{2022}
\acmYear{2022}
\acmDOI{}
\acmPrice{}
\acmISBN{}

\acmSubmissionID{58}

\title[AAMAS-2022 Formatting Instructions]{REMAX: Relational Representation for Multi-Agent Exploration}

\author{Heechang Ryu}
\affiliation{
  \institution{Samsung Research}
  \city{Seoul}
  \country{Republic of Korea}}
\email{heechang.ryu21@gmail.com}

\author{Hayong Shin}
\affiliation{
  \institution{KAIST}
  \city{Daejeon}
  \country{Republic of Korea}}
\email{hyshin@kaist.ac.kr}

\author{Jinkyoo Park\textsuperscript{$\ast$}}
\affiliation{
  \institution{KAIST}
  \city{Daejeon}
  \country{Republic of Korea}}
\email{jinkyoo.park@kaist.ac.kr}

\begin{abstract}
Training a multi-agent reinforcement learning (MARL) model with a sparse reward is generally difficult because numerous combinations of interactions among agents induce a certain outcome (\emph{i.e.,} success or failure). Earlier studies have tried to resolve this issue by employing an intrinsic reward to induce interactions that are helpful for learning an effective policy. However, this approach requires extensive prior knowledge for designing an intrinsic reward. To train the MARL model effectively without designing the intrinsic reward, we propose a learning-based exploration strategy to generate the initial states of a game. The proposed method adopts a variational graph autoencoder to represent a game state such that (1) the state can be compactly encoded to a latent representation by considering relationships among agents, and (2) the latent representation can be used as an effective input for a coupled surrogate model to predict an exploration score. The proposed method then finds new latent representations that maximize the exploration scores and decodes these representations to generate initial states from which the MARL model starts training in the game and thus experiences novel and rewardable states. We demonstrate that our method improves the training and performance of the MARL model more than the existing exploration methods.
\end{abstract}

\keywords{Multi-Agent Reinforcement Learning; Multi-Agent Exploration}

\newcommand{\BibTeX}{\rm B\kern-.05em{\sc i\kern-.025em b}\kern-.08em\TeX}

\begin{document}


\pagestyle{fancy}
\fancyhead{}


\maketitle 


\section{Introduction}

Along with deep neural networks,\blfootnote{\textsuperscript{$\ast$}Corresponding author} reinforcement learning (RL) has dramatically improved in the past decades, exceeding human-level performance in challenging games \cite{mnih2015human,silver2016mastering,silver2017mastering}. 
Learning with the games or physical simulators helps to simulate and solve real-world problems, such as smart grids \cite{dall2013distributed}, logistics \cite{ying2005multi,cao2013overview}, and distributed vehicles$/$robots  \cite{corke2005networked,fax2004information,matignon2012coordinated}.
The advances in RL have naturally generated a significant interest in MARL as the achievements in RL can be extended to more complex problems involving multiple interacting agents. 
However, the training of a MARL model is difficult even for a simple multi-agent task because it needs to learn implicitly how multiple agents interact with each other along with the environment and how these interactions induce certain task outcomes from only a reward signal.
Furthermore, training the MARL model using typical exploration methods that are effective for RL has been unsuccessful because of a large number of possible interactions that the MARL model needs to explore. 

Exploration methods employed by researchers for RL can be categorized into three categories: exploration bonus, goal conditioning, and initial state generation. The exploration bonus methods provide an intrinsic reward to an agent if the agent visits a state that has not been visited. Count-based exploration \cite{bellemare2016unifying,ostrovski2017count,tang2017exploration}, and curiosity-driven exploration \cite{pathak2017curiosity,burda2018large,burda2018exploration} are some examples of this method. Goal-conditioning methods manipulate a task's goal in the simulator to ensure that the agent keeps receiving a reward to achieve different goals. %
These methods virtually endorse the agent the experience of ``success'' to make training faster using virtual reward signals of success \cite{andrychowicz2017hindsight,florensa2017automatic}. 
Finally, initial state generation methods adjust the initial state distribution in the simulator to generate initial states where the agent can easily complete a task and receive a reward \cite{nair2018overcoming, resnick2018backplay,florensa2017reverse,goyal2018recall}. 
These methods assume that they can arbitrarily reset the agent into any initial state at the beginning of the training episodes in the simulator. 
The common purpose of these three types of exploration methods is to expose the agent to as many new and rewardable experiences as possible. In particular, goal-conditioning and initial state generation methods are mainly used for goal-oriented tasks where the reward signal is typically sparsely given and depends on whether a goal is achieved or not.

On the contrary, MARL exploration methods are limited in comparison to RL. Most MARL exploration methods focus on designing an intrinsic reward to induce certain collective behaviors of agents that are believed to help solve a multi-agent task or game.
For example, intrinsic rewards are provided when one agent's action affects other agents' state transition \cite{wang2019influence,bohmer2019exploration}, and when all agents explore only different (novel) or same (rewardable) state space \cite{iqbal2019coordinated}. However, designing a good intrinsic reward is difficult because it requires the prior knowledge of interaction types that can help solve a multi-agent task or game. In addition, designing an effective intrinsic reward often requires an iterative reward shaping procedure until satisfactory performance is achieved. 


We herein propose a learning-based exploration method called \textbf{RE}lational representation for \textbf{M}ulti-\textbf{A}gent e\textbf{X}ploration (REMAX) to effectively generate novel and rewardable initial states for improving the training of a MARL model without designing an intrinsic reward. As an initial state generation method, REMAX has additional networks that interact with the MARL model to generate initial states in the simulator. These networks consist of two parts: state representation and generation. For state representation, REMAX employs a variational graph autoencoder (VGAE) \cite{kipf2016variational} to extract latent vectors from the game states 
such that they can be used as an effective input for a  coupled surrogate model to predict exploration scores. The exploration score quantifies the balance between exploitation and exploration to generate states that are useful for MARL training.
In particular, REMAX effectively encodes the states by representing the relationships among agents using an encoder constructed using a graph attention network (GAT) \cite{velivckovic2017graph} in VGAE. For state generation, REMAX optimizes the surrogate model with respect to a latent vector to find new latent vectors with high exploration scores and decodes them through a VGAE decoder to generate new states. The generated states are then used as the initial states of the training episodes in the simulator to train the MARL model.
To summarize, REMAX learns how to optimally generate initial states that are helpful in boosting the MARL model training through the coupled VGAE and surrogate model.

\section{Related Work}

A representative task with sparse rewards is a goal-oriented task where a binary reward is given only when agents achieve goals (\emph{i.e.,} complete a task). Goal-conditioning and initial state generation methods are mainly used as RL exploration methods for goal-oriented tasks. As one of the goal-conditioning methods, HER \cite{andrychowicz2017hindsight} virtually exposes an agent to a task's goal by setting the final state of a training episode as a virtual goal in a simulator. Similarly, Goal GAN \cite{florensa2017automatic} generates goals with an increasing difficulty using a generator network. Meanwhile, RCG \cite{florensa2017reverse}, one of the initial state generation methods, adjusts the initial state of a training episode in the simulator to ensure that the probability of an agent reaching a task's goal is within a specific bound. While adjusting the initial state, RCG needs to know at least one state that can reach the goal unconditionally, such as an exact goal state. Some methods perform imitation learning on the predicted state trajectory to obtain a reward \cite{goyal2018recall}, or determine the initial state using expert demonstrations on the task \citep{nair2018overcoming, resnick2018backplay}.

MARL exploration methods for a multi-agent task with sparse rewards, such as a multi-agent goal-oriented task, have not been extensively studied. As one of the possible exploration strategies, GENE \cite{jiang2019generative} has been proposed to generate initial states in the simulator for boosting the exploration of MARL models. This method represents the state of a game as a latent vector using a variational autoencoder (VAE) \cite{kingma2013auto} and estimates the density of the latent vectors using a kernel density estimation (KDE) \cite{davis2011remarks}. By sampling latent vectors from KDE and decoding them, GENE generates new initial states that are believed to be novel and rewardable. However, because GENE trains the state representation module (VAE) and the state density estimation module (KDE) separately, the latent representations may not be the best for generating novel and rewardable initial states. Another critical limitation of the method is that it lacks of considering the relationships among agents. A simple example illustrating this issue is when the relative positions of $n$ homogeneous agents can be represented as $n!$ different states unless a permutation invariance is imposed on the state representation. When focusing on the relative configuration among agents, $n!$ different states can be considered as the same state. It can reduce the state space to be explored and increase the efficiency of sampling states during exploration.

We consider GENE as the main comparison method because GENE and REMAX are both initial state generation methods that can be used for multi-agent goal-oriented tasks without shaping a reward. We also consider additional baseline methods for goal-oriented tasks, such as HER and RCG. However, for a non-goal-oriented task, where there is no goal to be reached, we exclude HER and RCG. 
Instead, we consider EDTI \cite{wang2019influence}, which uses an intrinsic reward to quantify the influence of an agent's action on the expected returns of other agents.
Note that we exclude the MARL exploration methods using the intrinsic reward because they require a user-designed or a domain-specific reward signal. 
Our study assumes that we can arbitrarily reset agents into any initial state at the beginning of training episodes in the simulator, which can be easily satisfied for the current study that is focusing on offline training using simulation environments.



\section{Background}

\subsection{Multi-Agent Reinforcement Learning}
We consider a partially observable Markov game \cite{littman1994markov}, which is an extension of the partially observable Markov decision process for a game with multiple agents. A partially observable Markov game for $\mathnormal{N}$ agents is defined as follows: $s\in\mathcal{S}$ denotes the global state of the game; ${o_i}\in\mathcal{S}\mapsto\mathcal{O}_{i}$ denotes a local observation correlated with the state that agent $i$ can acquire; and $a_i\in\mathcal{A}_{i}$ is an action of agent $i$. The reward for agent $i$ is obtained as a function of state $s$ and joint action $\mathbf{a}$ as ${r}_{i}:\mathcal{S}\times\mathcal{A}_1\times\dots\times\mathcal{A}_N\mapsto\mathbb{R}$. The state evolves to the next state according to the state transition function $\mathcal{T}:\mathcal{S}\times\mathcal{A}_1\times\dots\times\mathcal{A}_N\mapsto\mathcal{S}$.
Agent $i$ aims to maximize its discounted return, $R_i=\sum_{t=0}^{T} \gamma^t r_{i,t}$, where $\gamma\in[0,1]$ is a discount factor.
In MARL, each agent learns an individual policy that maps the observation to its action to maximize its expected return, which is approximated by a $Q$-function. While the policy can be deterministic $(a_i=\mu_i(o_i))$ or stochastic $(a_i\sim\pi_i(\cdot\arrowvert o_i))$, the multi-agent deep deterministic policy gradient (MADDPG) \cite{lowe2017multi} adopts a deterministic policy. MADDPG comprises individual $Q$-networks and policy networks for each agent. 
MADDPG lets the $Q$-network of agent $i$ be trained by minimizing the loss (TD error) as follows: $\mathcal{L}(\varphi_i)={\mathbb{E}_{\mathbf{o},\mathbf{a},r,{\mathbf{o}'}\sim \mathcal{D}}}[(Q_i^{\mu}(\mathbf{o},\mathbf{a};\varphi_i)-y_i)^2], y_i=r_i+\gamma {Q_i^{\mu'}}({\mathbf{o}'},\mathbf{a}';{\varphi'_i})\arrowvert_{a_j'=\mu_j'(o'_j;\vartheta_j')},$ where $\mathbf{o}=(o_1,\dots,o_N)$ and $\mathbf{a}=(a_1,\dots,a_N)$ represent the joint observation and joint action of all agents, respectively. $\mathcal{D}$ is an experience replay buffer that stores $(\mathbf{o},\mathbf{a},r,{\mathbf{o}'})$ samples obtained from the training episodes. $Q^{\mu'}$ and $\mu'$ are target networks for the stable learning of $Q$ and policy networks, respectively. The policy network, $\mu_i(o_i;\vartheta_i)$, of agent $i$ is optimized using the gradient computed as $\mathbb{E}_{\mathbf{o},\mathbf{a}\sim \mathcal{D}}[\nabla_{\vartheta_i}Q_i^{\mu}(\mathbf{o},\mathbf{a};\varphi_i )]
=\mathbb{E}_{\mathbf{o},\mathbf{a}\sim \mathcal{D}}[\nabla_{\vartheta_i}\mu_i(o_i;\vartheta_i)\nabla_{a_i}Q_i^{\mu}(\mathbf{o},\mathbf{a};\varphi_i )\arrowvert_{a_i=\mu_i(o_i;\vartheta_i)}]$.

\subsection{Variational Graph Autoencoder}
VGAE is an unsupervised representation learning model for graph-structured data. 
It consists of a graph convolutional network (GCN) \cite{kipf2016semi} encoder and a simple inner product decoder. The GCN encoder performs a posterior inference for latent representations using a node feature matrix and an adjacency matrix of graph-structured data. The decoder performs an inner product between the encoded latent variables to reconstruct the adjacency matrix.

\subsection{Graph Attention Network}
GAT is an effective model for processing graph-structured data. It proposes to compute new node embeddings $h'_{i}$ for target node $i$ in a graph by aggregating previous node embeddings $h_{j}$ from neighboring nodes $\{j\in \mathcal{N}_i\}$ that are connected to target node $i$ as $h'_{i}=\sigma(\sum_{j\in \mathcal{N}_i} \alpha_{ij} \mathbf{W} h_j)$. The attention coefficient $\alpha_{ij}=\text{softmax}_{j}(e_{ij})$, where $e_{ij}=a(\mathbf{W}h_i,\mathbf{W}h_{j})$, quantifies the importance of node $j$ to node $i$ in computing $h'_i$. The attention mechanism effectively differentiates the importance of different nodes in updating the embeddings of the target node. In addition, the attention mechanism can be extended to a multi-head attention \cite{vaswani2017attention}.

\begin{figure}[t]
    \centering
    \includegraphics[scale = 0.31]{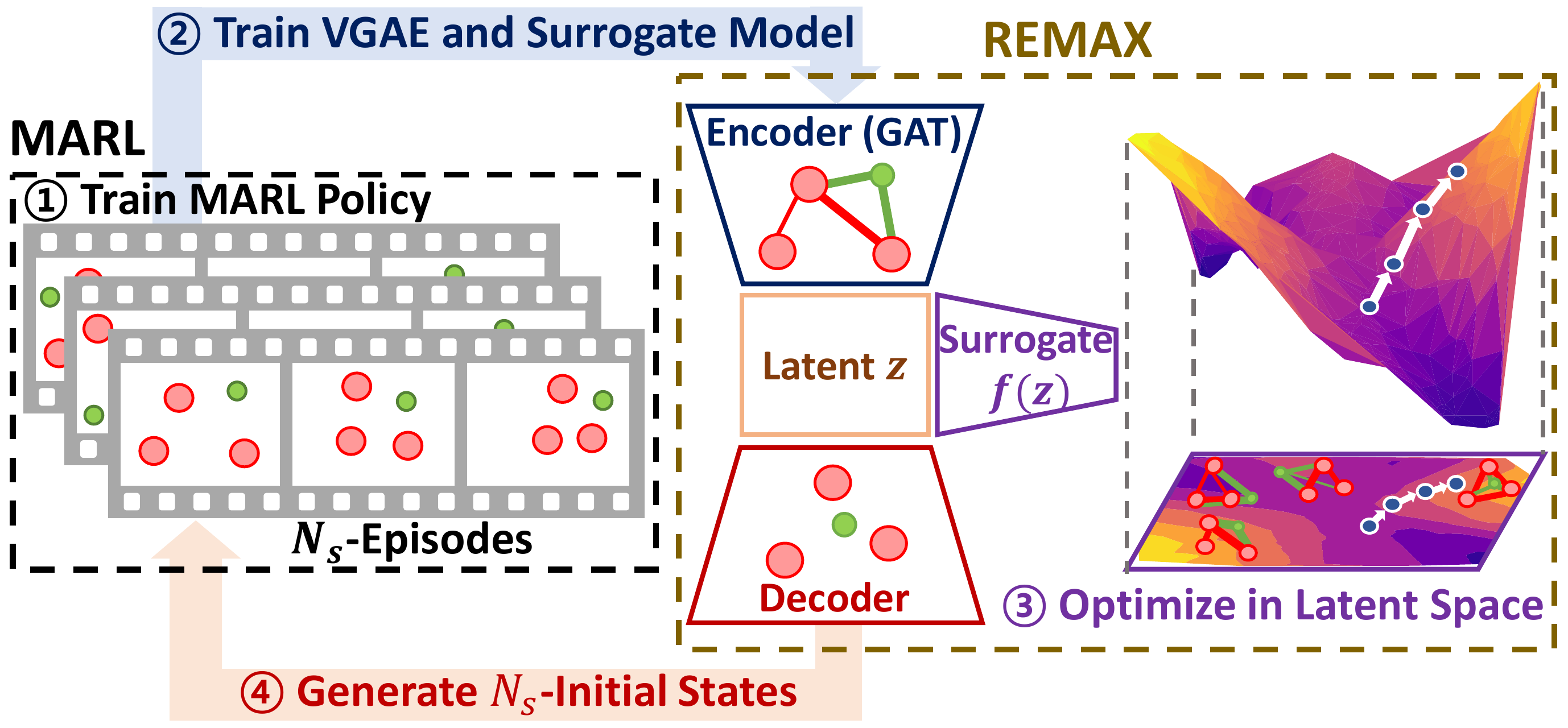}
    \caption{Overview of REMAX.}
    \label{fig:overview}
\end{figure}

\section{Methods}
Figure \ref{fig:overview} shows how REMAX runs with a coupled MARL model. First, MARL starts training by playing a game in a simulator. When a certain number of states and associated rewards are collected from the game during MARL training, REMAX trains (1) a VGAE to represent the states as latent vectors by considering the relationships among agents and (2) a surrogate model to map the latent vectors into the exploration scores. The VGAE and surrogate model undergo an end-to-end training. Once the training is finished, REMAX gathers a set of optimized latent vectors by maximizing the surrogate model with respect to a latent vector. Finally, REMAX decodes these optimized latent vectors using the trained VGAE decoder to generate new states. MARL then starts the next training episodes from the new initial states generated by REMAX in the simulator and collects states again during training. The newly collected states by MARL are used to retrain REMAX. In other words, the policy training phase of MARL and the surrogate model learning are alternated during training. 

\subsection{State Representation Using VGAE and Surrogate Model}

\subsubsection{Encoder of VGAE}
We propose GAT-based VGAE 
to transform states and their latent vectors in both directions (forward and backward) while considering the relationships among agents. Unlike the original VGAE using GCN, we use GAT in the encoder part of VGAE to effectively consider the relationships among agents with the learnable and adjustable relational attentions \cite{lowe2017multi,smac}.

First, the GAT encoder reshapes state $s\in\mathbb{R}^{NF}$ as a set of node features $\mathbf{h}=\{h_1,...,h_N\}$, where $N$ is the number of agents (nodes), and $h_i\in\mathbb{R}^F$ is the local state or observation (node feature) of agent $i$. The importance of node $j$ to node $i$ is then computed as the attention coefficient $\alpha_{ij}=\text{softmax}_j(e_{ij})=\frac{\text{exp}(e_{ij})}{\sum_{m=1}^{N}\text{exp}(e_{im})}$, where $e_{ij}=\text{LeakyReLU}(\mathbf{v}^T[\mathbf{W}h_i\mathbin\Vert\mathbf{W}h_j])$. The trainable parameters $\mathbf{W}\in\mathbb{R}^{F'\times F}$ and $\mathbf{v}\in\mathbb{R}^{2F'}$ are shared for all nodes. The computed attention coefficients for every node serve as ``soft'' edges in a graph representing a state, which means the coefficients softly quantify the level of interactions among agents to be between 0 and 1 in the state.

Employing a multi-head attention with $K$ heads, we compute the updated node embedding $h'_i\in\mathbb{R}^{KF'}$ as $h'_{i}=\Vert_{k=1}^{K}\text{ReLU}(\sum_{j=1}^{N} \alpha^k_{ij} \mathbf{W}^k h_j)$, where $\alpha^k_{ij}$ and $\mathbf{W}^k$ are the attention coefficients and parameters corresponding to the $k$-th head attention. In multi-agent settings, using a multi-head attention is beneficial for extracting the distinctly different interactions as there are various types of interactions among agents, especially for non-homogeneous agents.

Once $\mathbf{h'}=\{h'_1,...,h'_N\}$ is computed, it is concatenated as $\|_{i=1}^N h'_i$. The concatenated node embedding is then used to output the mean $\mu$ and standard deviation $\sigma$ for constructing the distribution for a latent vector $z$ as $z\sim q_\phi(z\vert s)=\mathcal{N}(\mu,\sigma^2\mathbf{I})$ with parameters $\phi$. 

\subsubsection{Decoder of VGAE}
A multi-layer perceptron (MLP) decoder $g_\theta$ is adopted to reconstruct the states $\tilde{s}$ from the latent vector $z$, \emph{i.e.,} $\tilde{s}\sim p_\theta(s\vert z)$ or simply $\tilde{s}=g_\theta(z)$ with parameters $\theta$.



\subsubsection{Surrogate Model}
While VGAE is trained to represent states as latent vectors, an MLP surrogate model $f_\psi(z)$, where $z$ is a latent vector in VGAE, is simultaneously learned to map $z$ to an exploration score $y_s=f_\psi(z)$. The exploration score can be flexibly determined depending on the MARL model. For example, if the MARL model is MADDPG, then the $Q$ and policy networks can be used to define the exploration scores $y_s$ as  
\begin{equation} \label{eq:1}
\frac{1}{N}\sum_{j=1}^{N} \Big[{Q_j(\mathbf{s},\mathbf{a})}+\lambda\lvert {Q_j(\mathbf{s},\mathbf{a})-(r_j+\gamma Q_j'(\mathbf{s}',\mathbf{a}'))}\rvert\Big],
\end{equation}
where $\mathbf{s}$ and $\mathbf{a}$ are, respectively, the joint state and the joint action. In MADDPG, observations and rewards corresponding to the states are obtained from the experience replay buffer. In Equation \ref{eq:1}, the first term $\frac{1}{N}\sum_{j=1}^{N} Q_j(\mathbf{s},\mathbf{a})$, which is an empirical mean of $Q$, quantifies how valuable the corresponding $(\mathbf{s}',\mathbf{a}')$ is for all agents.
This term corresponds to the objective of policy networks to maximize their $Q$-values.
Meanwhile, the second term $\frac{1}{N}\sum_{j=1}^{N}\lvert {Q_j(\mathbf{s},\mathbf{a})-(r_j+\gamma Q_j'(\mathbf{s}',\mathbf{a}'))}\rvert$, which is an empirical mean of TD error, quantifies how novel the corresponding $(\mathbf{s}',\mathbf{a}')$ is for all agents. This term corresponds to the objective of $Q$-networks to minimize their TD errors as losses. In other words, the first and second terms respectively evaluate how rewardable (exploitable) and novel (explorable) the state is. It is a combination of exploration using the mean of $Q$ \cite{chen2017ucb} and TD error \cite{schaul2015prioritized}. The hyper-parameter $\lambda\geq0$ adjusts the balance between exploitation and exploration. 



\setcounter{figure}{-3}
\begin{figure*}[t]
\begin{minipage}[c]{.24\textwidth}
  \centering
  \includegraphics[width=0.6\textwidth]{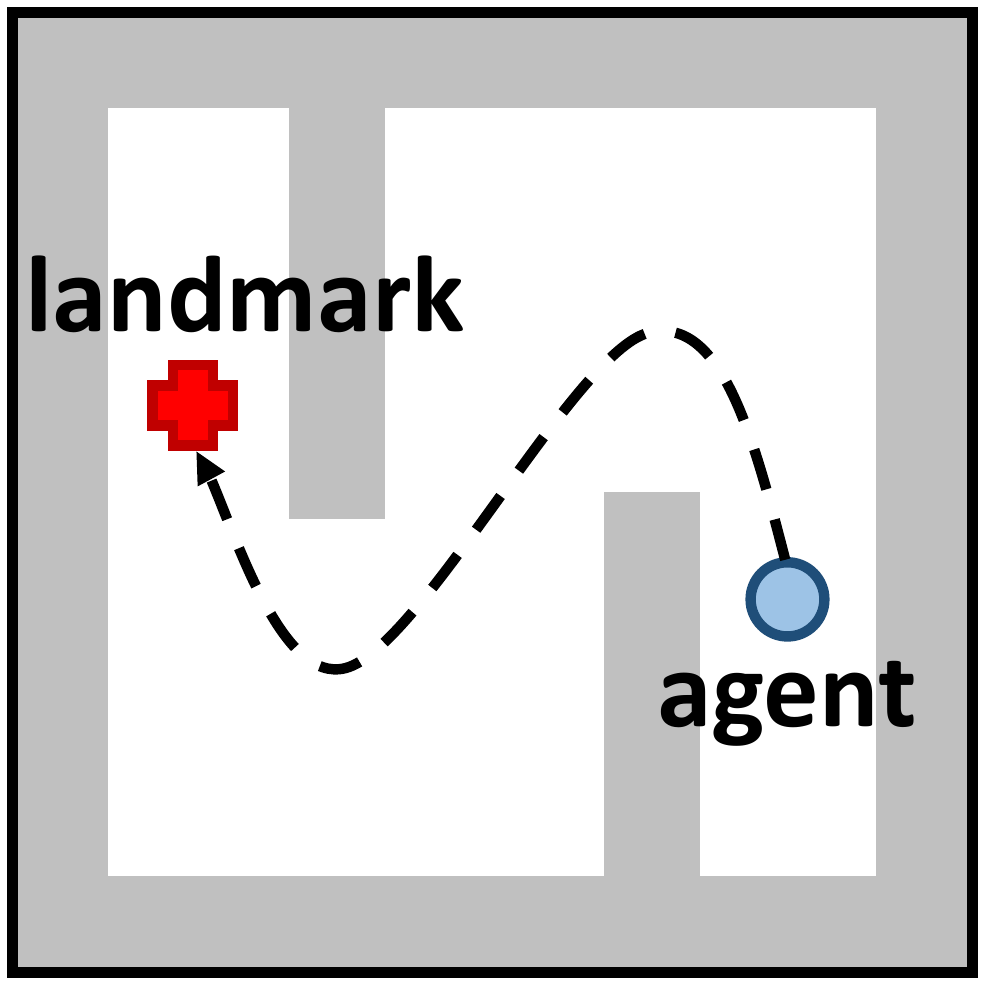}
  \captionsetup{labelformat=empty}
  \caption{a) Maze}
\end{minipage}
\begin{minipage}[c]{.24\textwidth}
  \centering
  \includegraphics[width=0.6\textwidth]{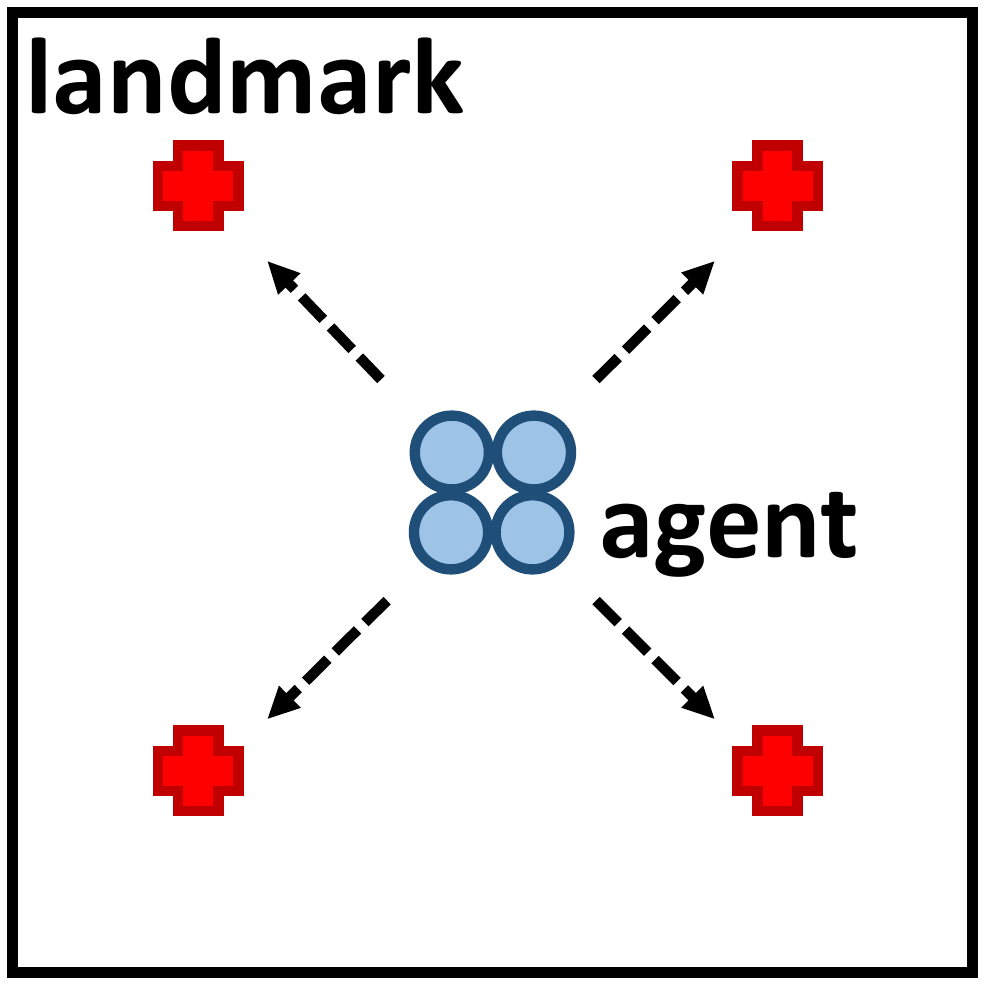}
  \captionsetup{labelformat=empty}
  \caption{b) Cooperative Navigation}
\end{minipage}
\begin{minipage}[c]{.24\textwidth}
  \centering
  \includegraphics[width=0.6\textwidth]{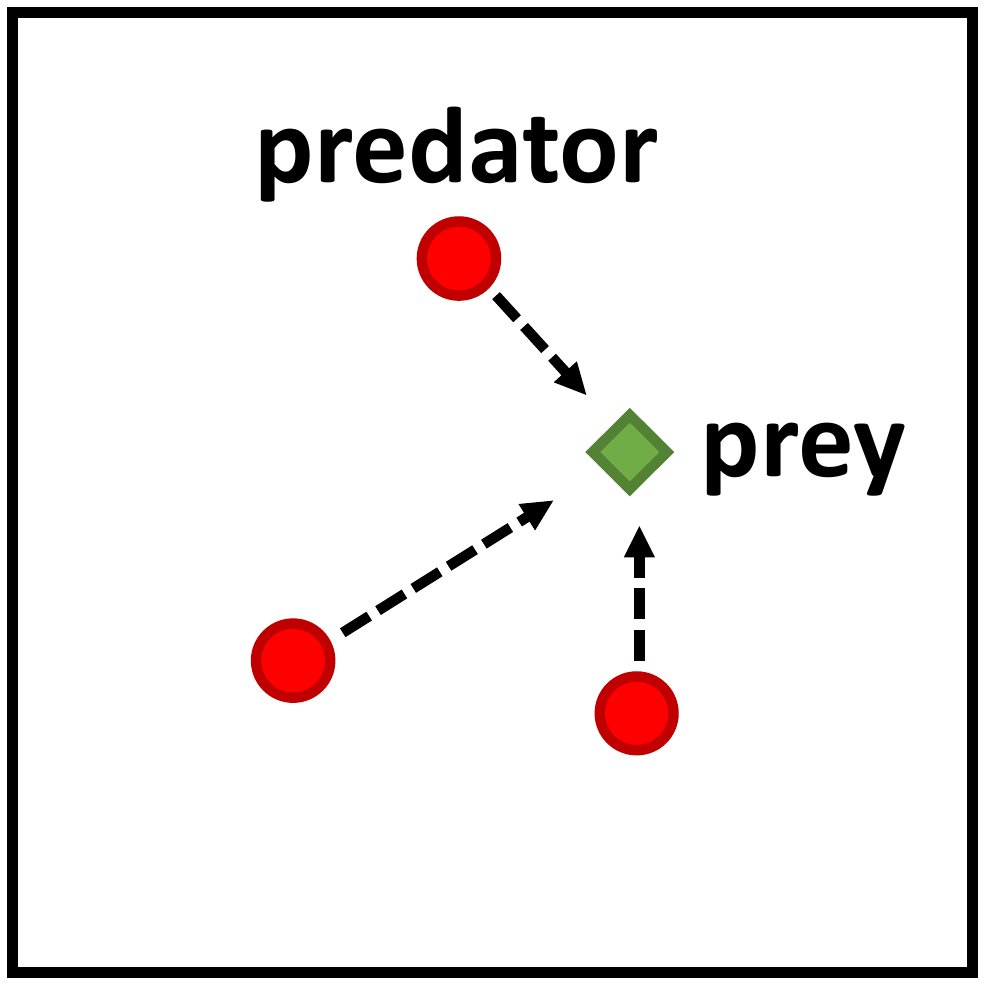}
  \captionsetup{labelformat=empty}
  \caption{c) Predator-Prey}
\end{minipage}
\begin{minipage}[c]{.24\textwidth}
  \centering
  \includegraphics[width=0.585\textwidth]{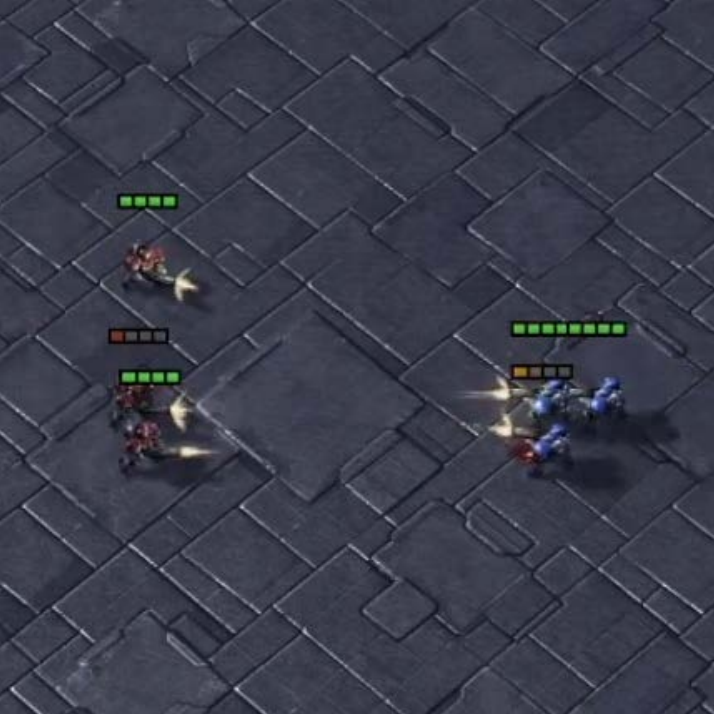}
  \captionsetup{labelformat=empty}
  \caption{d) SMAC - 3 marines}
\end{minipage}

\caption{Illustrations of the experimental environments.}
\label{fig:fig_illustrations}
\end{figure*}

The parameters $(\theta,\phi,\psi)$ of VGAE and the surrogate model are optimized by minimizing the total loss:
\begin{equation}
\mathcal{L}^{\text{total}}=\mathcal{L}^{\text{VGAE}}+\beta\mathcal{L}^{\text{surrogate}},
\end{equation}
where $\mathcal{L}^{\text{VGAE}}$ is the negative variational lower bound expressed as $-\mathbb{E}_{z\sim q_\phi(z\vert s)}[\text{log} \,p_\theta(s\vert z)]+\text{KL}(q_\phi(z\vert s)\|p(z))$, where KL is the KL divergence, and $p(z)$ is a prior expressed as $\mathcal{N}(0,\mathbf{I})$. $\mathcal{L}^{\text{surrogate}}$ is the mean squared error loss expressed as $\mathbb{E}_{z\sim q_\phi(z\vert s)}[(f_\psi(z)-y_s)^2]$, where $z\sim q_\phi(z\vert s)$ is the latent vector encoded from states $s$.
$\beta>0$ is a hyper-parameter.

\subsection{State Generation by Maximizing Surrogate Model}
\subsubsection{Optimization in Latent Space}
After training the surrogate model $f_\psi(z)$ to predict the exploration scores of states based on their latent vectors $z$, we maximize $f_\psi(z)$ with respect to $z$ to find new latent vectors with high exploration scores.

First, the $N_s$ number of latent vectors $\mathbf{z^0}=\{z^0_1,...,z^0_{N_s}\}$ are sampled from $\mathcal{N}(0,\mathbf{I})$. Every vector $z^0_i$ for $i=1,...,N_s$ is then optimized using the stochastic gradient ascent.
\begin{equation} \label{eq:3}
z^{t+1}_i=z^t_i+\delta(\nabla_{z^t_i}f(z^t_i)+\eta_t),
\end{equation}
where $\delta$ is a step size, and $\eta_t$ is a Gaussian noise with zero mean and decreasing variance. The noise is injected to prevent the latent vectors from being stuck in local optima over a latent space. After $L$ iterations of Equation \ref{eq:3}, a set of optimized latent vectors $\mathbf{z^*}=\{z^*_1,...,z^*_{N_s}\}$ is obtained.

\subsubsection{Decoding Latent Vectors}
Once $\mathbf{z^*}$ is obtained, it is decoded to a set of new states $\mathbf{s^*}=\{s^*_1,...,s^*_{N_s}\}$ through the decoder $s_i^*=g_\theta(z_i^*)$ of VGAE. MARL then uses $\mathbf{s^*}$ as the new initial states for the next training episodes in the game.

\subsection{Training MARL Policy}

MARL policies are then trained while playing the game with the training episodes having the initial states $\mathbf{s^*}$ generated by REMAX. To maintain a certain level of default initial states, MARL uses the initial states generated from REMAX and the default initial states provided by the game with a certain probability, such as 0.8:0.2. After training MARL using $N_s$ training episodes, REMAX is trained again using newly collected states from MARL and generates new initial states for training MARL in the next round. In other words, for every $N_s$ number of training episodes, REMAX is retrained and generates new initial states for training MARL. The parameters $(\theta,\phi,\psi)$ of REMAX are reinitialized and retrained using only newly collected states from just previous training episodes to reflect the evolution of MARL. This is similar to employing curriculum learning in that the game condition is dynamically being adjusted depending on the current performance of the intermediate policy that is being trained.

\section{Experiments}

Figure \ref{fig:fig_illustrations} shows the environments (games) used to evaluate the performances of the proposed and comparison methods. The environments are designed to render more sparse rewards than existing environments \cite{lowe2017multi,smac}. We assume that the agents in the environments have the positions and velocities (or health) of all agents as a state. In the environments, we consider random exploration and GENE as main comparison methods.



\begin{itemize}
\item \textbf{Goal-oriented tasks (a) and (b)}:
In single-and multi-agent goal-oriented tasks, we additionally consider HER and RCG for goal-oriented tasks.

\item \textbf{Mixed cooperative-competitive games (c) and (d):}
We exclude HER and RCG because they are not suitable for non-goal-oriented tasks. Instead, we additionally consider EDTI, which uses intrinsic rewards that quantify the influence of one agent on others.

\end{itemize}


We use MADDPG as the MARL model. Note that any MARL model can be used; however, only MADDPG, one of the most general MARL models, is used as a representative
in this study because the focus of this study is on investigating the effectiveness of exploration methods, irrespective of the MARL model used. 
In addition, all results in this study were averaged across five different seeds.
    

\begin{table}[t]
  \caption{Number of training episodes in maze.}
  \label{table:results_maze}
  \centering
    \begin{tabular}{ccc}
        \toprule
    
        \multicolumn{1}{c}{} & \multicolumn{1}{c}{Latent space dim.}&
        \multicolumn{1}{c}{Episode($\times$10)}\\
        \midrule
        \multicolumn{1}{c}{Random} &-& 640\scriptsize $\pm$ 223\\
        \multicolumn{1}{c}{HER} &-& 601\scriptsize $\pm$ 227\\
        \multicolumn{1}{c}{RCG} &-& 574\scriptsize $\pm$ 198\\
         \multicolumn{1}{c}{GENE}  &1&559\scriptsize $\pm$255\\
         \multicolumn{1}{c}{REMAX}&1& \textbf{456}\scriptsize $\pm$202\\ 
         \multicolumn{1}{c}{REMAX}&2&  495\scriptsize $\pm$139\\
        \bottomrule
    \end{tabular}
\end{table}

\begin{figure*}[t]
\begin{minipage}[c]{.153\textwidth}
  \centering
  \includegraphics[width=0.85\textwidth]{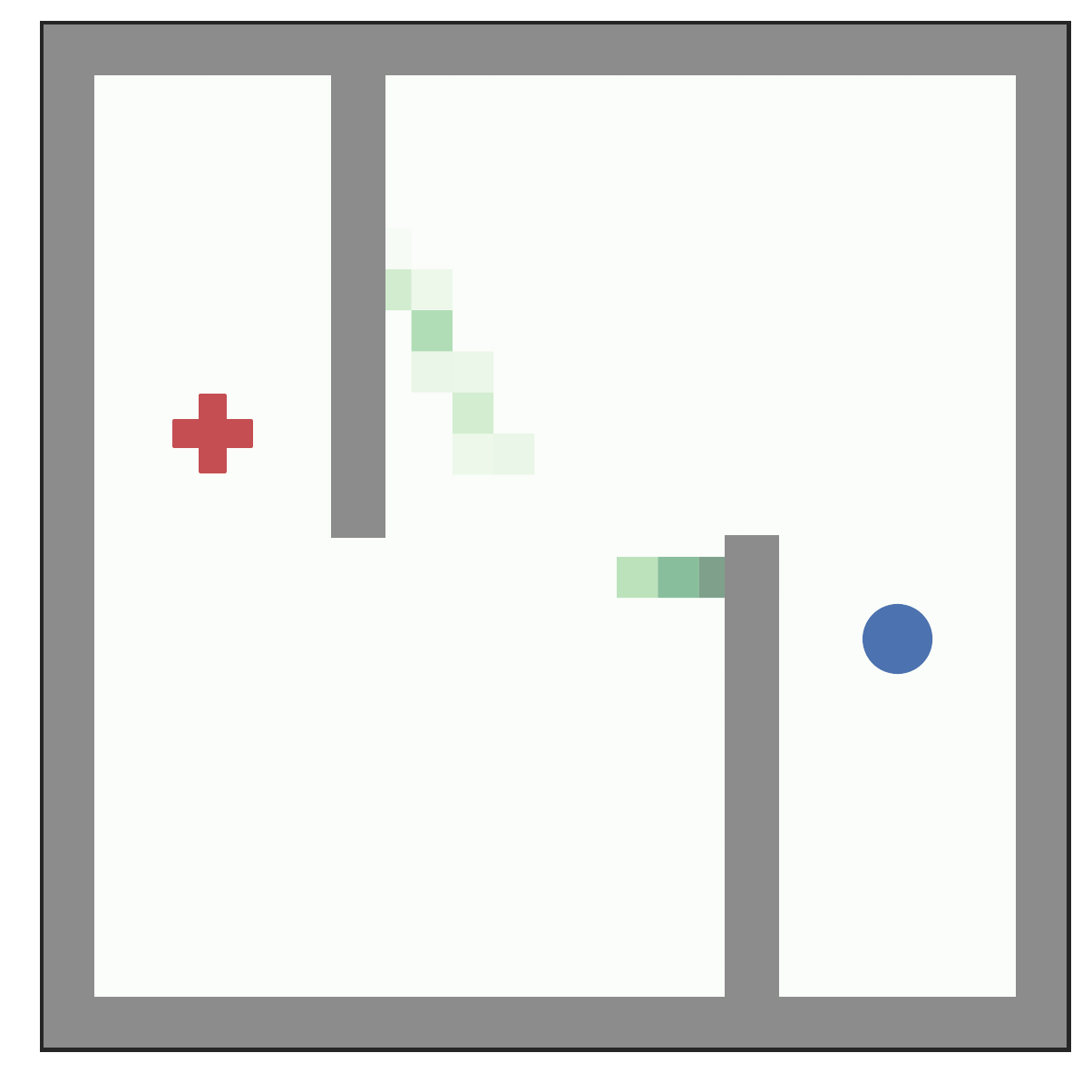}
\end{minipage}
\hspace{0.0cm}
\begin{minipage}[c]{.153\textwidth}
  \centering
  \includegraphics[width=0.85\textwidth]{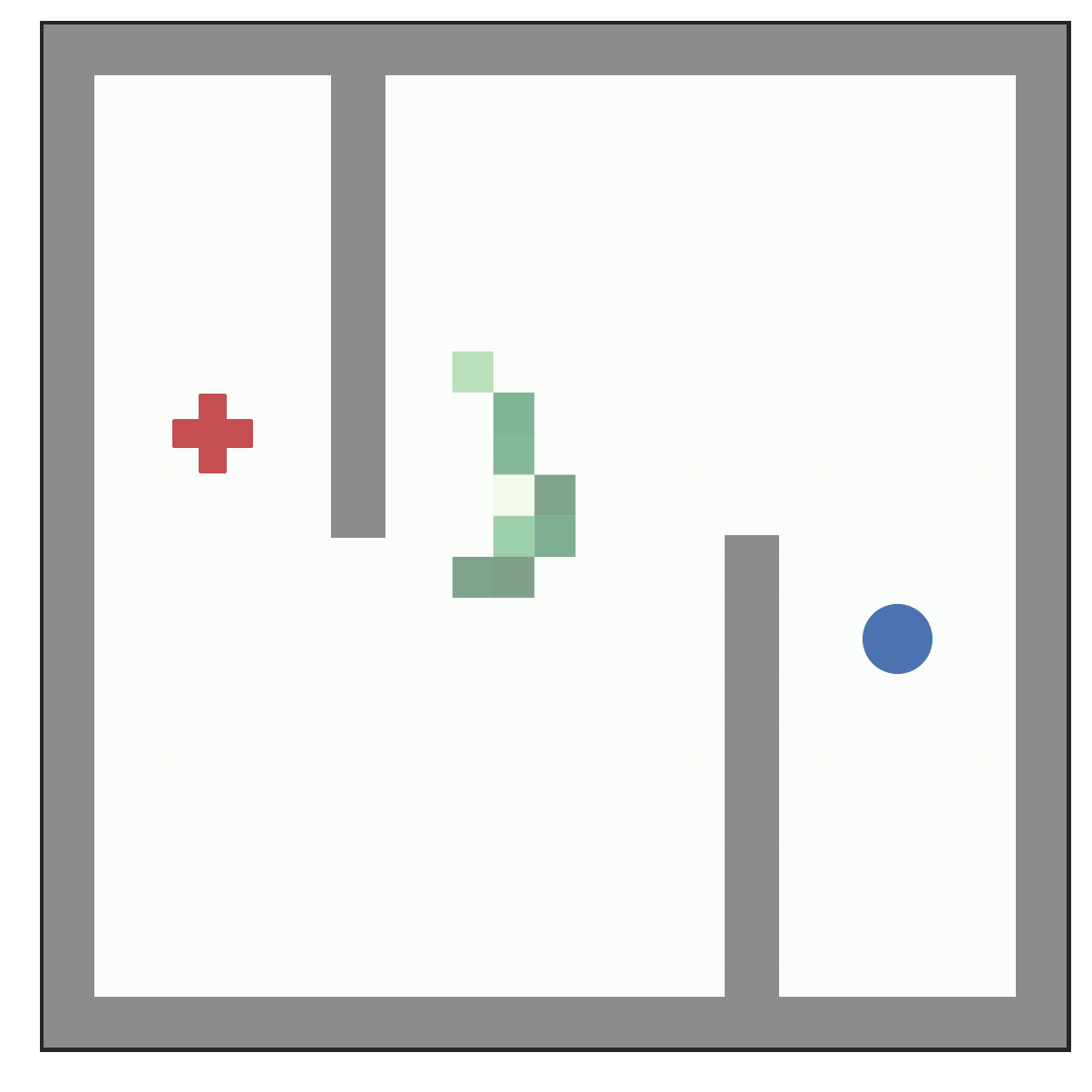}
\end{minipage}
\hspace{0.0cm}
\begin{minipage}[c]{.153\textwidth}
  \centering
  \includegraphics[width=0.85\textwidth]{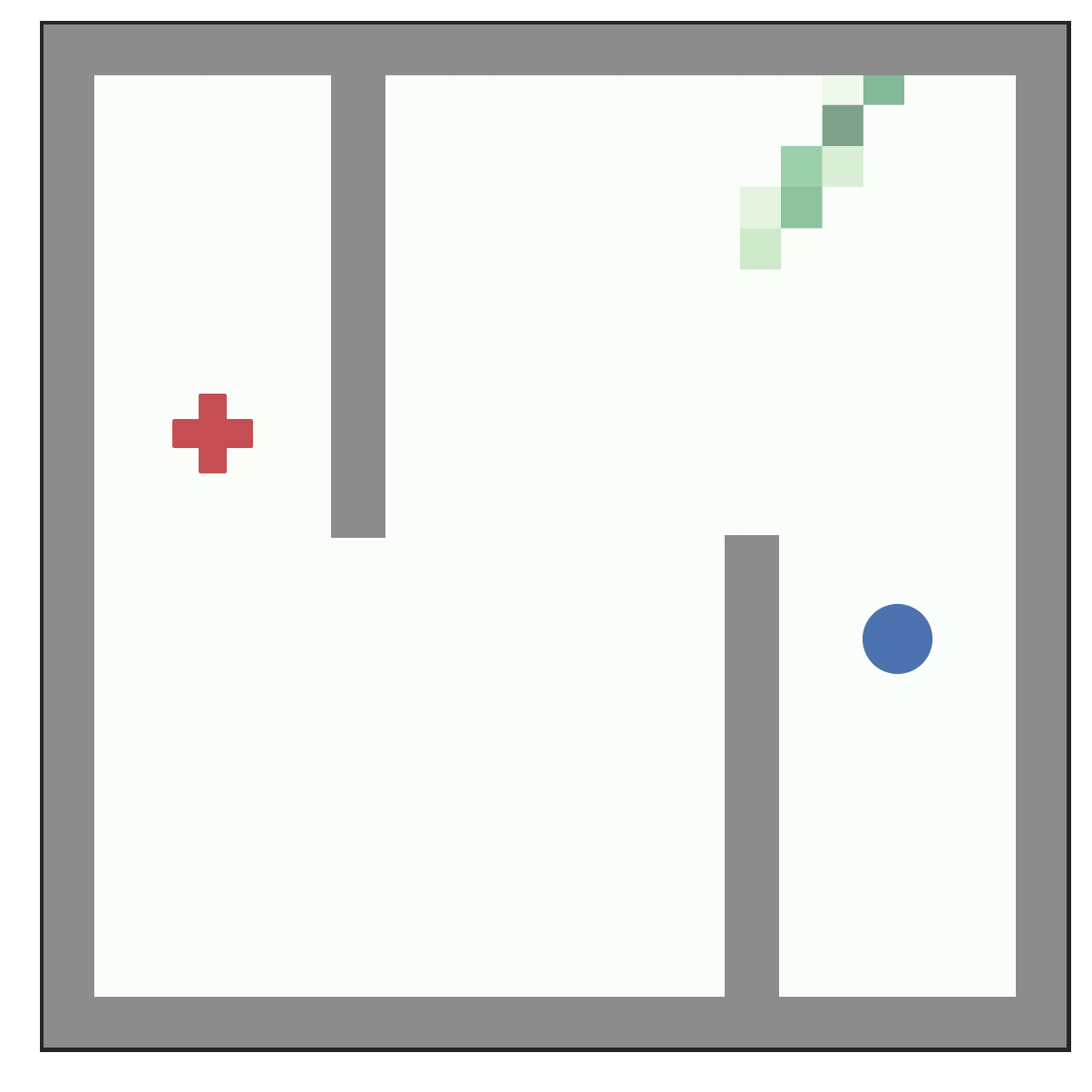}
\end{minipage}
\hspace{0.0cm}
\begin{minipage}[c]{.153\textwidth}
  \centering
  \includegraphics[width=0.85\textwidth]{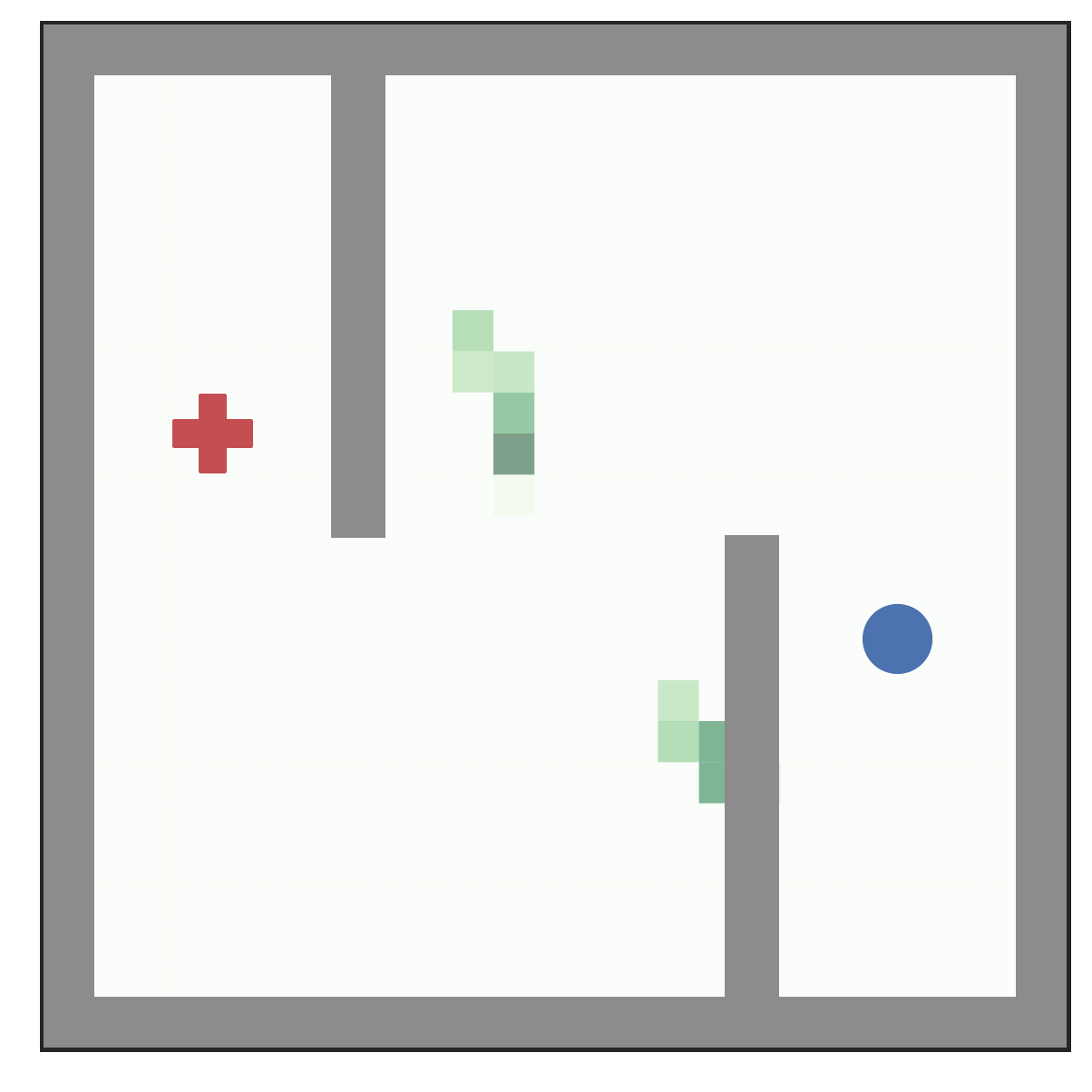}
\end{minipage}
\hspace{0.0cm}
\begin{minipage}[c]{.153\textwidth}
  \centering
  \includegraphics[width=0.85\textwidth]{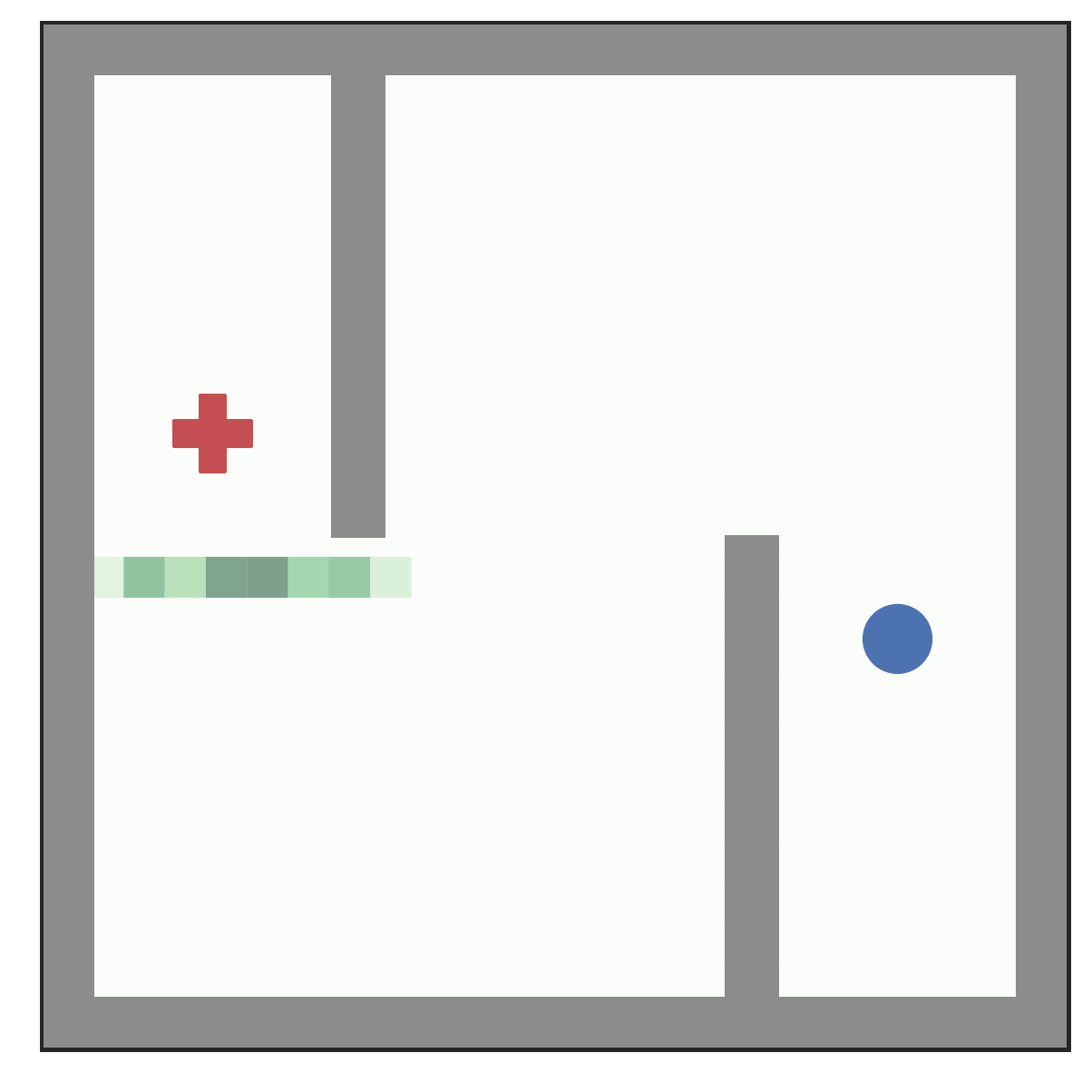}
\end{minipage}
\hspace{0.0cm}
\begin{minipage}[c]{.153\textwidth}
  \centering
  \includegraphics[width=0.85\textwidth]{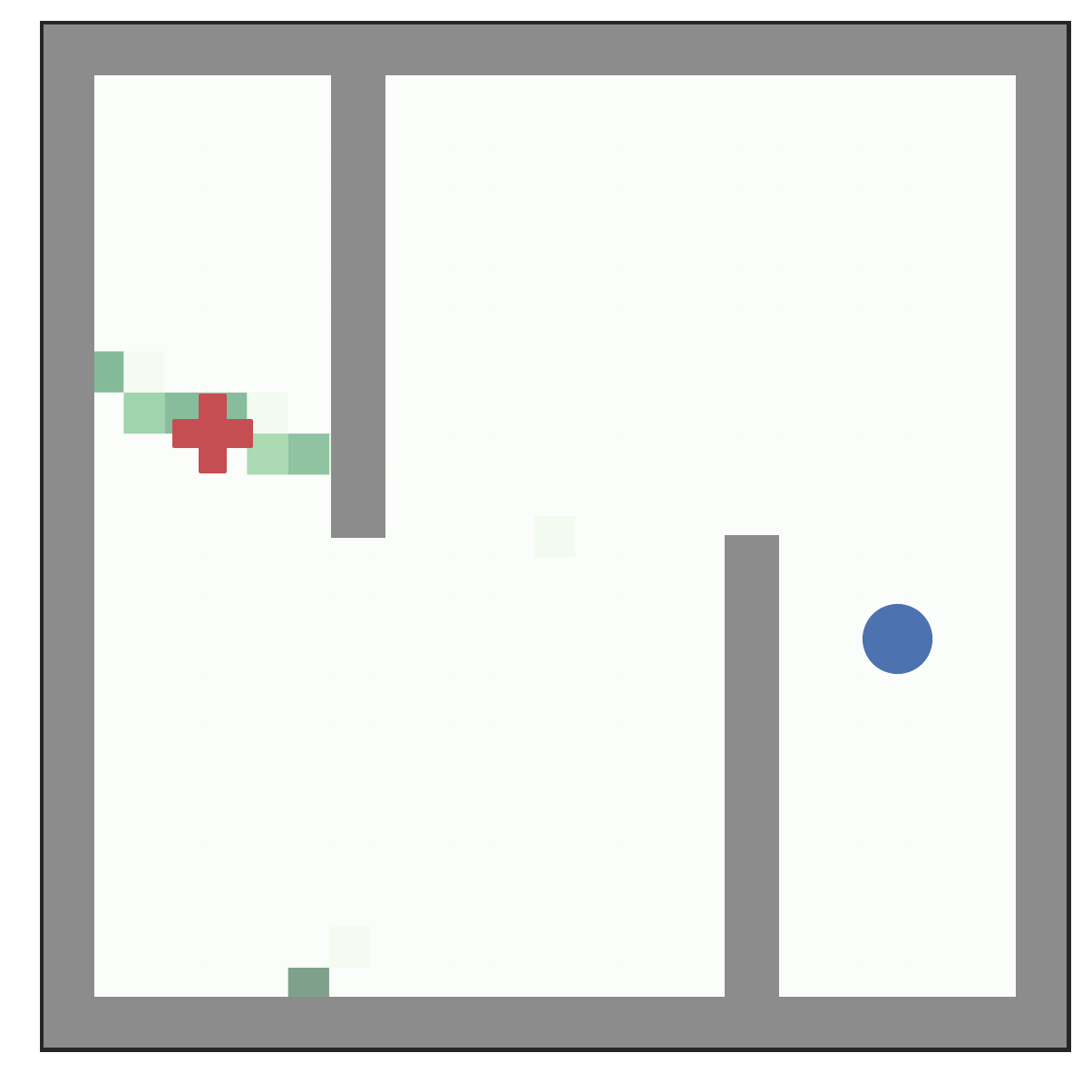}
\end{minipage}

\begin{minipage}[c]{.153\textwidth}
  \centering
  \includegraphics[width=0.84\textwidth]{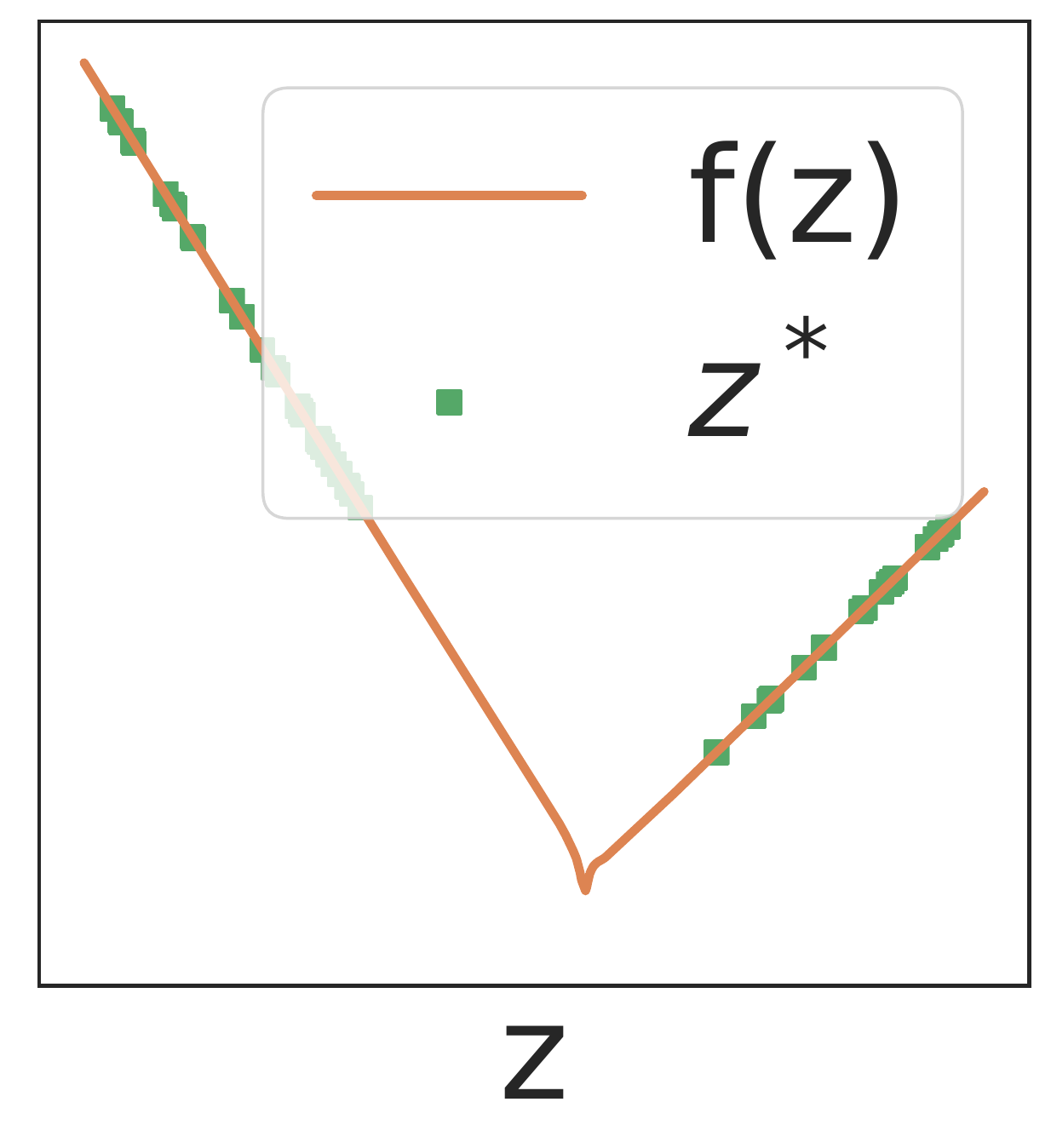}
\end{minipage}
\hspace{0.0cm}
\begin{minipage}[c]{.153\textwidth}
  \centering
  \includegraphics[width=0.84\textwidth]{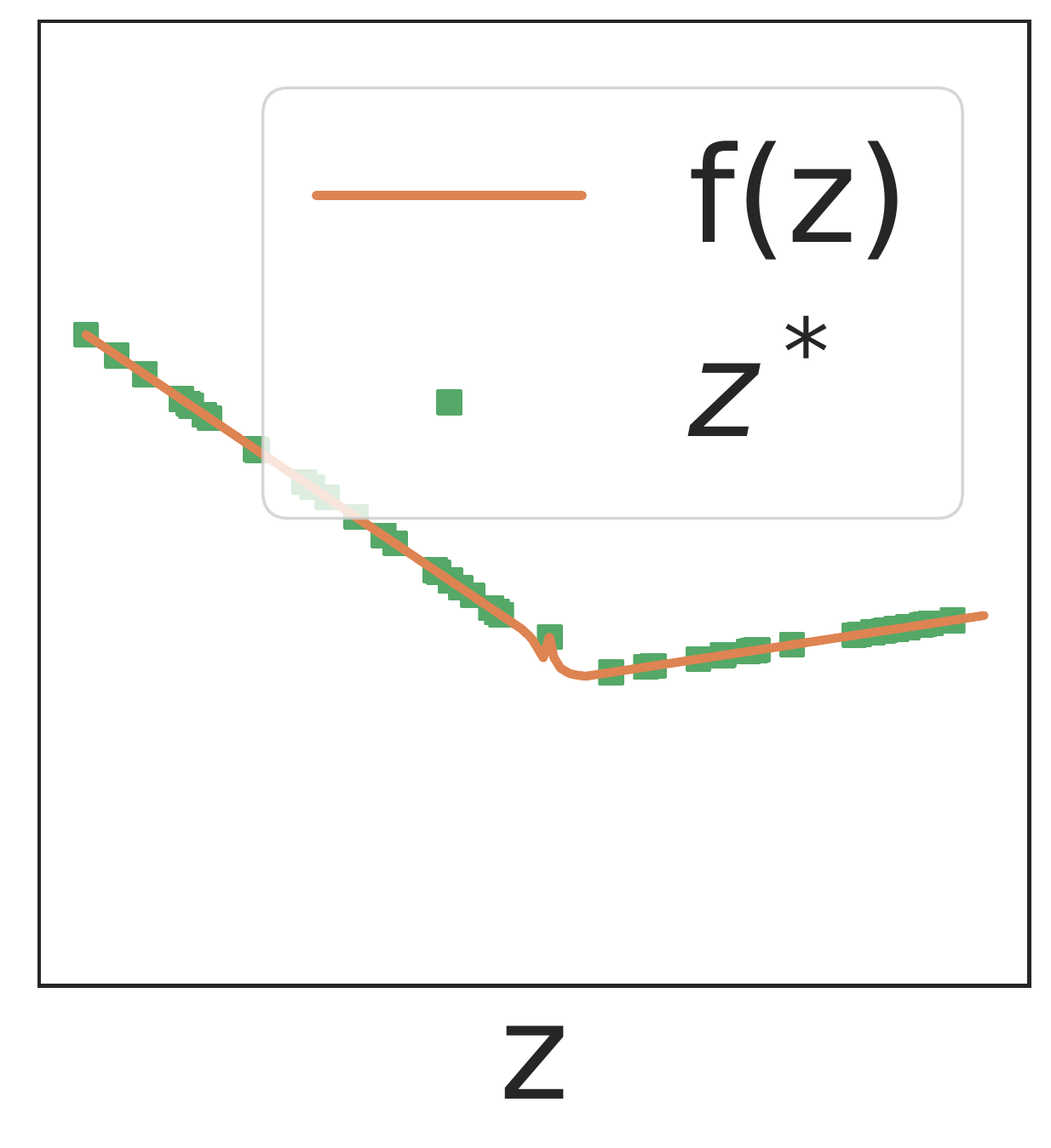}
\end{minipage}
\hspace{0.0cm}
\begin{minipage}[c]{.153\textwidth}
  \centering
  \includegraphics[width=0.84\textwidth]{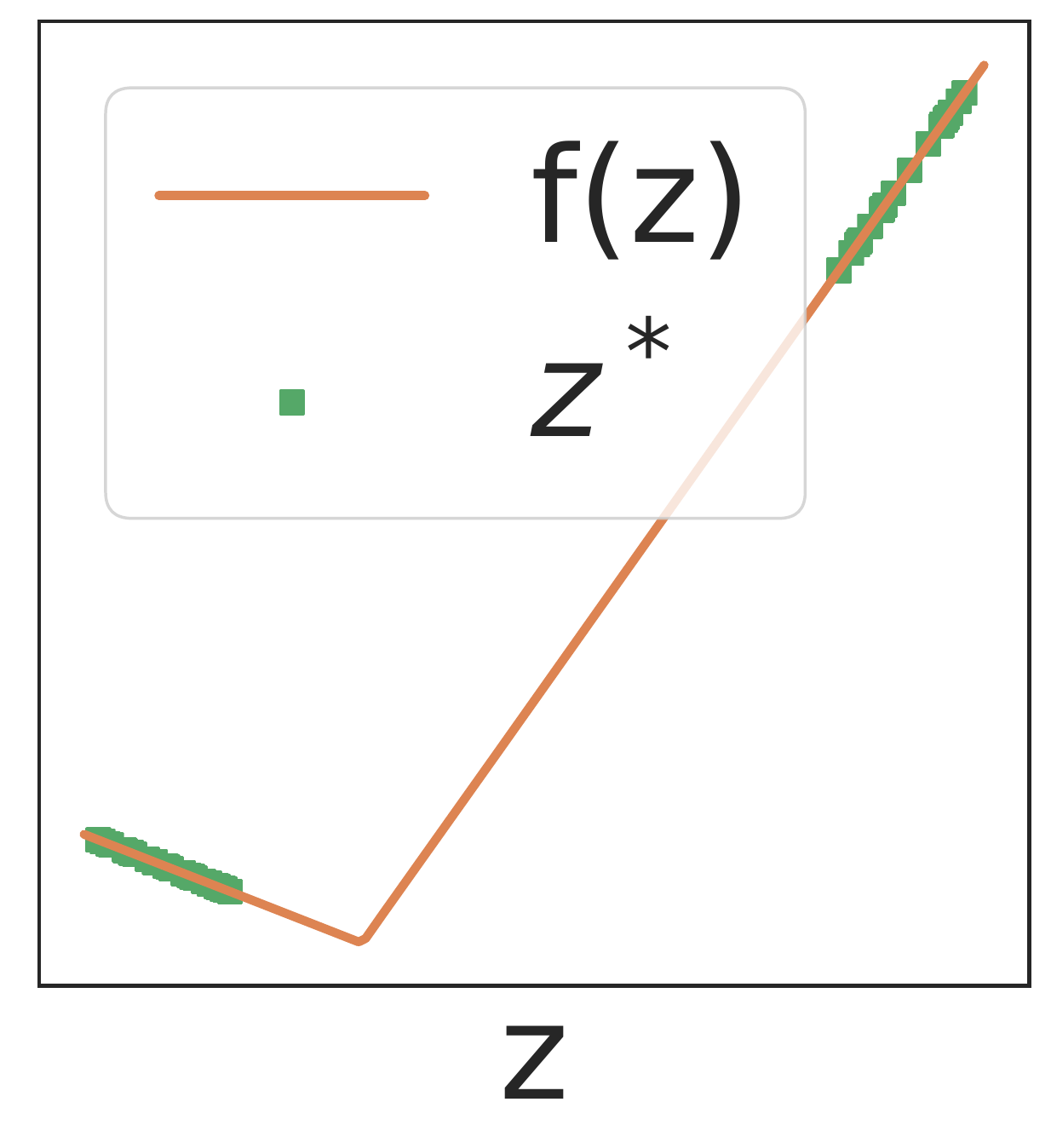}
\end{minipage}
\hspace{0.0cm}
\begin{minipage}[c]{.153\textwidth}
  \centering
  \includegraphics[width=0.84\textwidth]{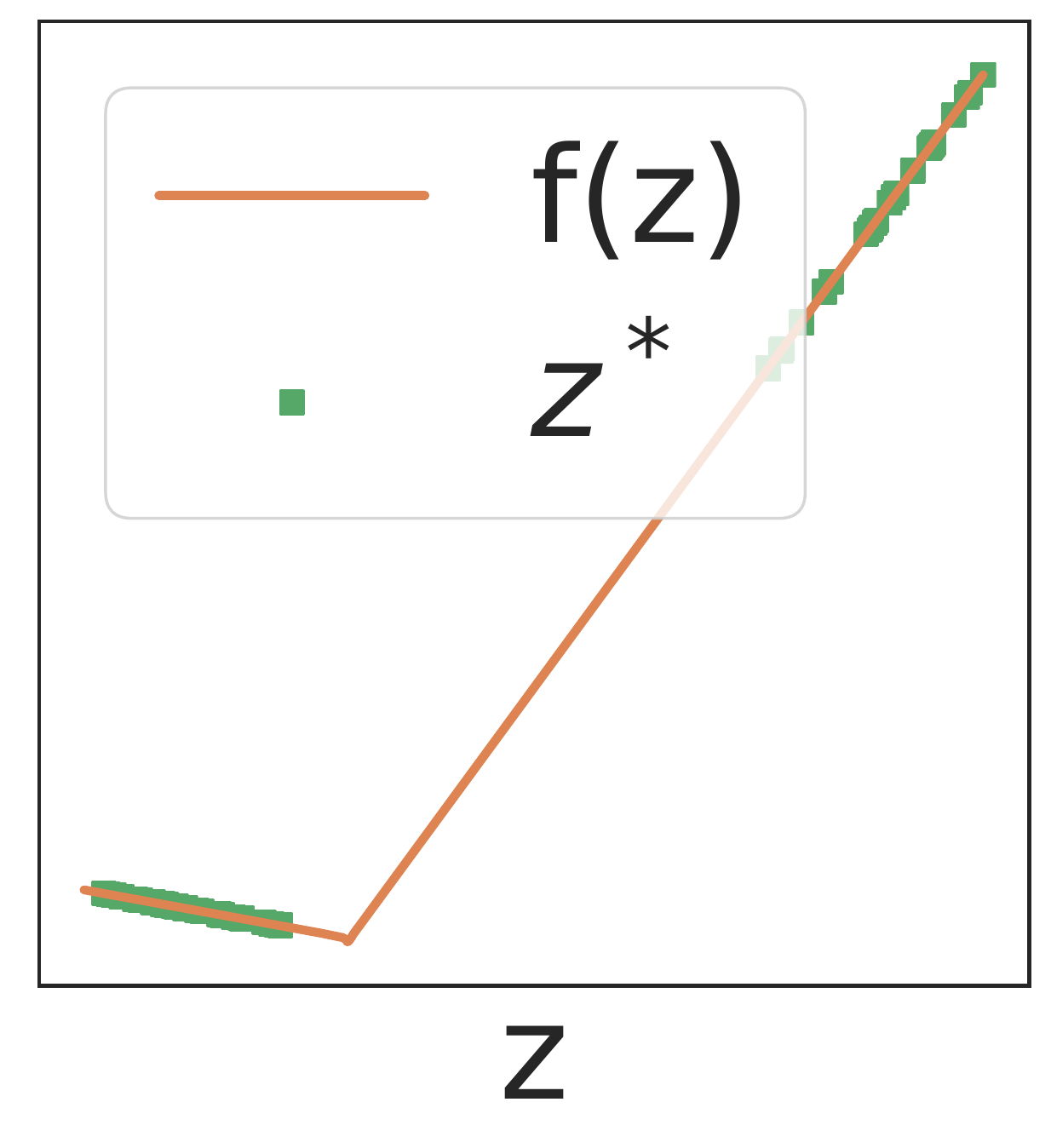}
\end{minipage}
\hspace{0.0cm}
\begin{minipage}[c]{.153\textwidth}
  \centering
  \includegraphics[width=0.84\textwidth]{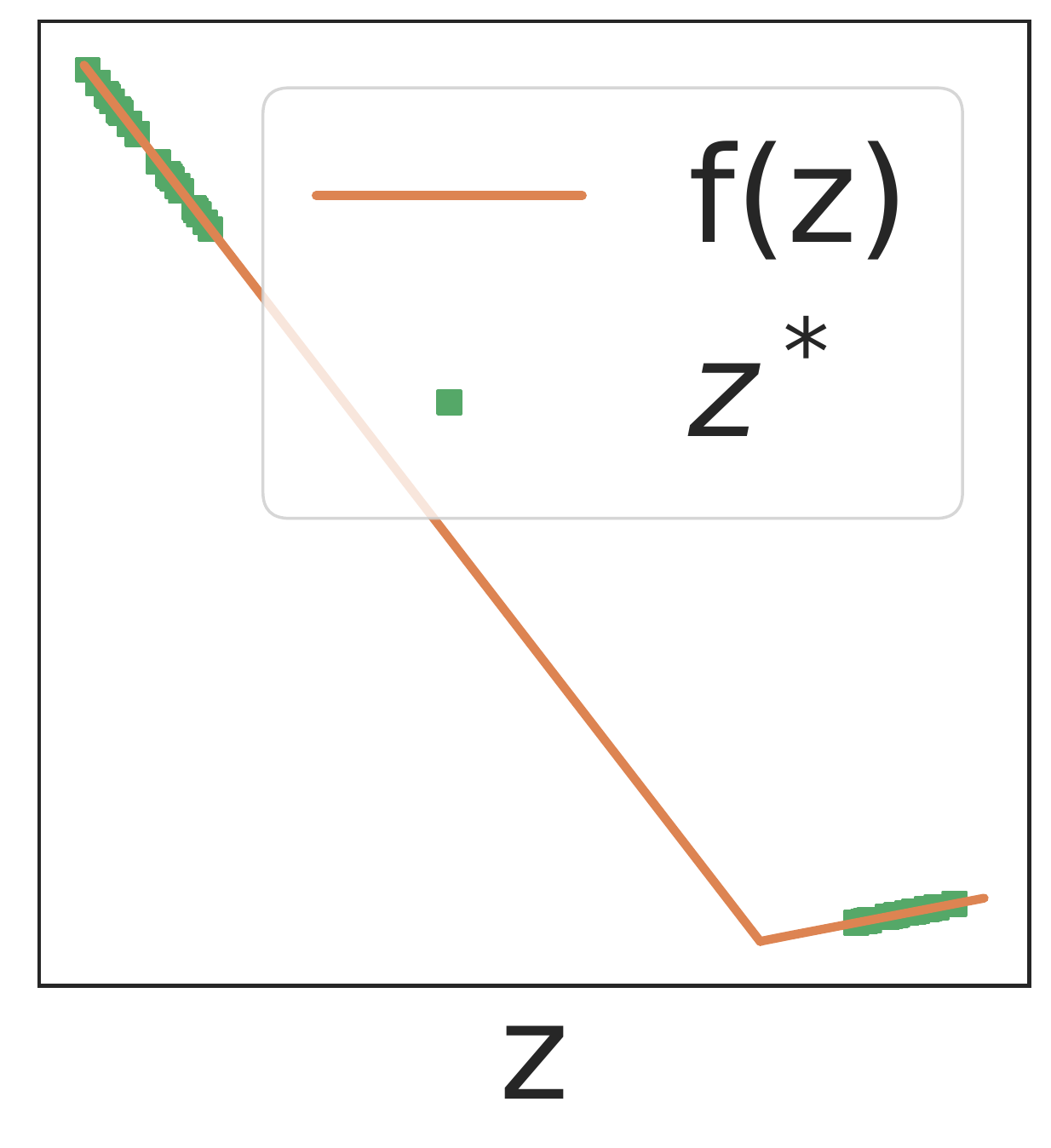}
\end{minipage}
\hspace{0.0cm}
\begin{minipage}[c]{.153\textwidth}
  \centering
  \includegraphics[width=0.84\textwidth]{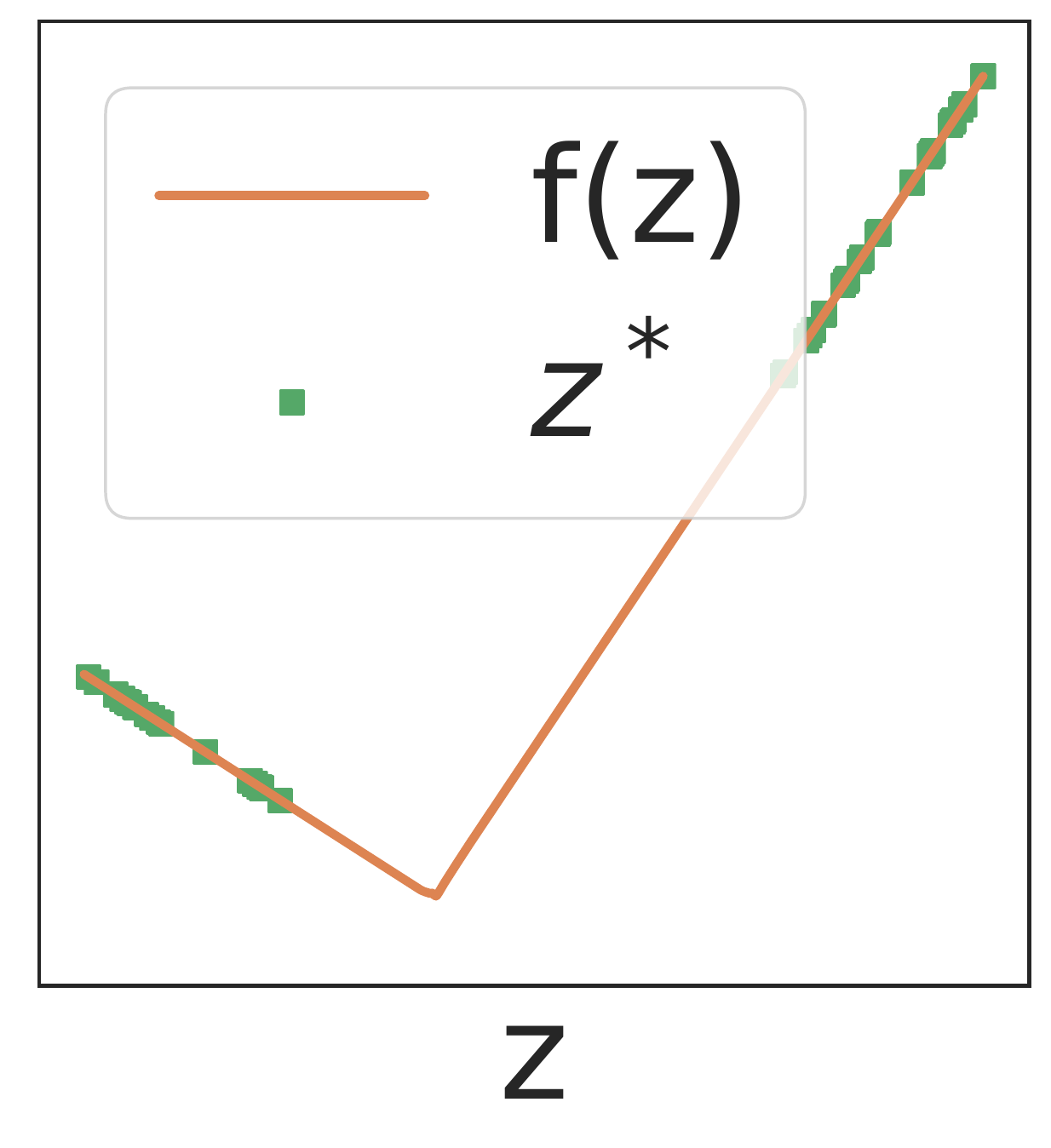}
\end{minipage}
\hspace{0.0cm}

\begin{minipage}[c]{.99\textwidth}
  \centering
  \includegraphics[width=0.92\textwidth]{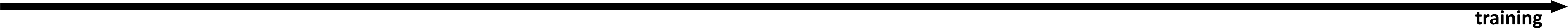}
\end{minipage}
 \caption{Generated states and surrogate models with one-dimensional latent space in maze.}
  \label{fig:fig_maze_1d}
\end{figure*}

\begin{table*}[t]
  \caption{Number of training episodes ($\times$10) in cooperative navigation. In the table, ``$>$2000'' indicates that the policy cannot be learned to complete the task by the corresponding method within the given number of training episodes.}
  \label{table:results_cn}
  \centering
    \begin{tabular}{cccccccccc}
        \toprule
        \multicolumn{1}{c}{} &  \multicolumn{4}{c}{Additional architecture} & \multicolumn{5}{c}{Number of agents}\\
        \cmidrule(r){2-5}
        \cmidrule(r){6-10}
        \multicolumn{1}{c}{} & \multicolumn{1}{c}{VAE}&
        \multicolumn{1}{c}{KDE}&
        \multicolumn{1}{c}{GAT}&
        \multicolumn{1}{c}{Surrogate}&
        \multicolumn{1}{c}{2}&
        \multicolumn{1}{c}{3}&
        \multicolumn{1}{c}{4}&
        \multicolumn{1}{c}{6}&
        \multicolumn{1}{c}{8}\\
        \midrule
        \multicolumn{1}{c}{Random} & & & & &583\scriptsize $\pm$ 271 & 834\scriptsize $\pm$ 168 & 1178\scriptsize $\pm$ 542 & $>$2000 & $>$2000\\
        \multicolumn{1}{c}{HER} & & & & &489\scriptsize $\pm$ 140 & 790\scriptsize $\pm$ 217 & 1071\scriptsize $\pm$ 618 & $>$2000 & $>$2000\\
        \multicolumn{1}{c}{RCG} & & & & &353\scriptsize $\pm$ 133 & 674\scriptsize $\pm$ 309 & 877\scriptsize $\pm$ 679 & $>$2000 & $>$2000\\
         \multicolumn{1}{c}{GENE}  &\checkmark&\checkmark& & & 342\scriptsize $\pm$ 145 & 632\scriptsize $\pm$ 265 & 742\scriptsize $\pm$ 540 & 1401\scriptsize $\pm$ 625 & $>$2000\\
         \multicolumn{1}{c}{GENE with GAT}&\checkmark& \checkmark&\checkmark& &177\scriptsize $\pm$ 64$\;$ & 400\scriptsize $\pm$256 &
         454\scriptsize $\pm$310 & 920\scriptsize $\pm$ 551 & 1204\scriptsize $\pm$ 698\\
         \multicolumn{1}{c}{REMAX}&\checkmark& &\checkmark&\checkmark&\textbf{161}\scriptsize $\pm$ 69$\;$ & \textbf{272}\scriptsize $\pm$80 $\,$&
         \textbf{331}\scriptsize $\pm$114 & \textbf{602} \scriptsize $\pm$ 287 & \textbf{908} \scriptsize $\pm$ 601\\
        \bottomrule
    \end{tabular}
\end{table*}

\subsection{Maze}
Maze, shown in Figure \ref{fig:fig_illustrations} (a), is a single-agent environment which requires an agent (blue circle) to search for a landmark (red cross). The agent obtains +1 as a reward when it reaches the landmark, and 0 otherwise. Because there is only one agent in this environment, considering the relationships among agents is unnecessary. Thus, REMAX has VGAE without GAT, which is similar to VAE. Although both REMAX and GENE use VAE, they are different in that REMAX uses the surrogate model while GENE uses KDE.


Table \ref{table:results_maze} presents the number of training episodes required for training a policy to complete the task. Here, completing a task is defined as achieving ten consecutive successes, starting from the initial states provided by the environment. Thus, a smaller number in the table means that fewer training episodes are needed to train the successful policy. The latent space dimension denotes the dimension of the latent vector $z$.
GENE is reported as having the best performance with a one-dimensional latent space. The table shows that REMAX requires fewer episodes than HER (a goal-conditioning method) and RCG, although it does not need to specify the goal state during training. Note that to train and run RCG, one needs to specify the goal state. We believe that REMAX performs better than GENE because it trains the VAE and the surrogate model together in an end-to-end learning, while GENE separately trains VAE and KDE.

\subsubsection{Analysis of Generating States}
The first-row plots in Figure \ref{fig:fig_maze_1d} show the distribution of generated states $\mathbf{s^*}$ as green dots, and the second-row plots show how the surrogate model $f_\psi(z)$ and the optimized latent vectors $\mathbf{z^*}$ change with the training's progress. These states $\mathbf{s^*}$ are generated by decoding 
$\mathbf{z^*}$. 
As MARL and REMAX are trained with more samples, as shown in the figures, REMAX tends to generate states near the landmark because it learns that these states are easily rewardable and helpful for MARL training.


\begin{figure*}[t]
\begin{minipage}[c]{.153\textwidth}
  \centering
  \includegraphics[width=0.85\textwidth]{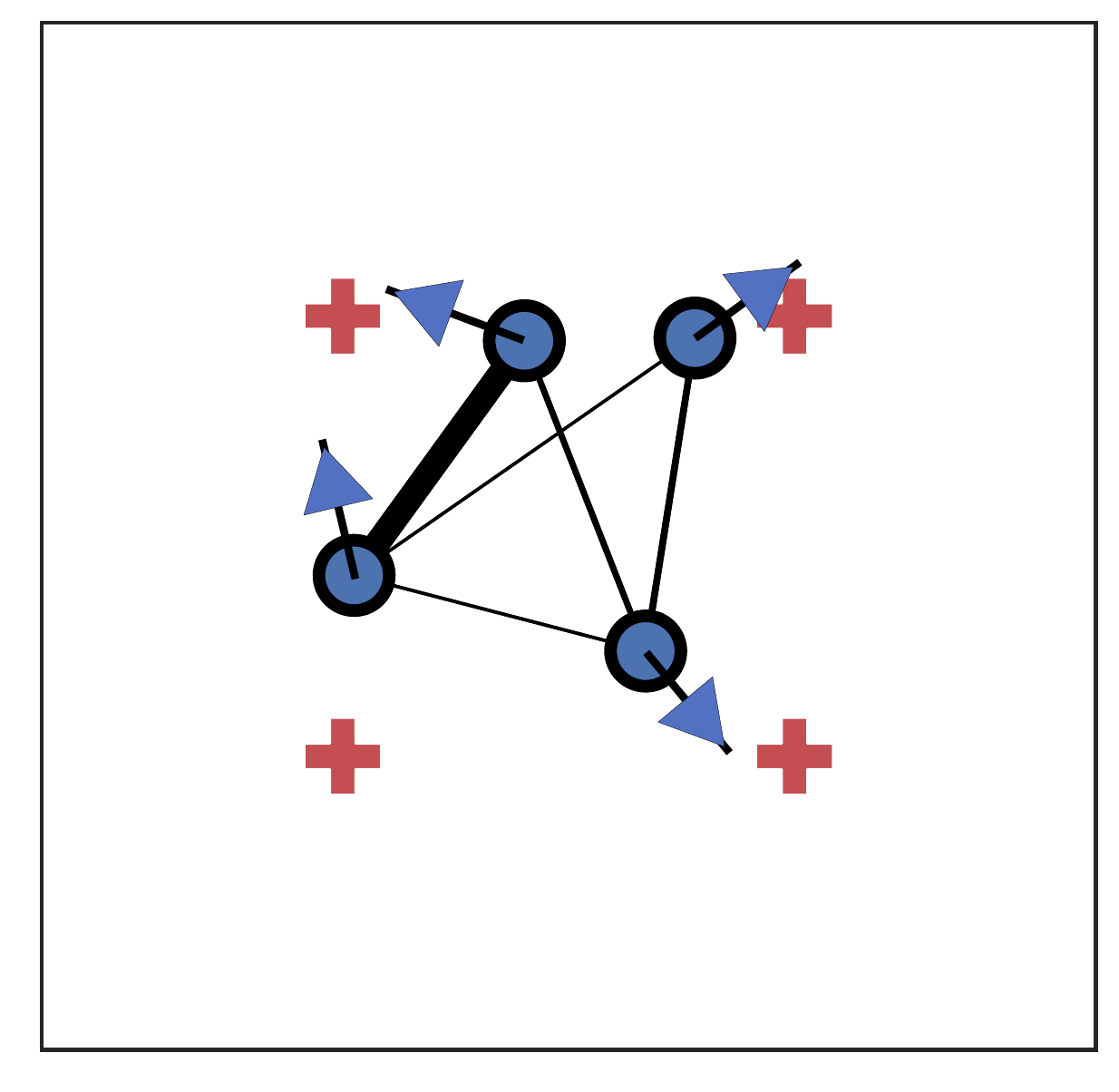}
\end{minipage}
\hspace{0.0cm}
\begin{minipage}[c]{.153\textwidth}
  \centering
  \includegraphics[width=0.85\textwidth]{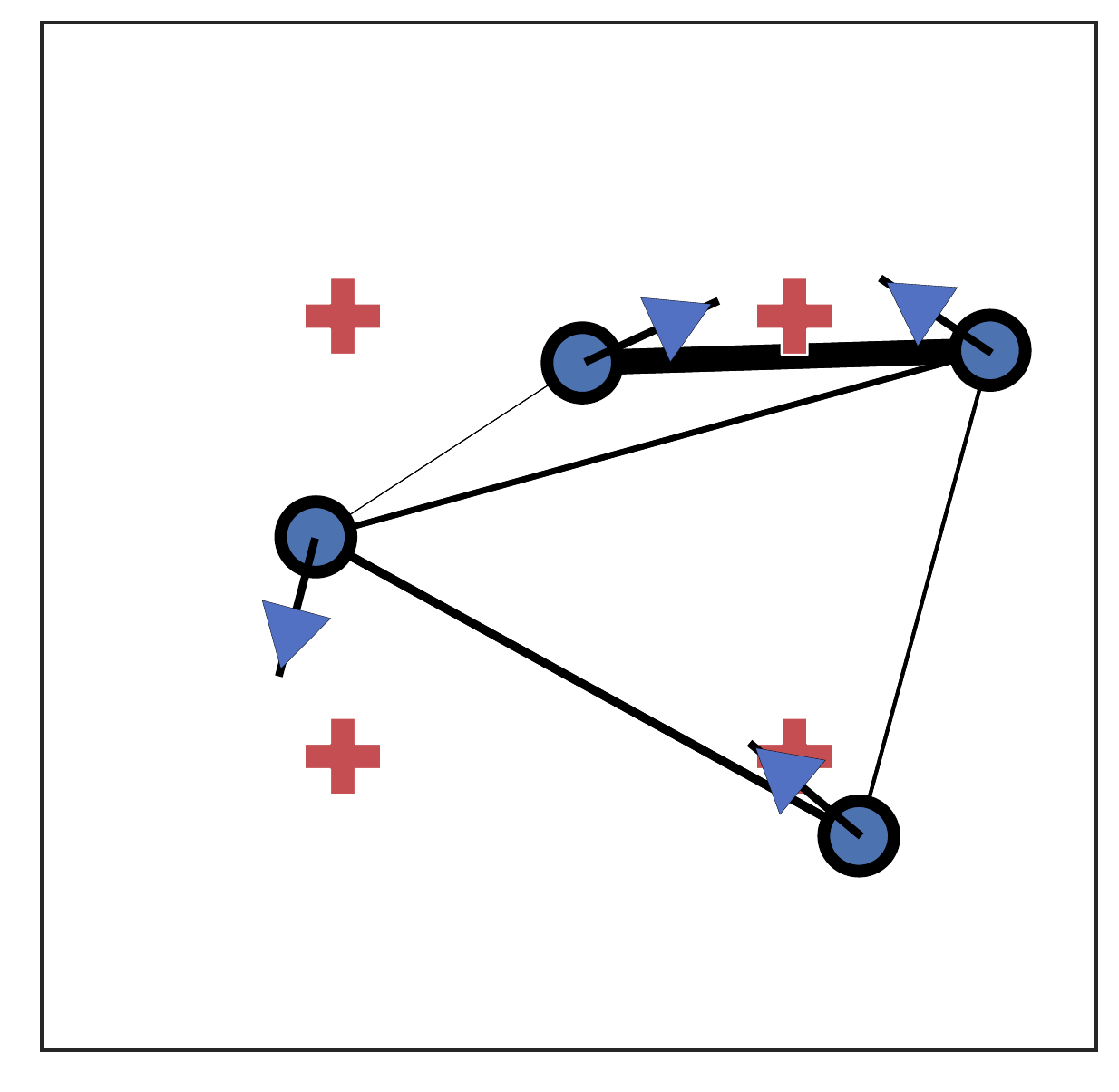}
\end{minipage}
\hspace{0.0cm}
\begin{minipage}[c]{.153\textwidth}
  \centering
  \includegraphics[width=0.85\textwidth]{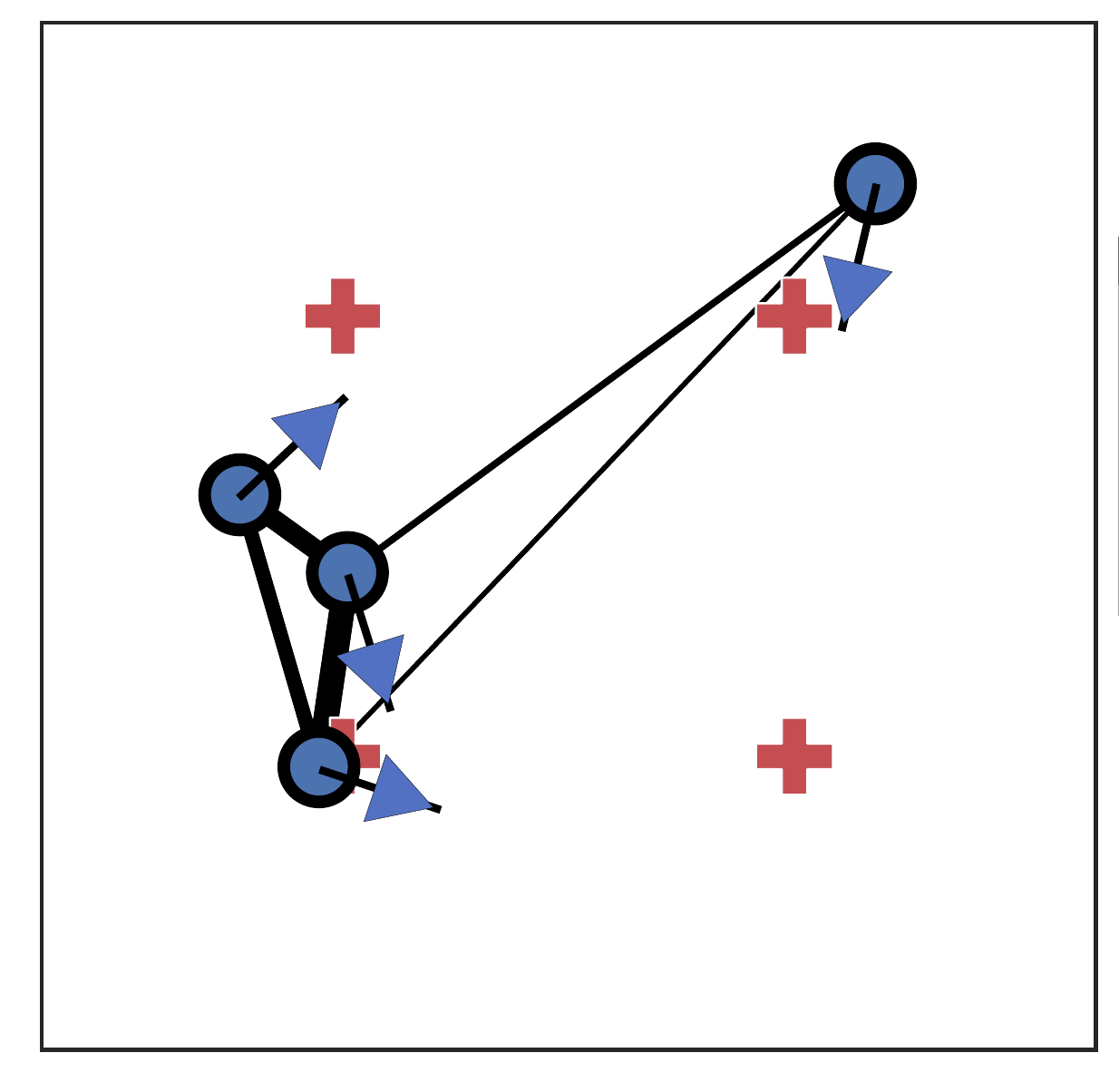}
\end{minipage}
\hspace{0.0cm}
\begin{minipage}[c]{.153\textwidth}
  \centering
  \includegraphics[width=0.85\textwidth]{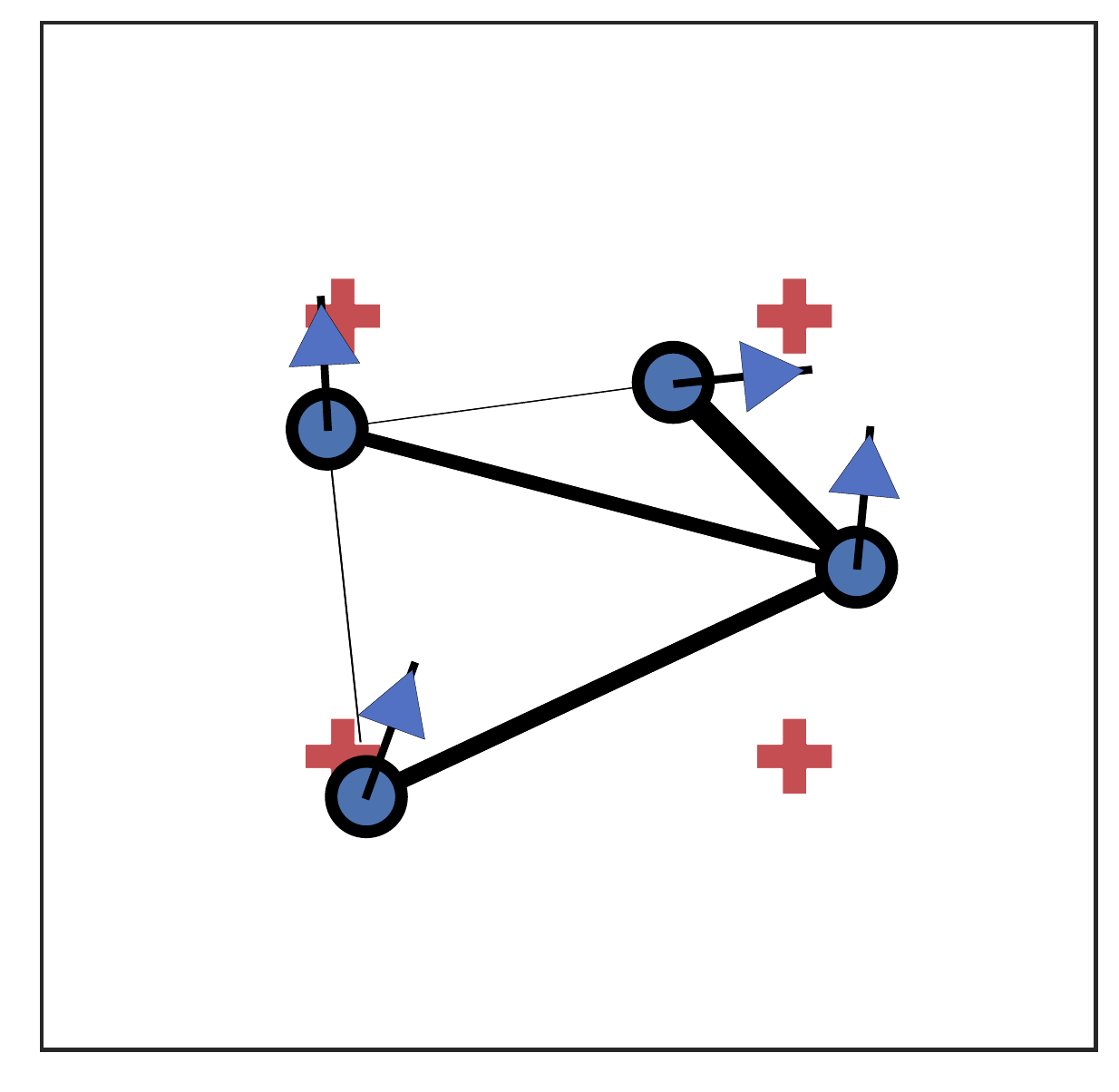}
\end{minipage}
\hspace{0.0cm}
\begin{minipage}[c]{.153\textwidth}
  \centering
  \includegraphics[width=0.85\textwidth]{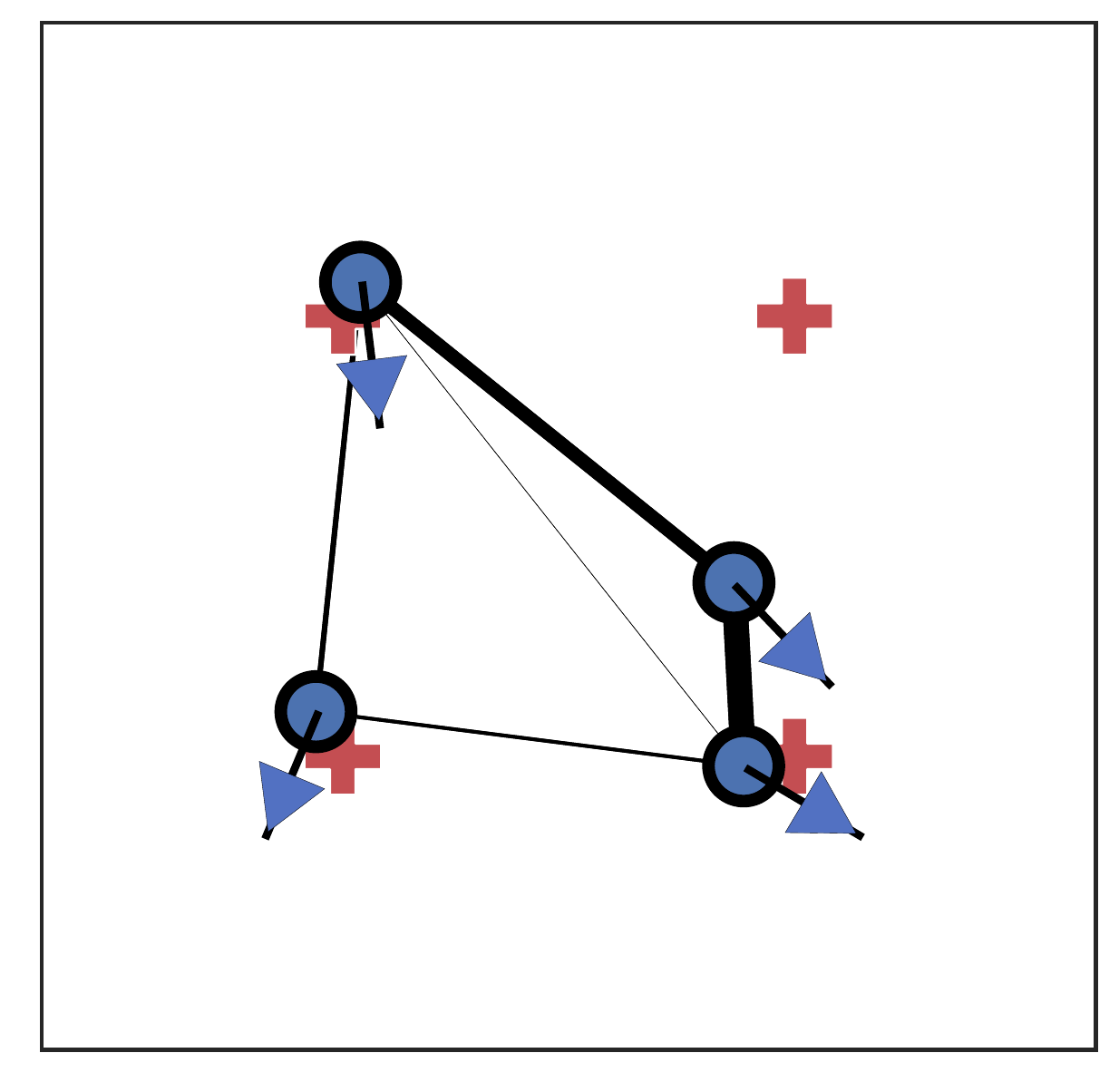}
\end{minipage}
\hspace{0.0cm}
\begin{minipage}[c]{.153\textwidth}
  \centering
  \includegraphics[width=0.85\textwidth]{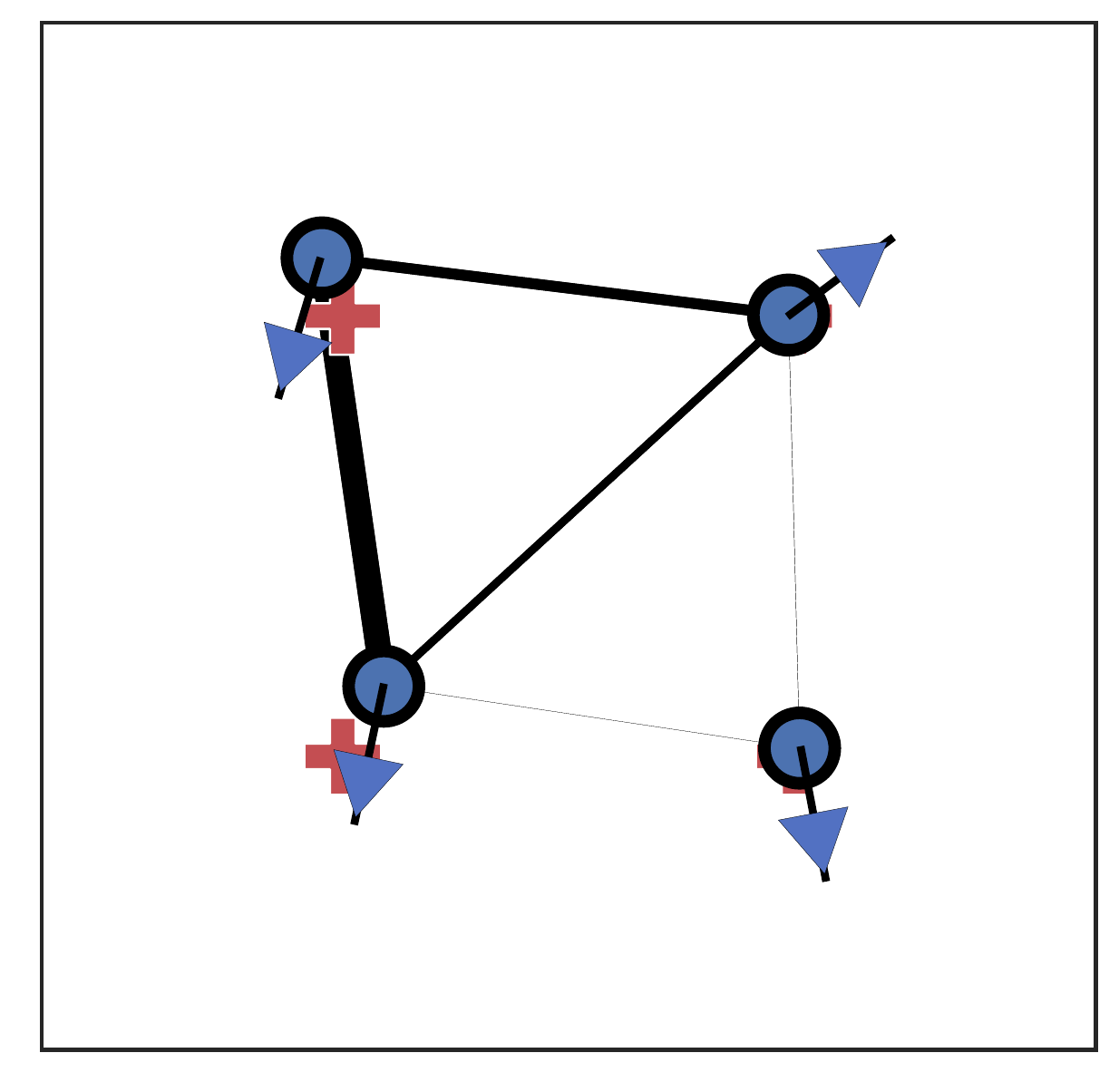}
\end{minipage}

\begin{minipage}[c]{.99\textwidth}
  \centering
  \includegraphics[width=0.92\textwidth]{method_figure_0512_time.pdf}
\end{minipage}
\caption{Attention of GAT in REMAX in cooperative navigation with 4 agents.}
\label{fig:fig_coopnavi_attention}
\end{figure*}

\begin{figure*}[t!]

\begin{minipage}[c]{.153\textwidth}
  \centering
  \includegraphics[width=0.85\textwidth]{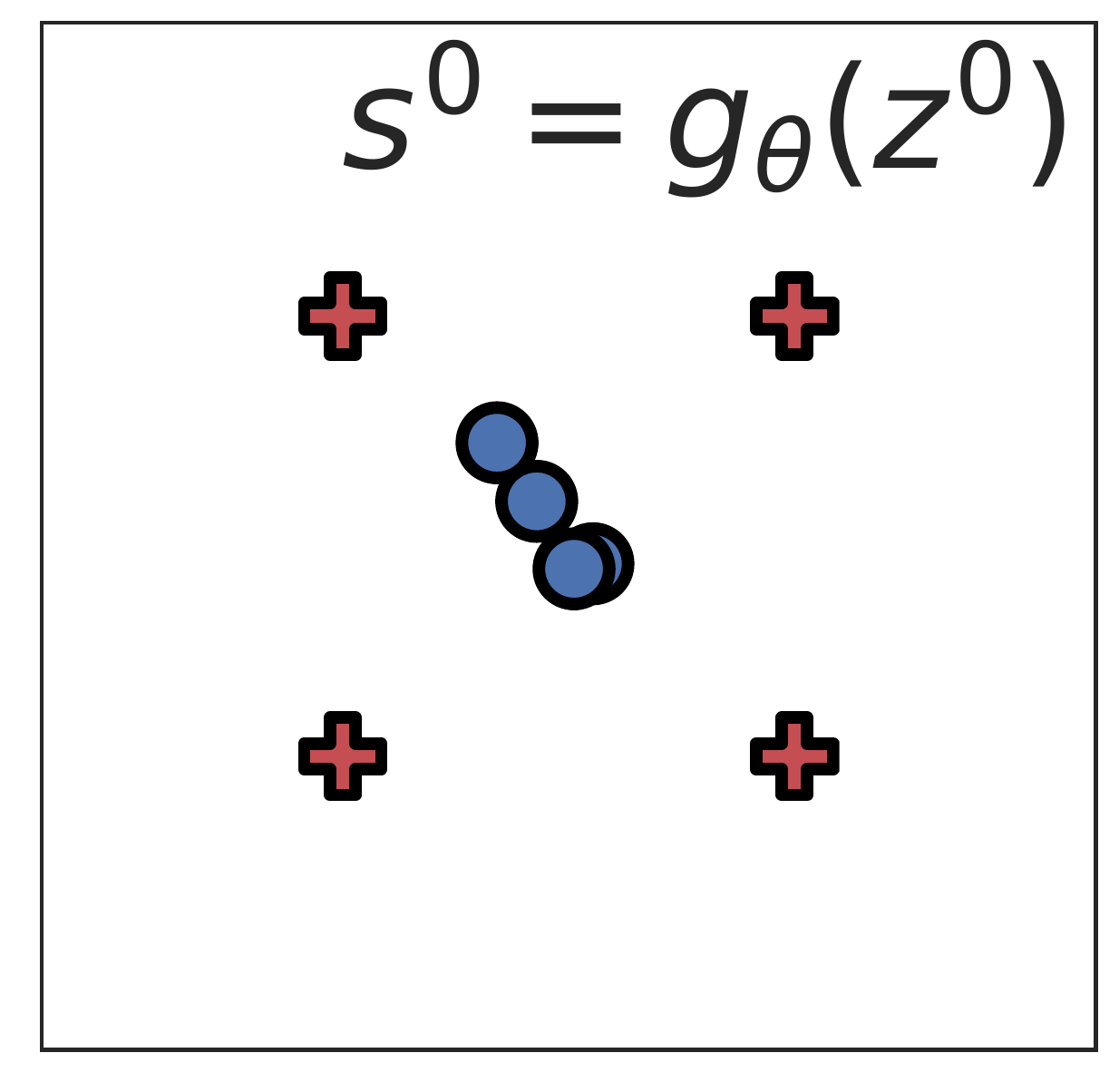}
\end{minipage}
\hspace{0.0cm}
\begin{minipage}[c]{.153\textwidth}
  \centering
  \includegraphics[width=0.85\textwidth]{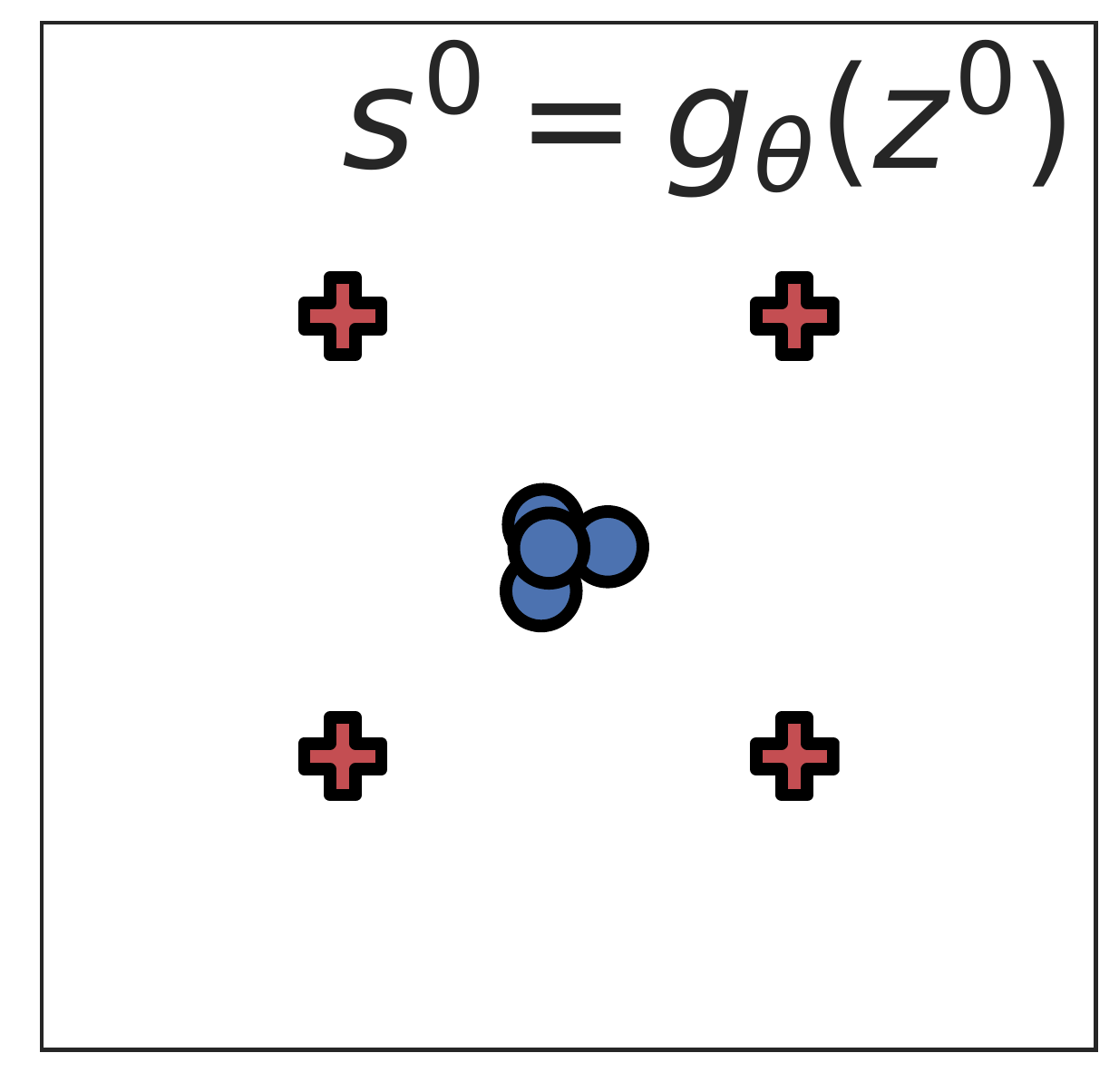}
\end{minipage}
\hspace{0.0cm}
\begin{minipage}[c]{.153\textwidth}
  \centering
  \includegraphics[width=0.85\textwidth]{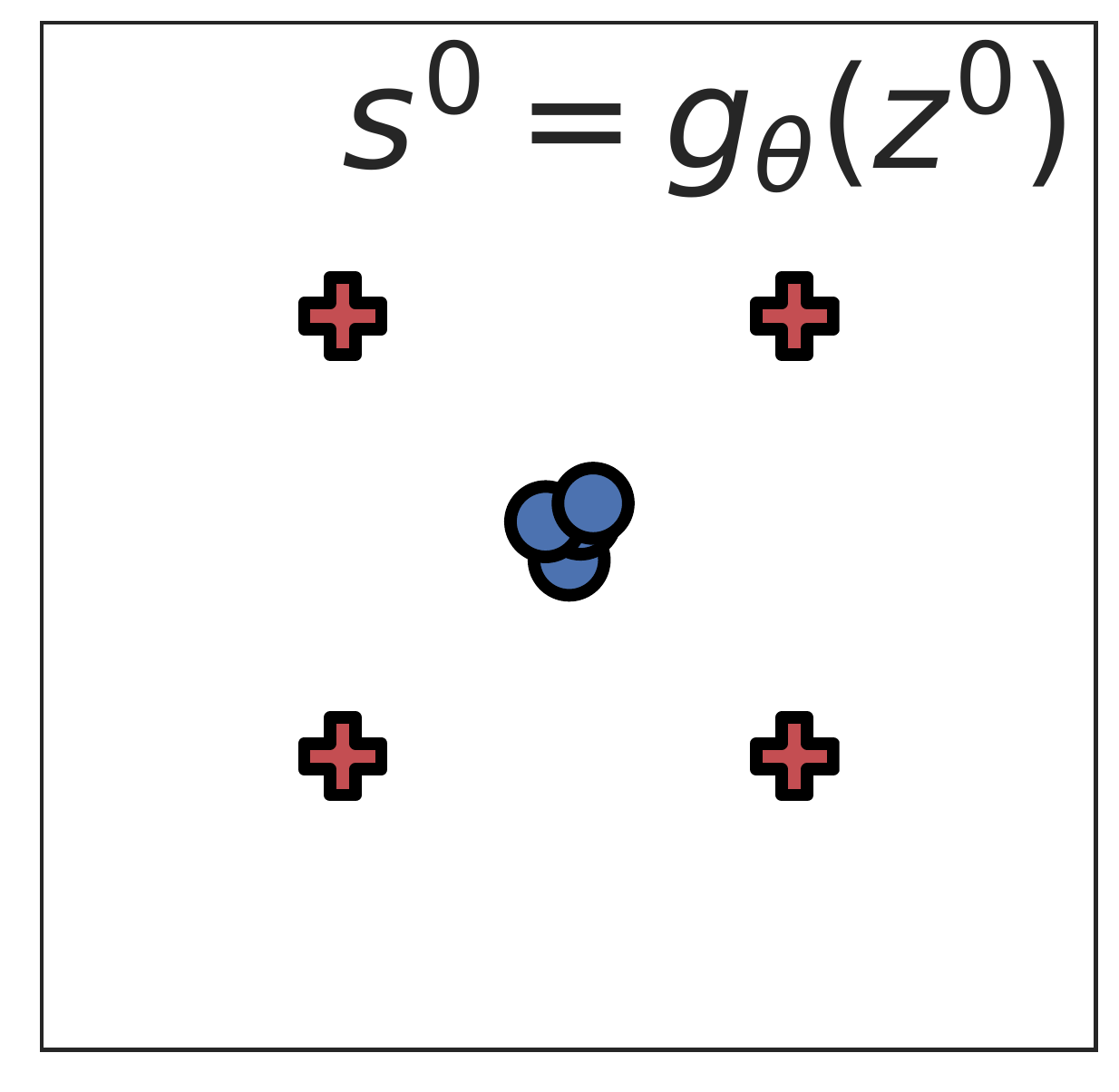}
\end{minipage}
\hspace{0.0cm}
\begin{minipage}[c]{.153\textwidth}
  \centering
  \includegraphics[width=0.85\textwidth]{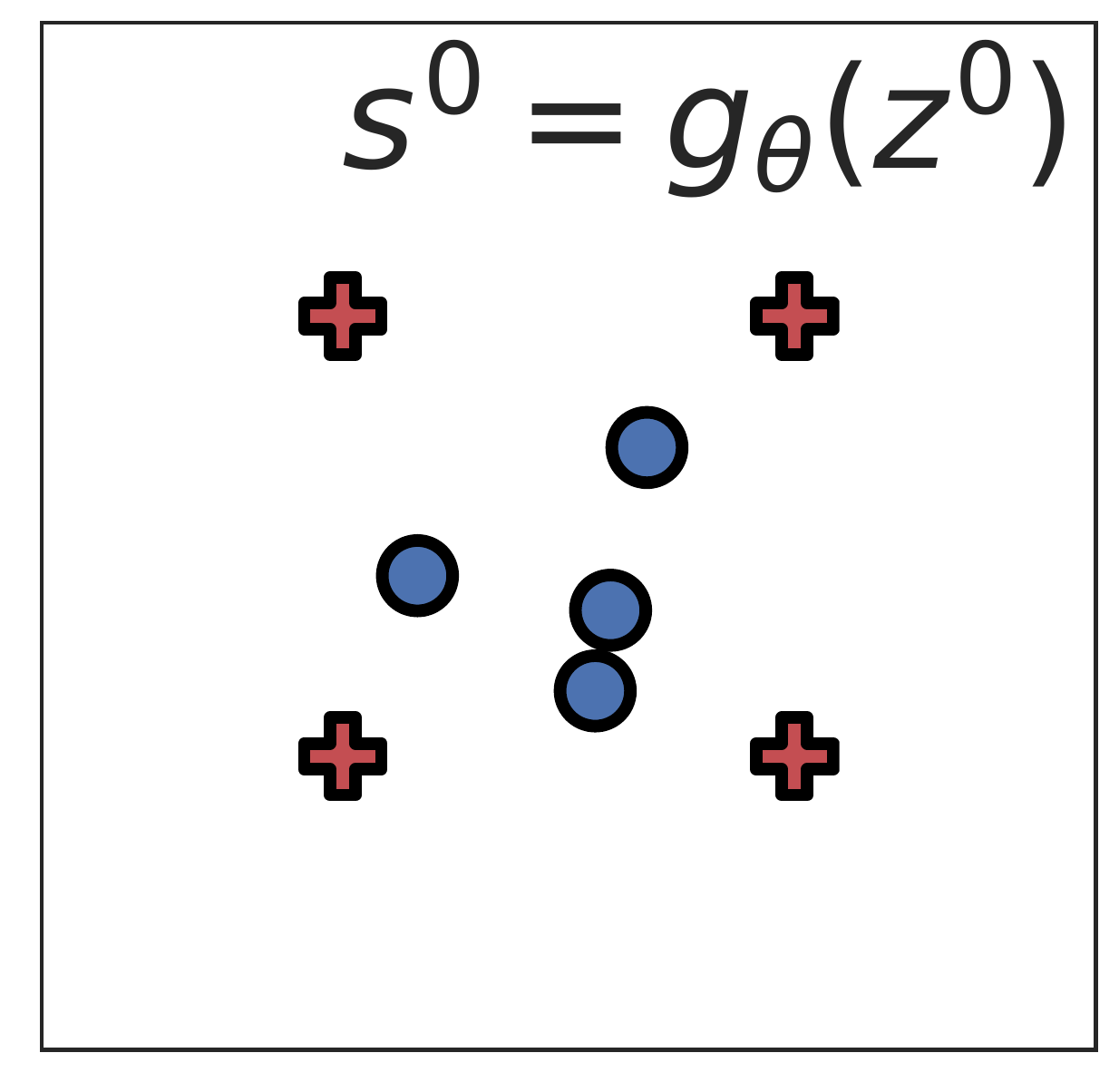}
\end{minipage}
\hspace{0.0cm}
\begin{minipage}[c]{.153\textwidth}
  \centering
  \includegraphics[width=0.85\textwidth]{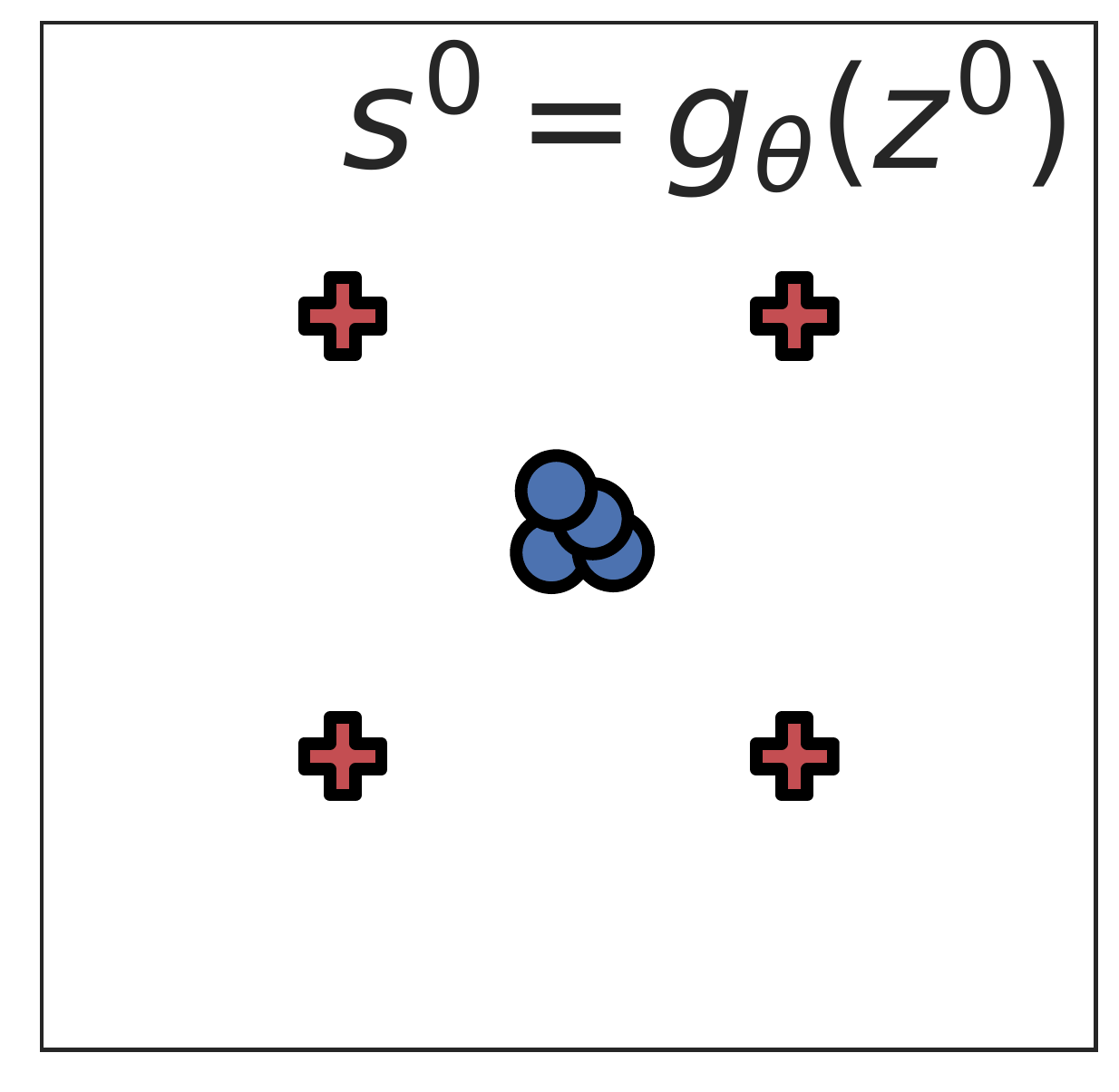}
\end{minipage}
\hspace{0.0cm}
\begin{minipage}[c]{.153\textwidth}
  \centering
  \includegraphics[width=0.85\textwidth]{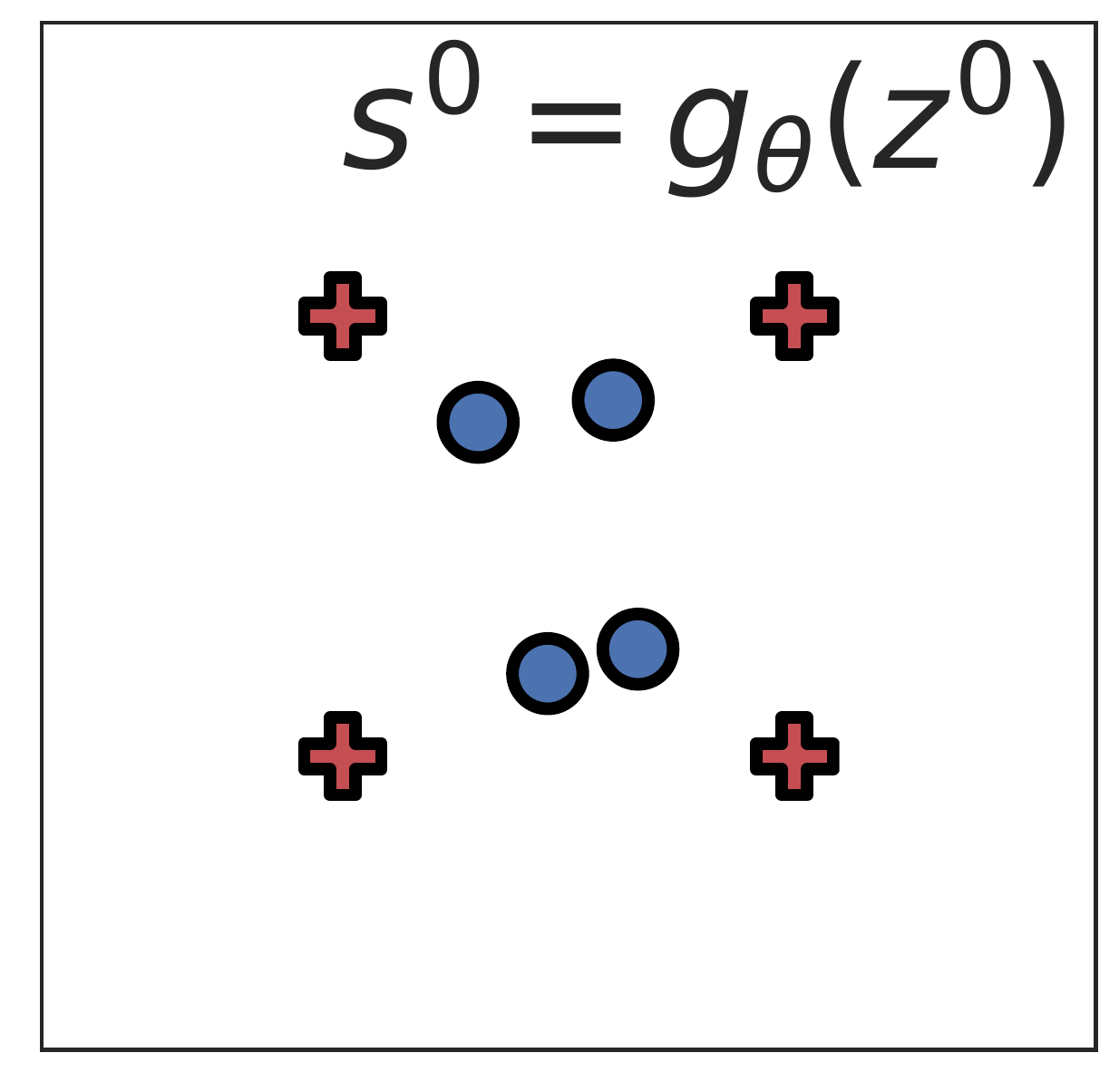}
\end{minipage}

\begin{minipage}[c]{.153\textwidth}
  \centering
  \includegraphics[width=0.85\textwidth]{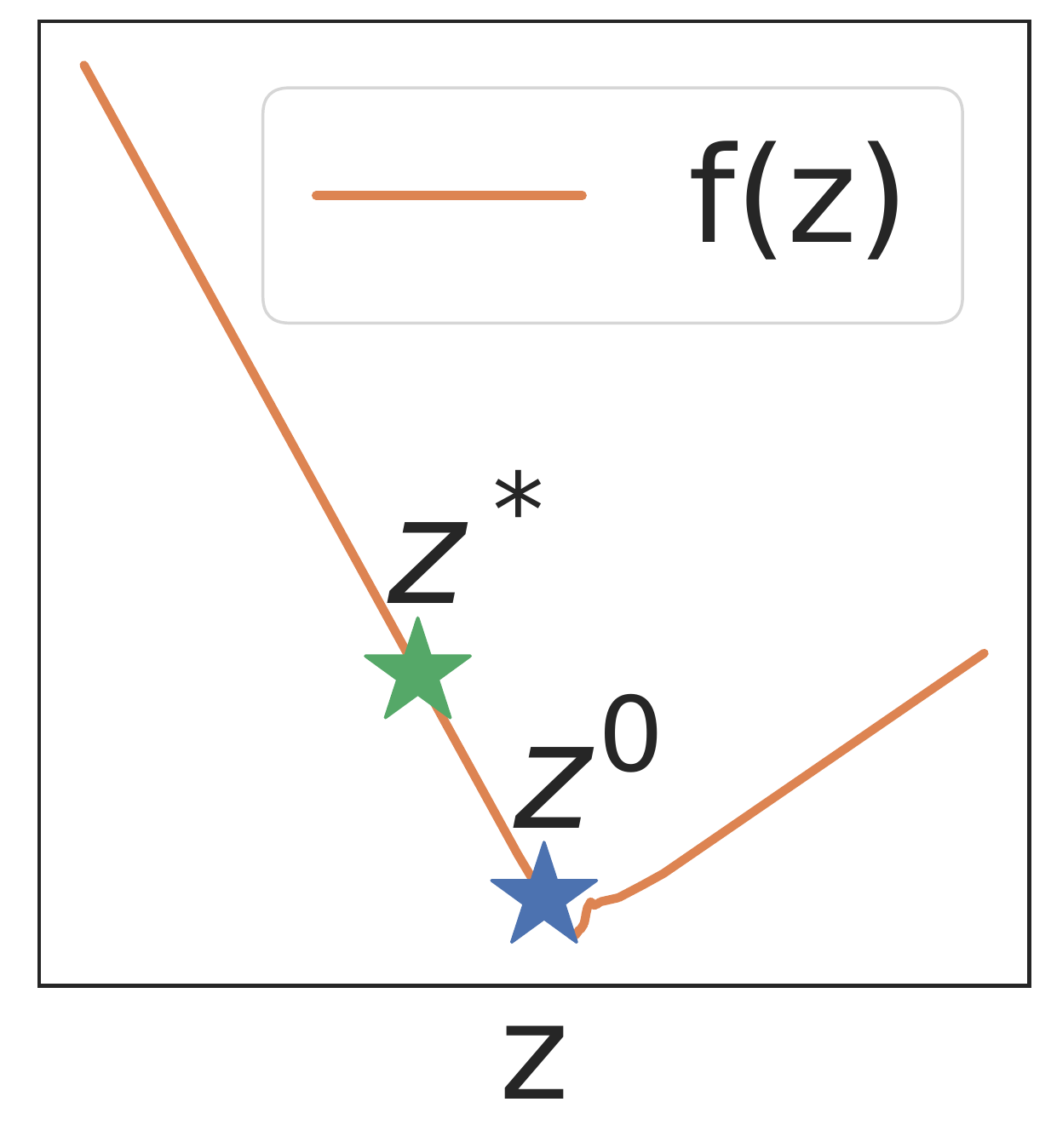}
\end{minipage}
\hspace{0.0cm}
\begin{minipage}[c]{.153\textwidth}
  \centering
  \includegraphics[width=0.85\textwidth]{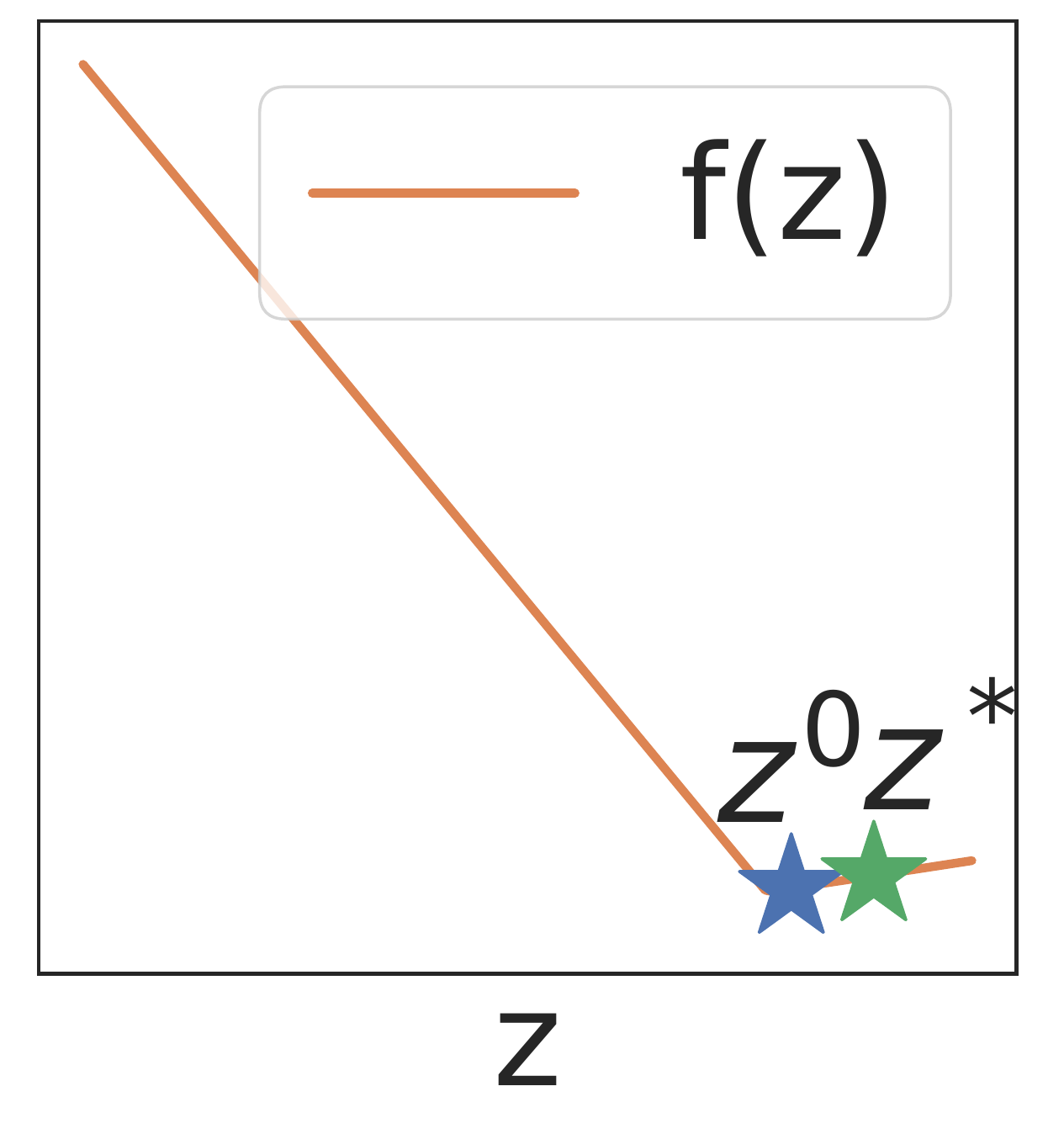}
\end{minipage}
\hspace{0.0cm}
\begin{minipage}[c]{.153\textwidth}
  \centering
  \includegraphics[width=0.85\textwidth]{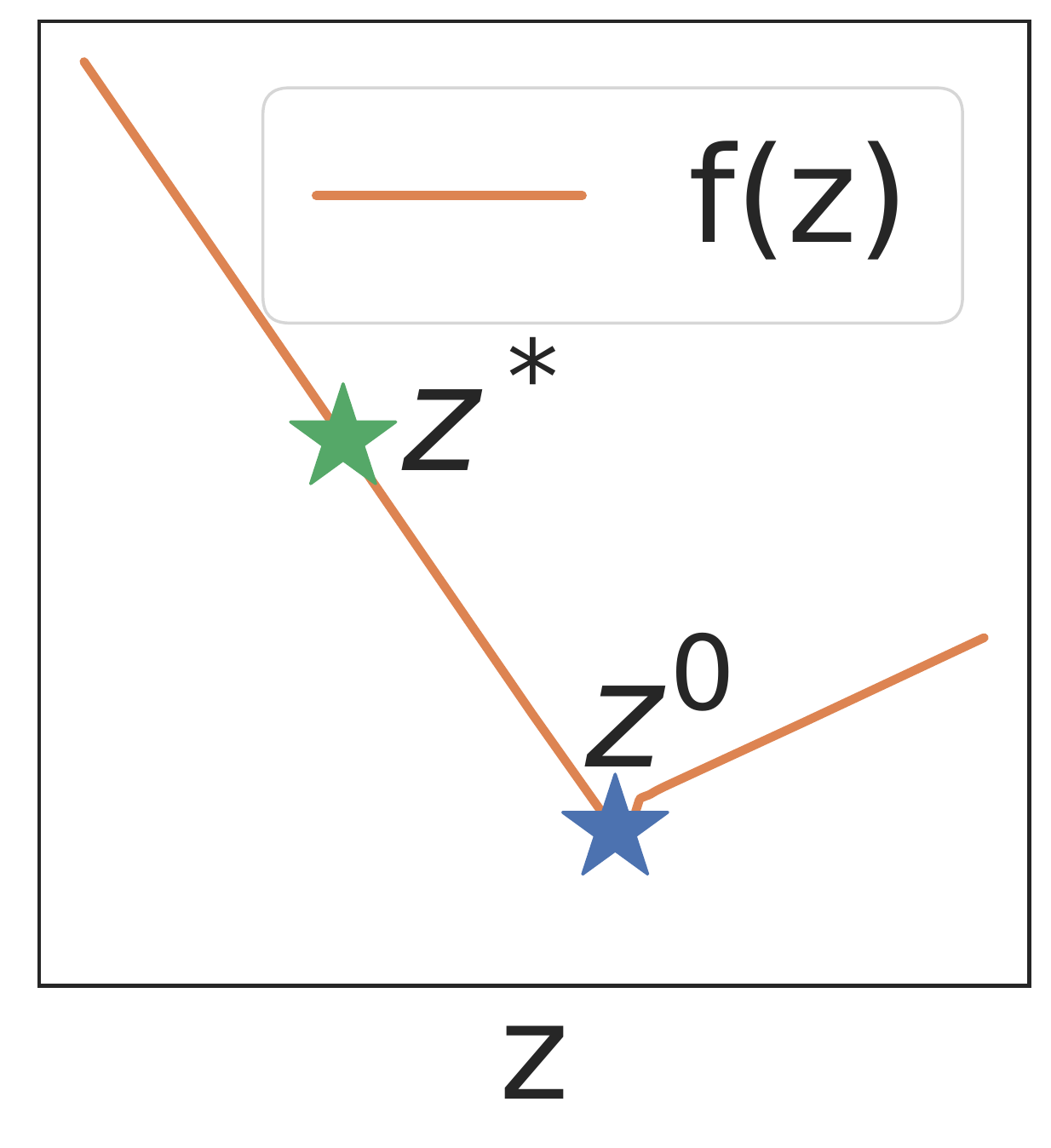}
\end{minipage}
\hspace{0.0cm}
\begin{minipage}[c]{.153\textwidth}
  \centering
  \includegraphics[width=0.85\textwidth]{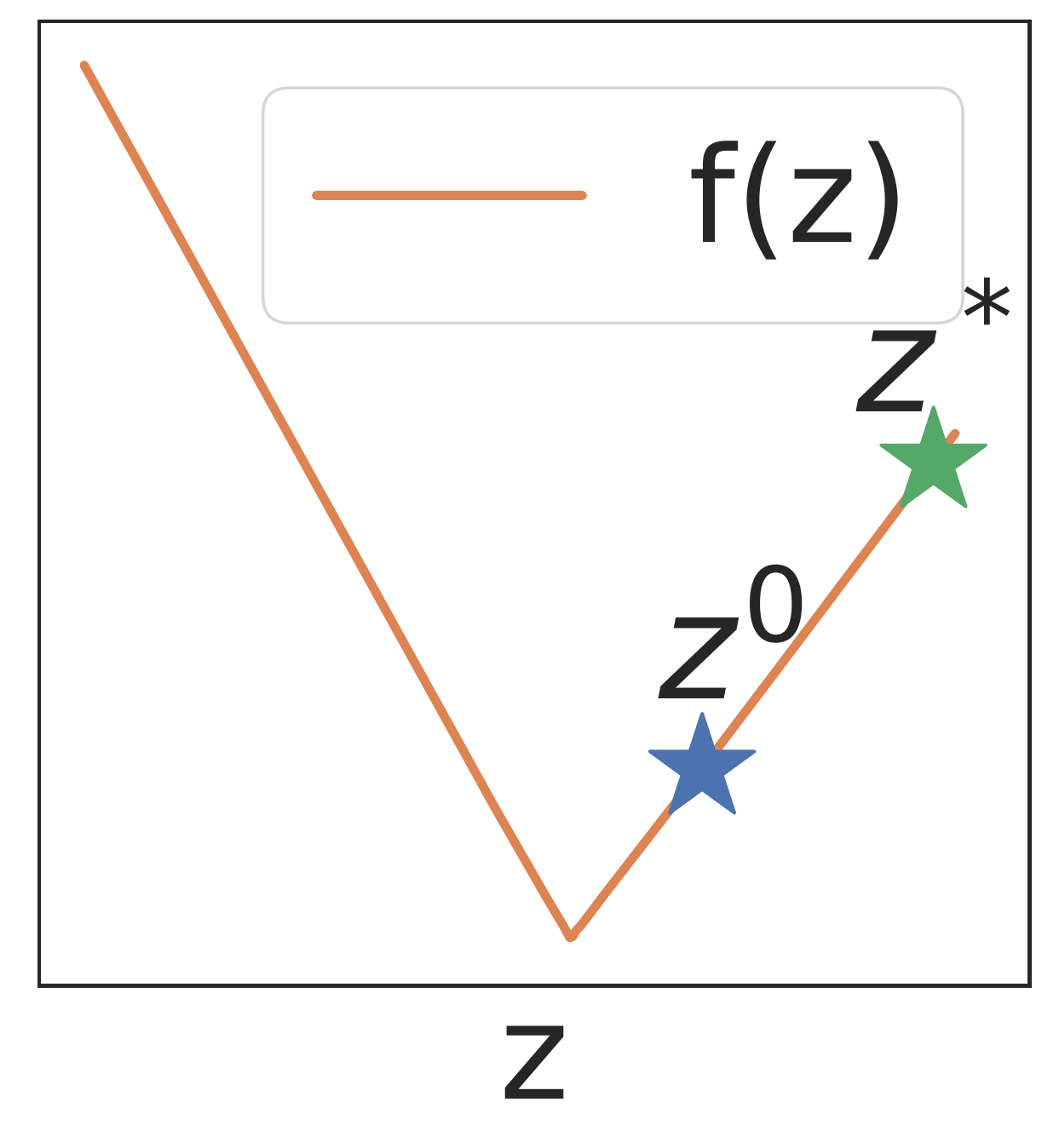}
\end{minipage}
\hspace{0.0cm}
\begin{minipage}[c]{.153\textwidth}
  \centering
  \includegraphics[width=0.85\textwidth]{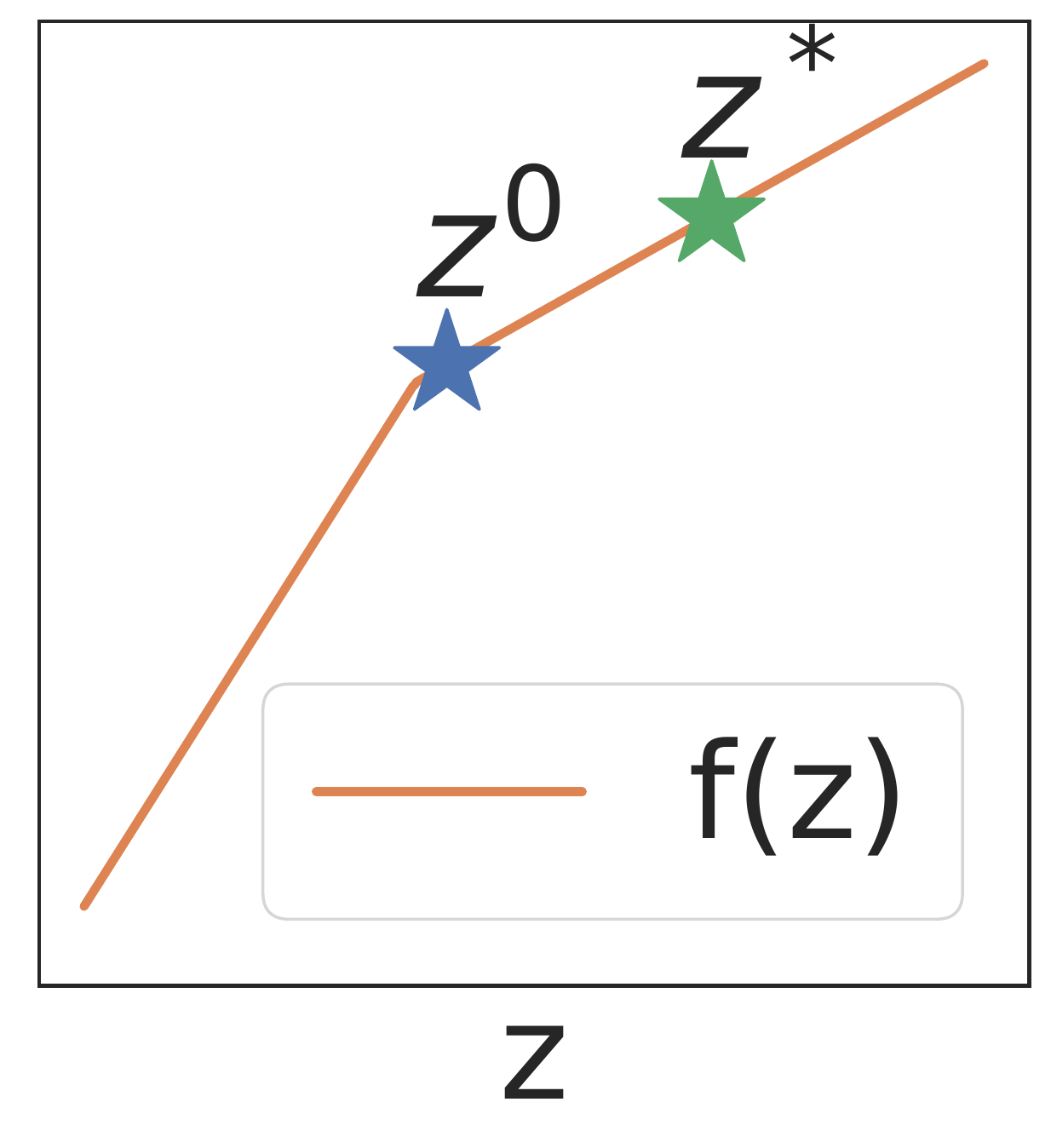}
\end{minipage}
\hspace{0.0cm}
\begin{minipage}[c]{.153\textwidth}
  \centering
  \includegraphics[width=0.85\textwidth]{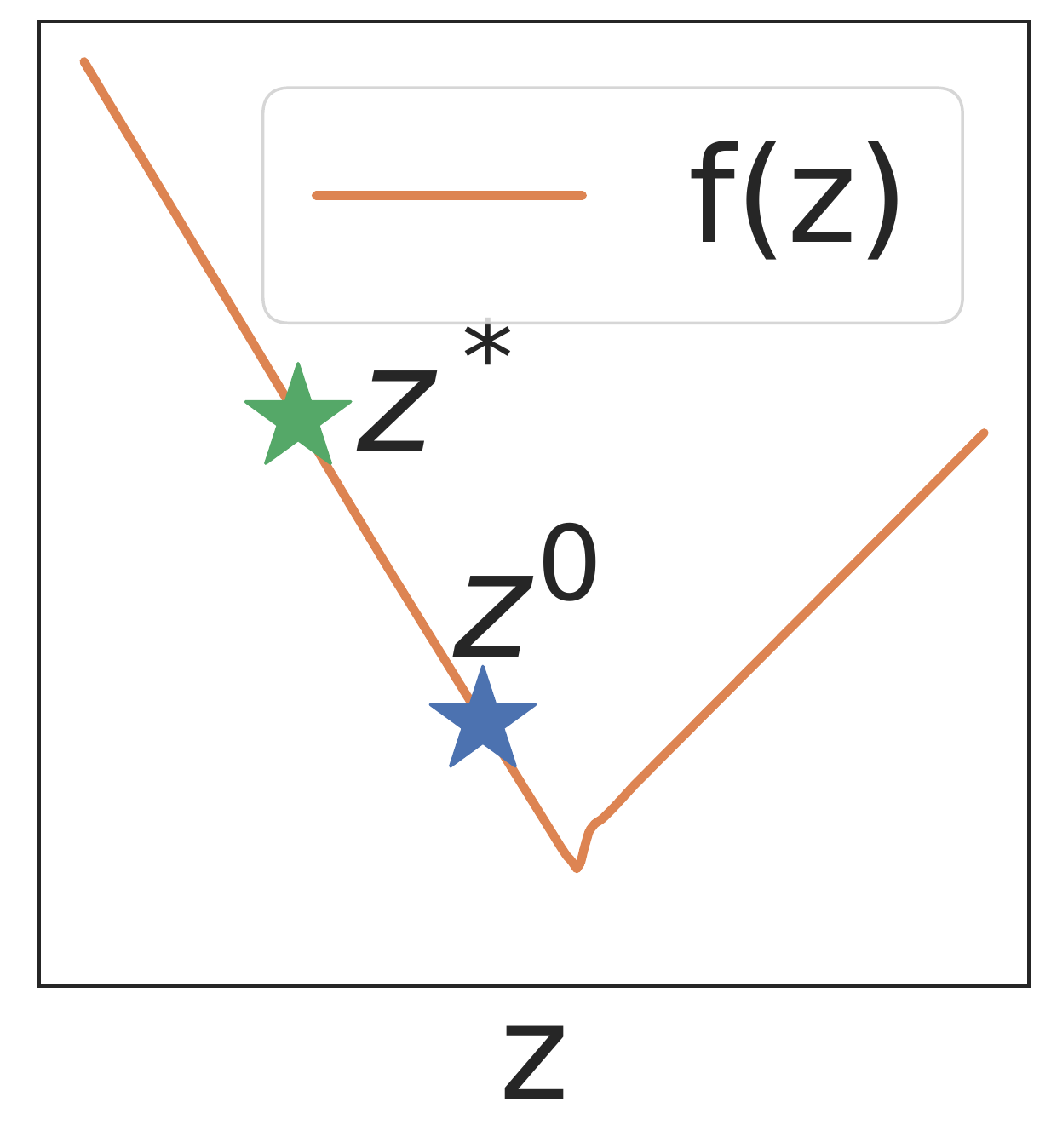}
\end{minipage}

\begin{minipage}[c]{.153\textwidth}
  \centering
  \includegraphics[width=0.85\textwidth]{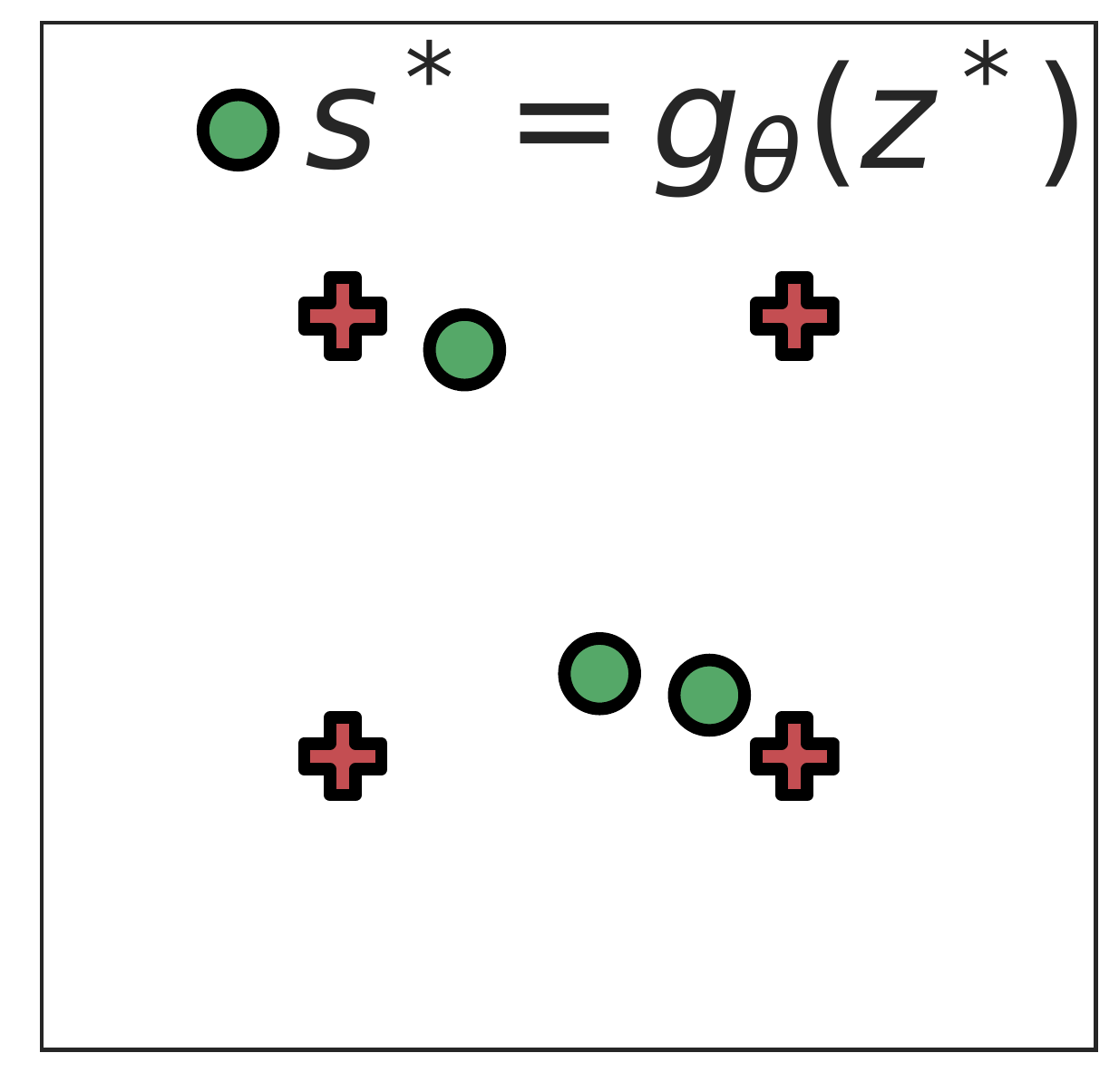}
\end{minipage}
\hspace{0.0cm}
\begin{minipage}[c]{.153\textwidth}
  \centering
  \includegraphics[width=0.85\textwidth]{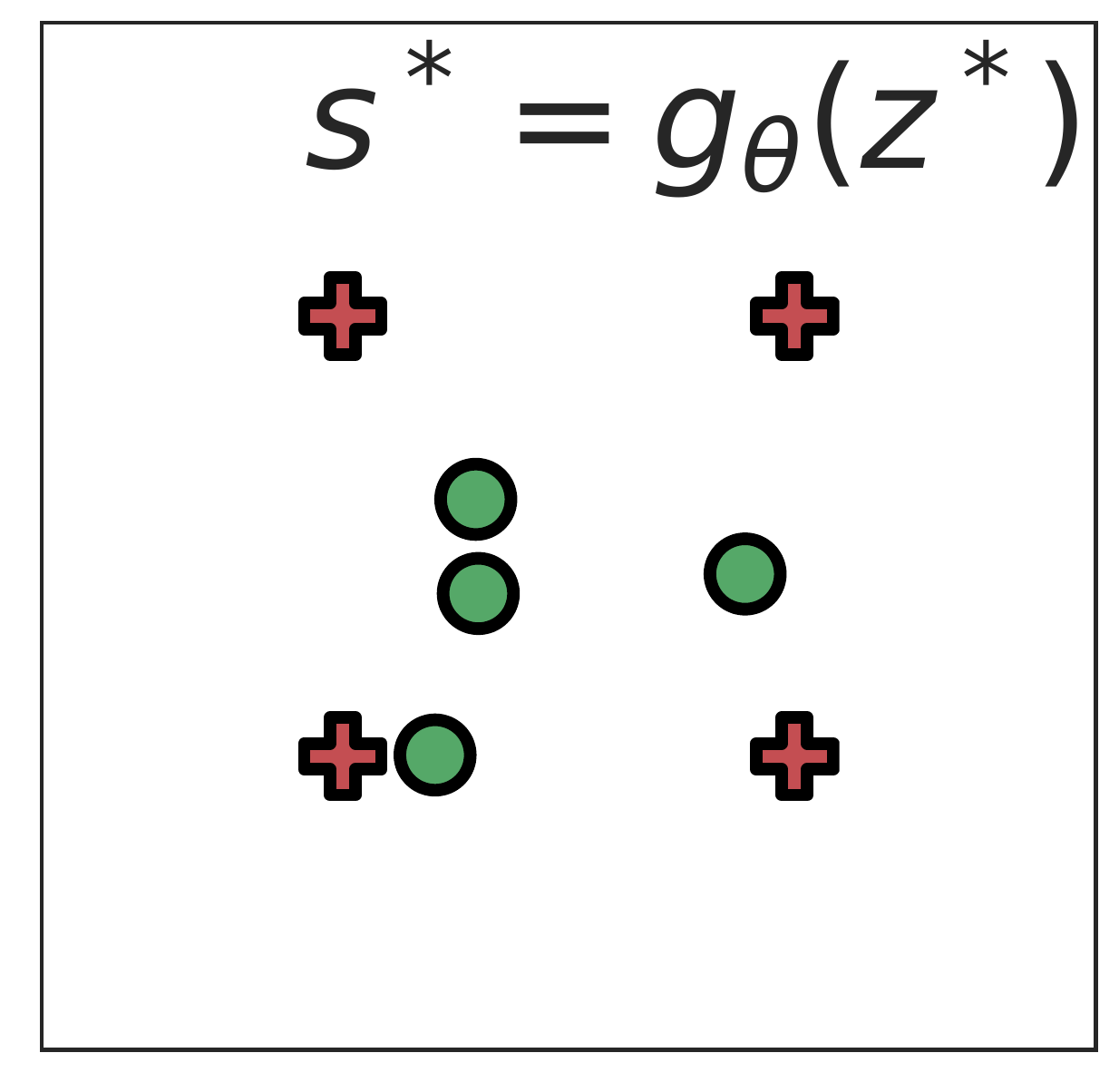}
\end{minipage}
\hspace{0.0cm}
\begin{minipage}[c]{.153\textwidth}
  \centering
  \includegraphics[width=0.85\textwidth]{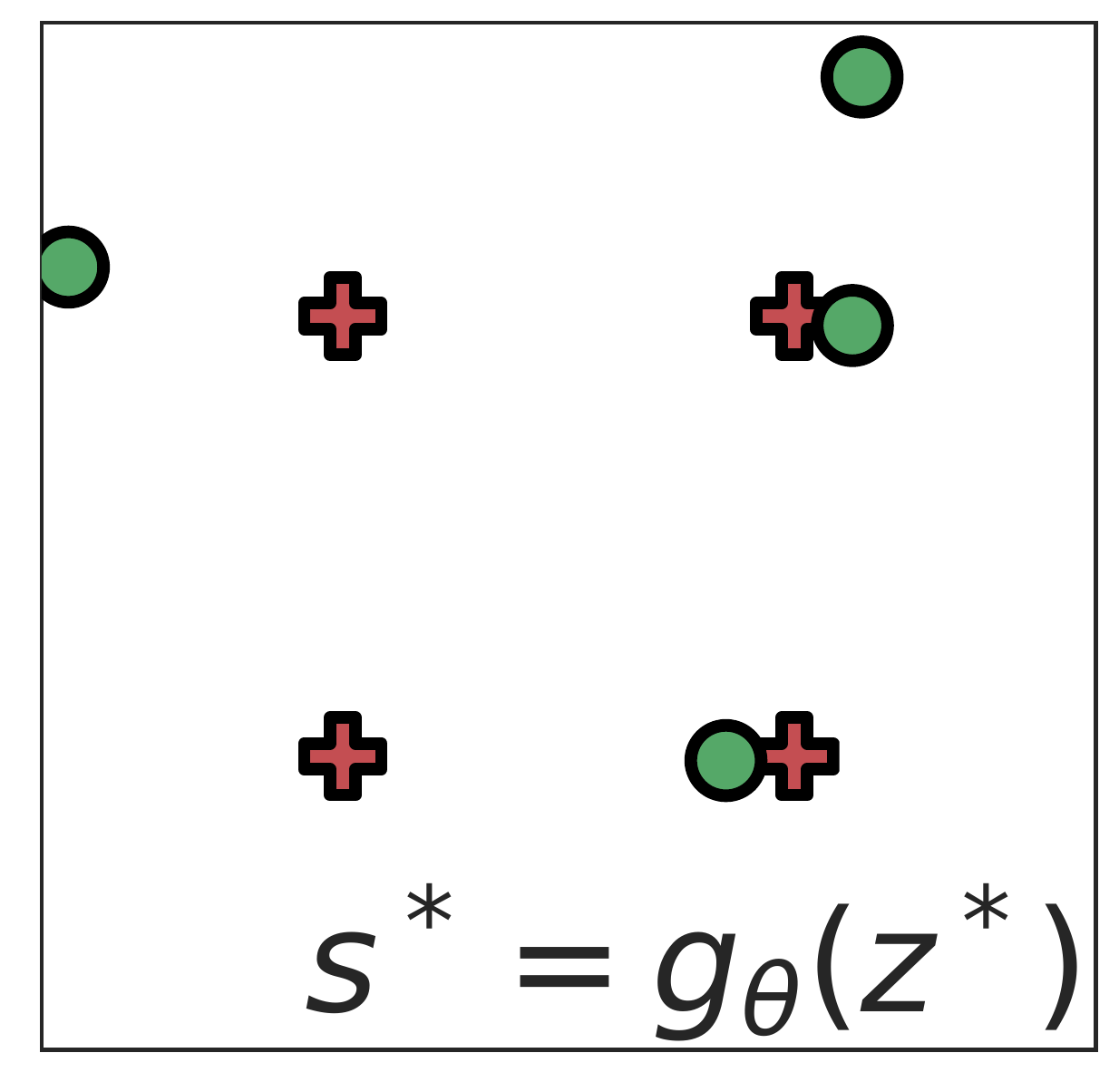}
\end{minipage}
\hspace{0.0cm}
\begin{minipage}[c]{.153\textwidth}
  \centering
  \includegraphics[width=0.85\textwidth]{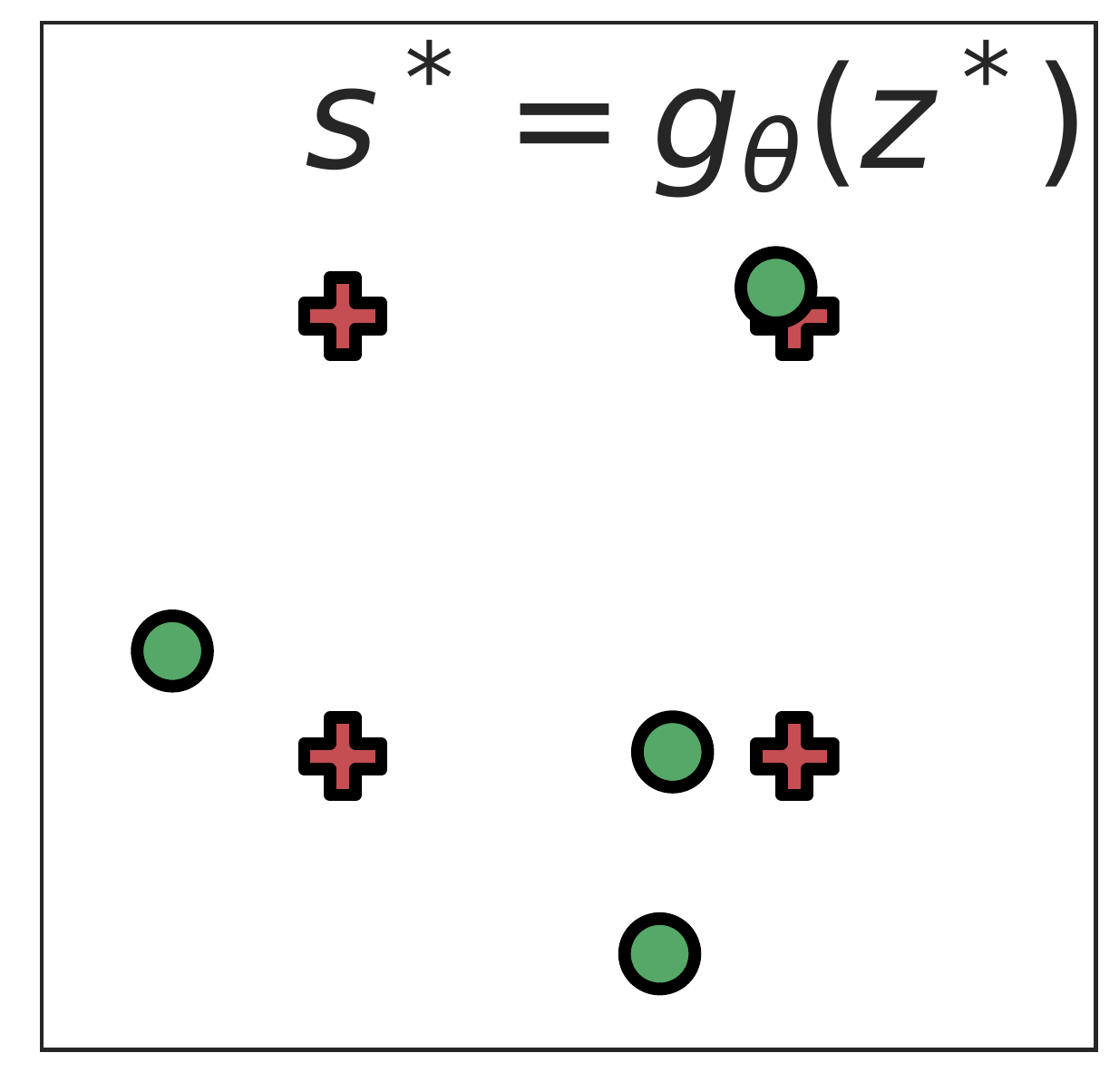}
\end{minipage}
\hspace{0.0cm}
\begin{minipage}[c]{.153\textwidth}
  \centering
  \includegraphics[width=0.85\textwidth]{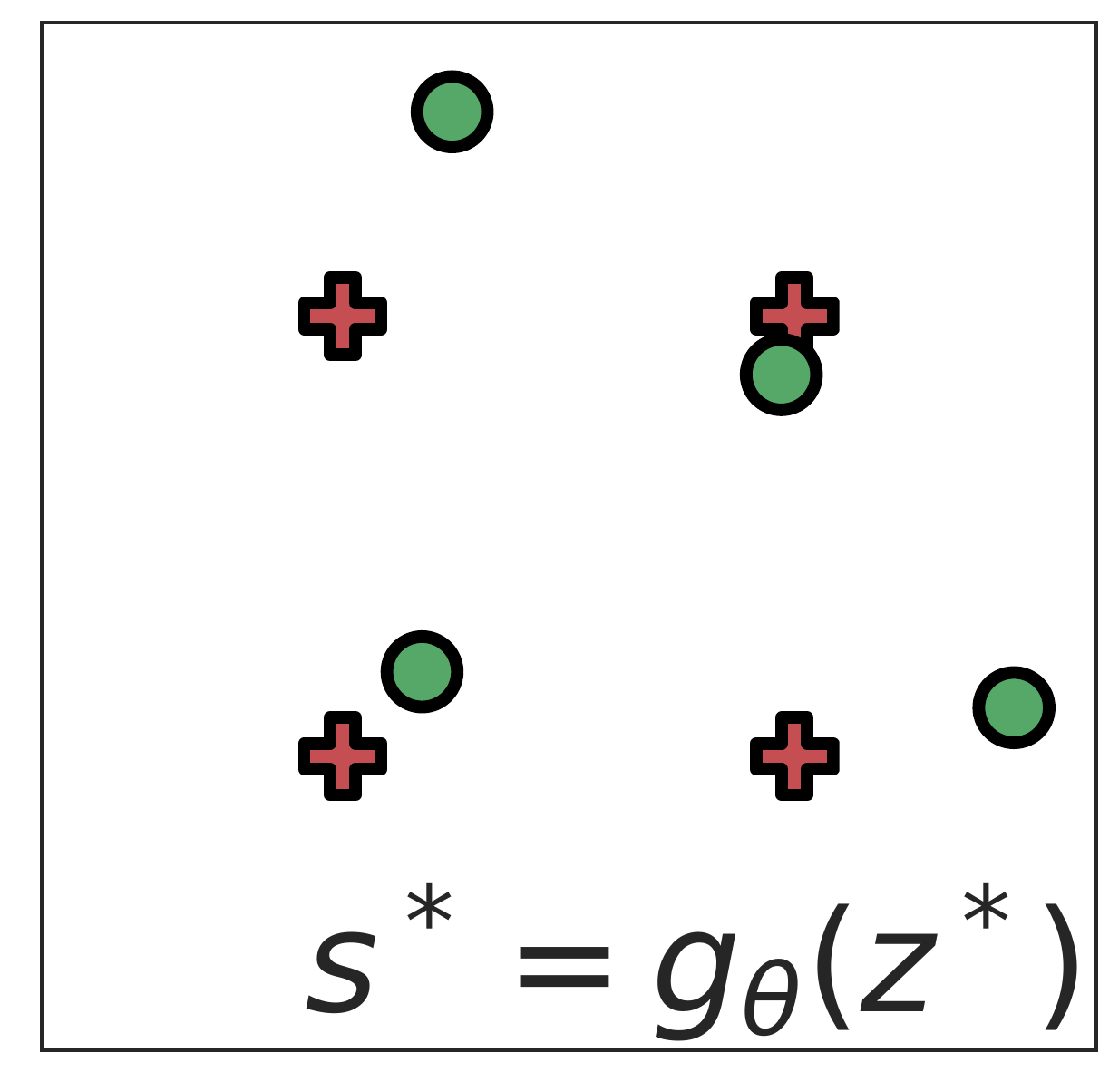}
\end{minipage}
\hspace{0.0cm}
\begin{minipage}[c]{.153\textwidth}
  \centering
  \includegraphics[width=0.85\textwidth]{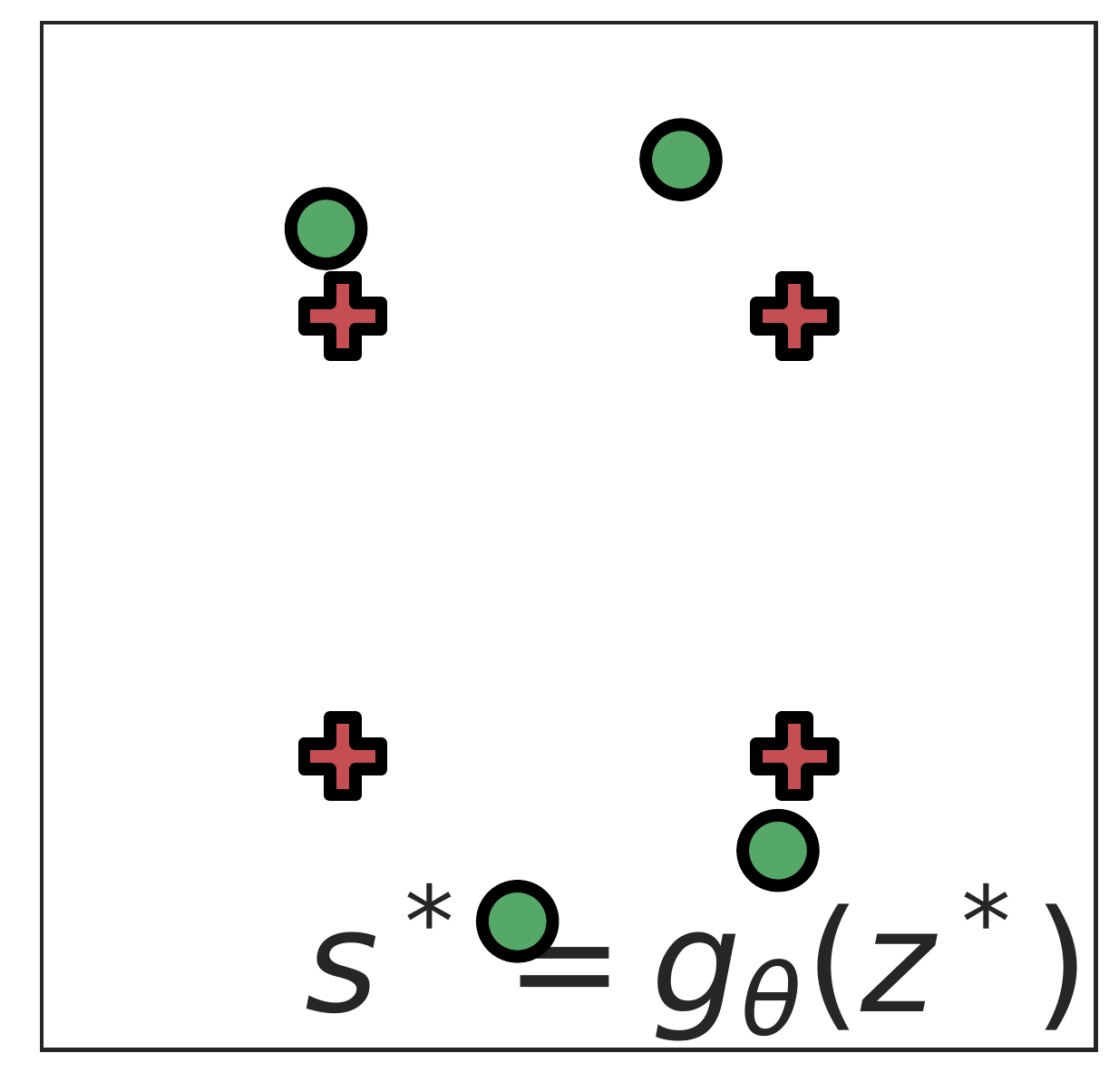}
\end{minipage}

\begin{minipage}[c]{.99\textwidth}
  \centering
  \includegraphics[width=0.92\textwidth]{method_figure_0512_time.pdf}
\end{minipage}
\caption{Surrogate models and generated states in cooperative navigation with 4 agents.}
\label{fig:fig_coopnavi_1d}
\end{figure*}

\begin{table*}[t]
  \caption{Number of training episodes ($\times$10) according to types of exploration scores for the surrogate model of REMAX. $f_i(s)$ is the inverse of state visit count of agent $i$ and $m(s)=\frac{1}{N}\sum_{j=1}^{N}f_j(s)$.}
  \label{table:table_surrogate}
  \centering
    \begin{tabular}{cccc}
        \toprule
        \multicolumn{1}{c}{Naive reward} &  
        \multicolumn{1}{c}{Intrinsic reward: Burrowing}&
        \multicolumn{1}{c}{Intrinsic reward: Covering}&
        \multicolumn{1}{c}{Proposed}\\
        \multicolumn{1}{c}{$r$} &  
        \multicolumn{1}{c}{$\sum_{i=1}^{N}f_i(s)\pmb{1}[f_i(s)>m(s)]$}&
        \multicolumn{1}{c}{$\sum_{i=1}^{N}f_i(s)\pmb{1}[f_i(s)<m(s)]$}&
        \multicolumn{1}{c}{Equation \ref{eq:1}}\\
        \midrule
        471\scriptsize $\pm$ 193 & 392\scriptsize $\pm$ 159 & 
        338\scriptsize $\pm$ 126 & \textbf{161}\scriptsize $\pm$ 69\\
        \bottomrule
    \end{tabular}
\end{table*}

\subsection{Cooperative Navigation}
Cooperative navigation shown in Figure \ref{fig:fig_illustrations} (b) is a multi-agent environment where homogeneous agents (blue circles) are required to be positioned at landmarks (red crosses) at the four corners. There are as many landmarks as there are agents. Each agent obtains a reward +1 when every landmark is occupied by one agent, and 0 otherwise. Thus, all agents should be coordinated to occupy a distinct landmark and thus receive the reward. 

\subsubsection{Effectiveness of the REMAX Structure}
We conducted ablation experiments to evaluate the synergistic effects of VGAE and the surrogate model in REMAX, and the results are summarized in Table \ref{table:results_cn}. The table presents the number of training episodes required for training a policy to complete the task. In the table, GENE with GAT indicates GENE whose VAE module is replaced with VGAE with the GAT encoder. 
Compared with GENE, adding GAT to GENE reduces the number of training episodes, as indicated in the table. 
Because REMAX requires fewer episodes than the other methods, regardless of the number of agents, the use of both VGAE and the surrogate model is validated as being effective for training the MARL model, especially when the number of agents is large. 
In addition, the number of episodes for REMAX increases more slowly with the number of agents, which implies that REMAX enables MARL to train the policy scalably even when the number of agents increases. Meanwhile, HER and RCG require a rapidly increasing number of episodes for training the policy with an increasing number of agents, and can no longer train the policy when the number of agents exceeds 6.

\subsubsection{Analysis of Representing/Generating States}
Figure \ref{fig:fig_coopnavi_attention} shows the normalized attention coefficients of GAT in the VGAE encoder in REMAX; the thicker the line is, the larger the coefficient is. The coefficients between agents increase as multiple agents with their actions (blue arrows) approach a common landmark. This trend can be observed in the second figure of Figure \ref{fig:fig_coopnavi_attention}, which shows the two agents approaching the upper-right landmark, and the fifth figure of Figure \ref{fig:fig_coopnavi_attention}, which shows the two agents approaching the lower-right landmark. 

Figure \ref{fig:fig_coopnavi_1d} compares the two types of states: $s^0$ decoded from an initial latent vector $z^0$ and $s^*$ decoded from an optimized latent vector $z^*$. The second-row figures show $z^0$ and $z^*$ over $f_\psi(z)$. In addition, the first-and third-row figures show the decoded states, $s^0$ and $s^*$, from $z^0$ and $z^*$ using $g_\theta$. As MARL and REMAX are trained with more samples, as shown in the figure, REMAX tends to generate states near the landmarks, enabling the agents to easily complete the task.



\setcounter{figure}{6}
\begin{figure*}[t]
\begin{minipage}[c]{.153\textwidth}
  \centering
  \includegraphics[width=0.85\textwidth]{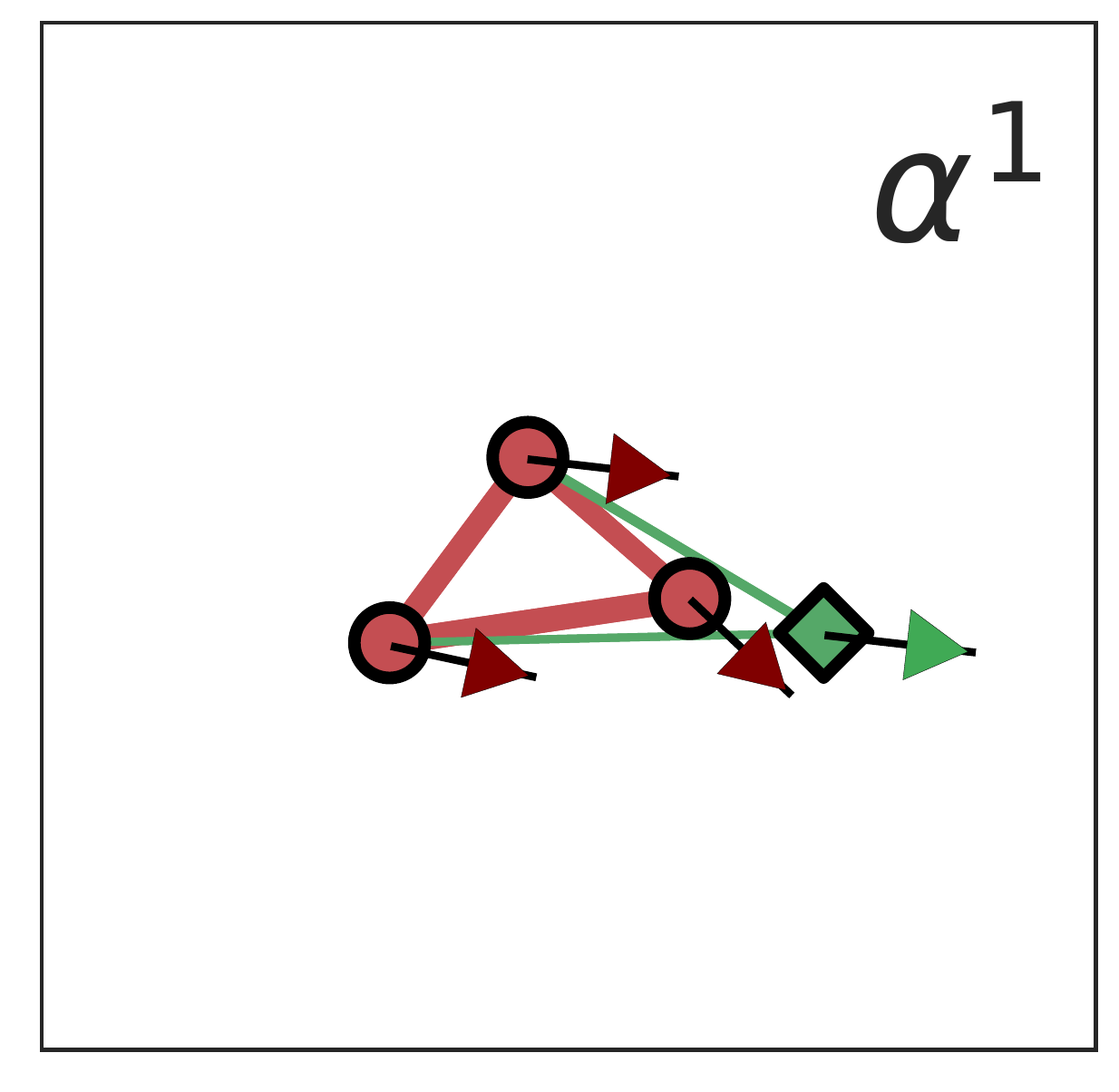}
\end{minipage}
\hspace{0.0cm}
\begin{minipage}[c]{.153\textwidth}
  \centering
  \includegraphics[width=0.85\textwidth]{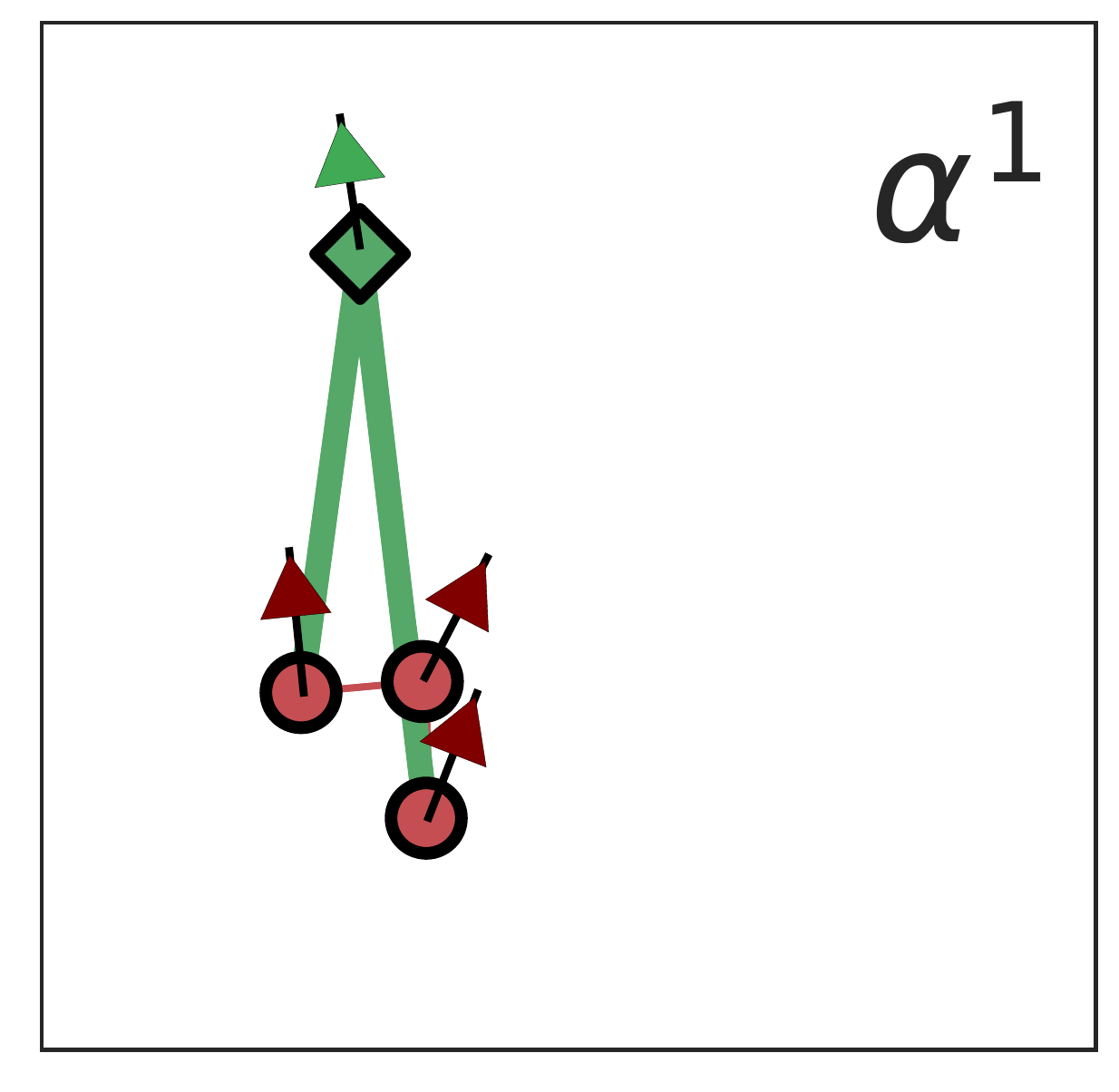}
\end{minipage}
\hspace{0.0cm}
\begin{minipage}[c]{.153\textwidth}
  \centering
  \includegraphics[width=0.85\textwidth]{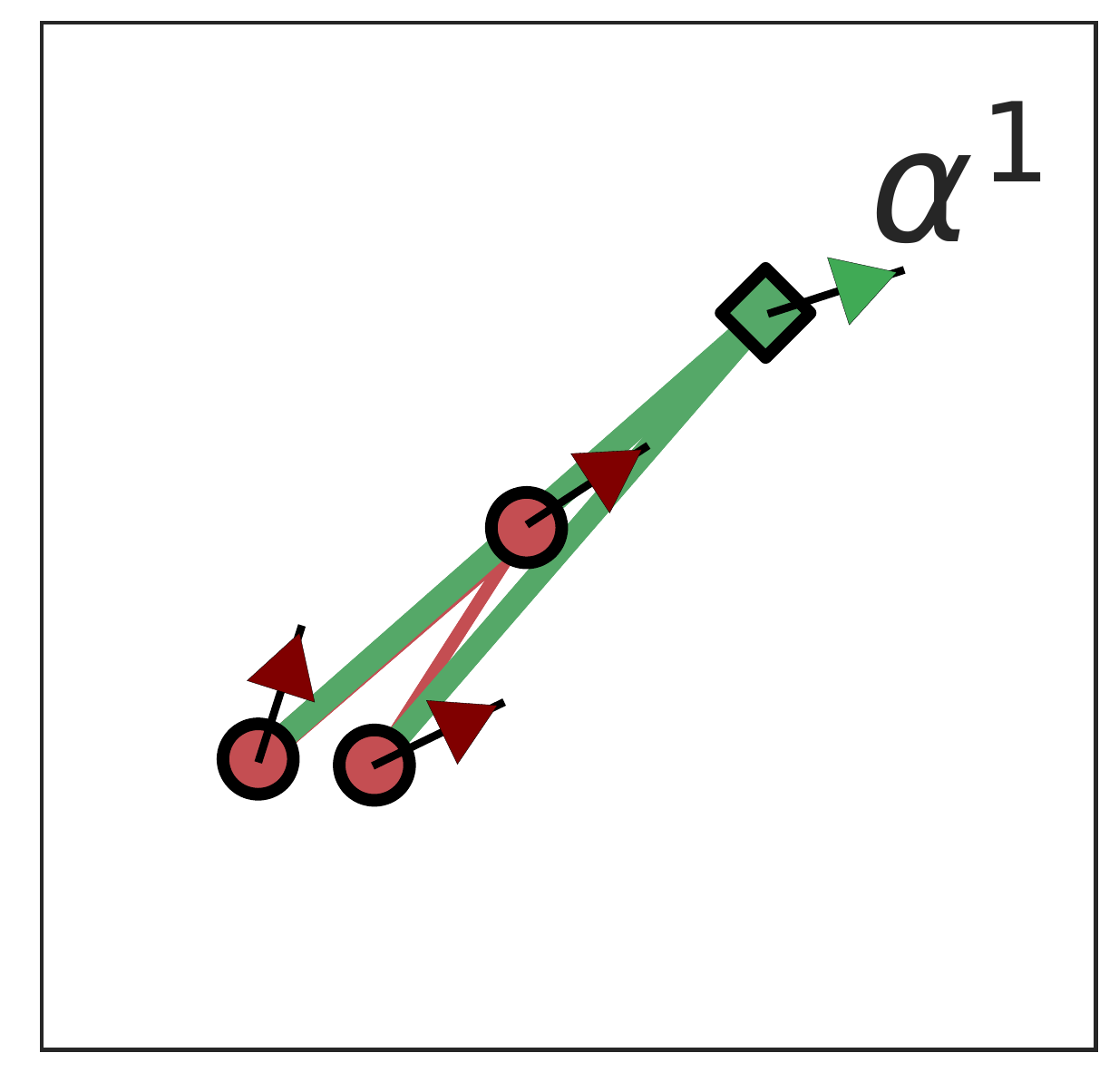}
\end{minipage}
\hspace{0.0cm}
\begin{minipage}[c]{.153\textwidth}
  \centering
  \includegraphics[width=0.85\textwidth]{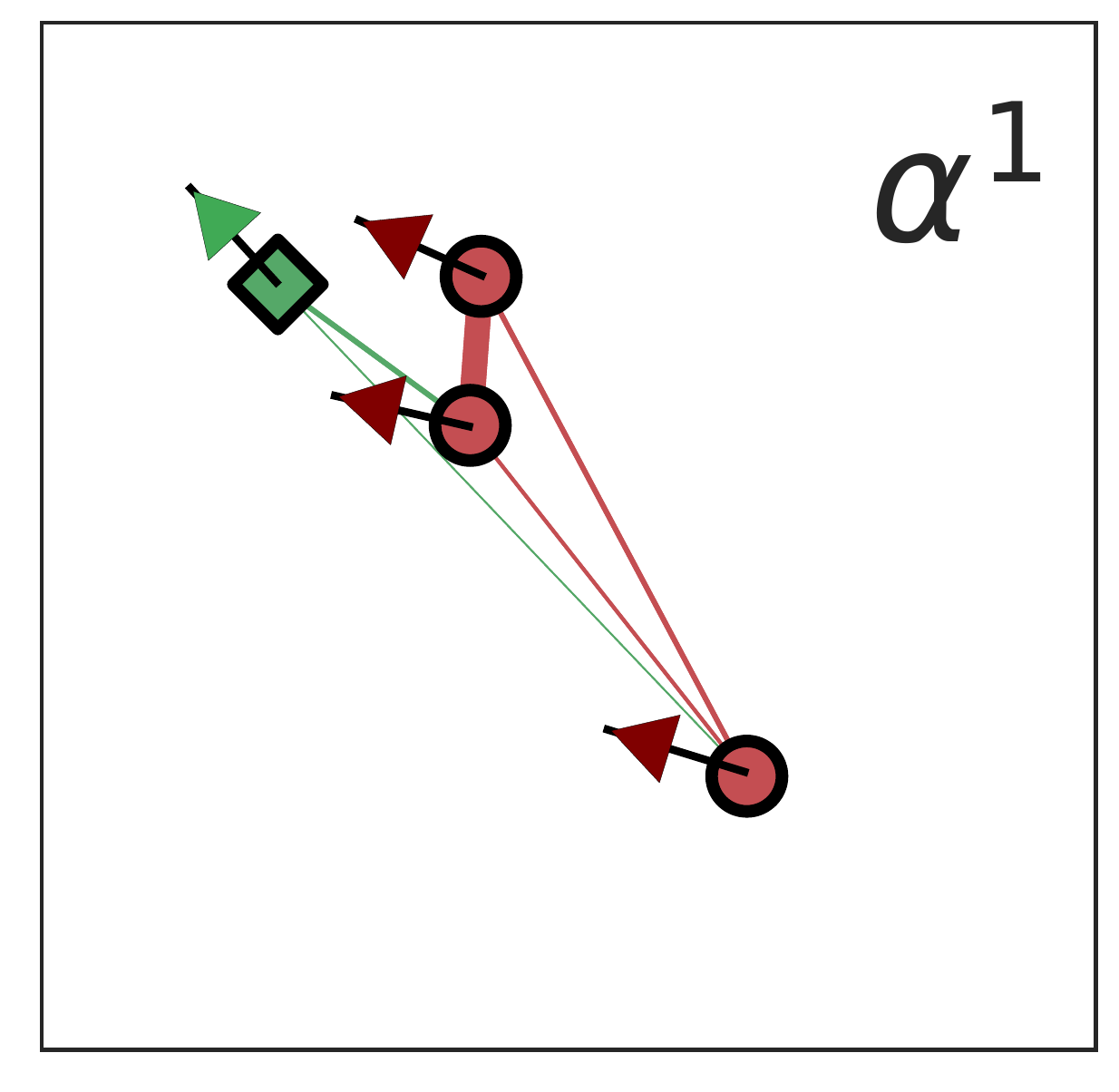}
\end{minipage}
\hspace{0.0cm}
\begin{minipage}[c]{.153\textwidth}
  \centering
  \includegraphics[width=0.85\textwidth]{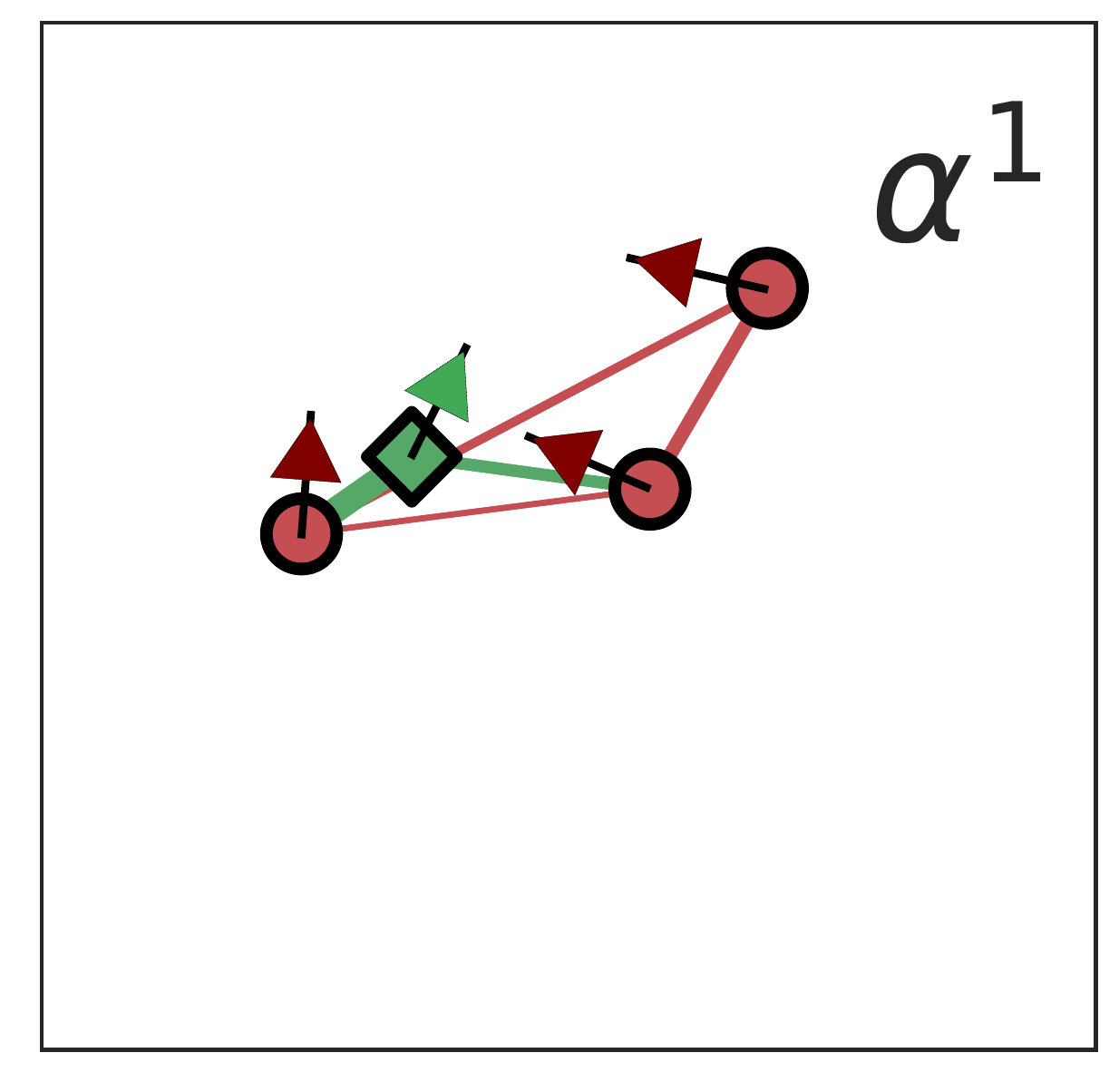}
\end{minipage}
\hspace{0.0cm}
\begin{minipage}[c]{.153\textwidth}
  \centering
  \includegraphics[width=0.85\textwidth]{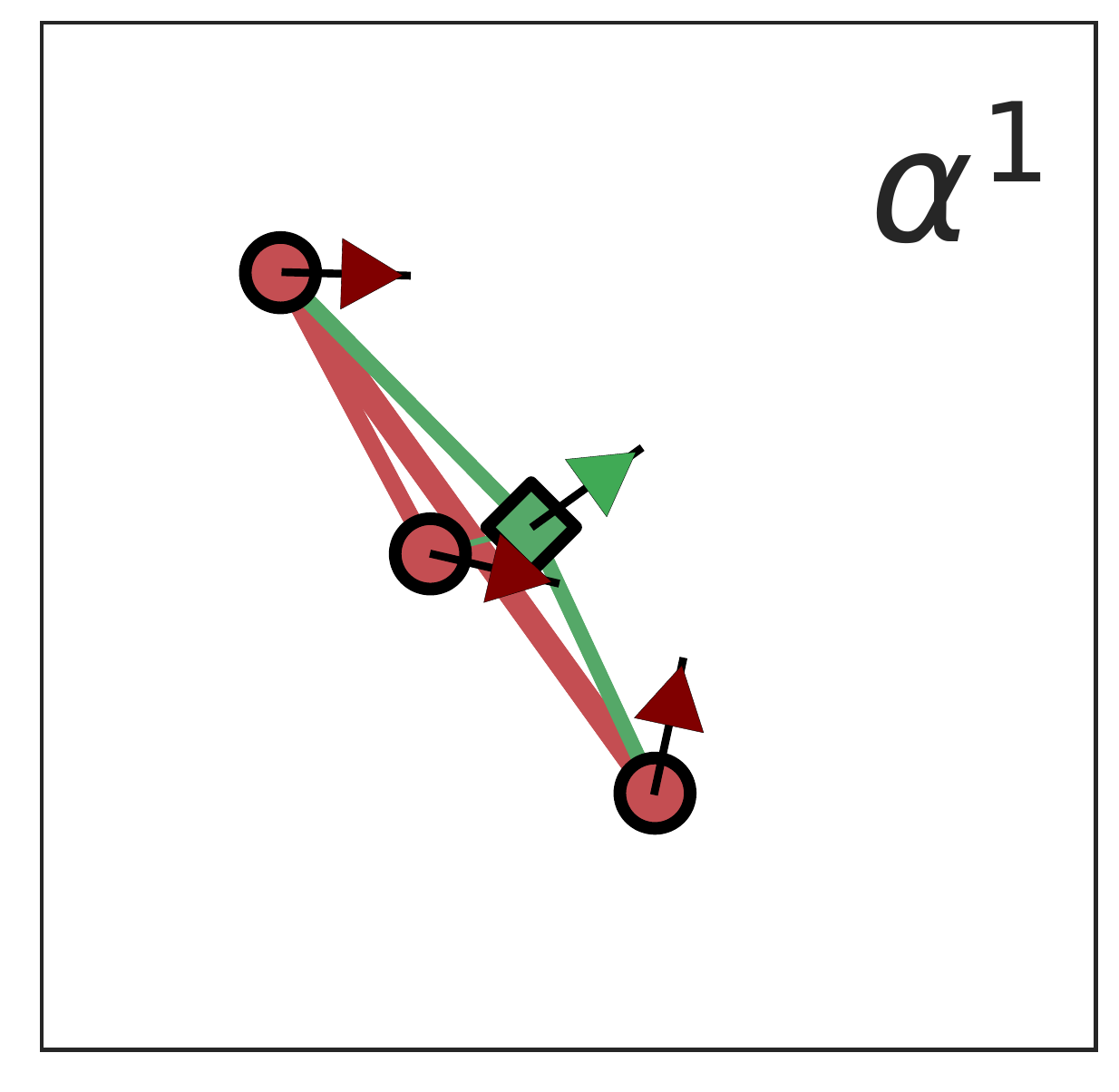}
\end{minipage}

\begin{minipage}[c]{.153\textwidth}
  \centering
  \includegraphics[width=0.85\textwidth]{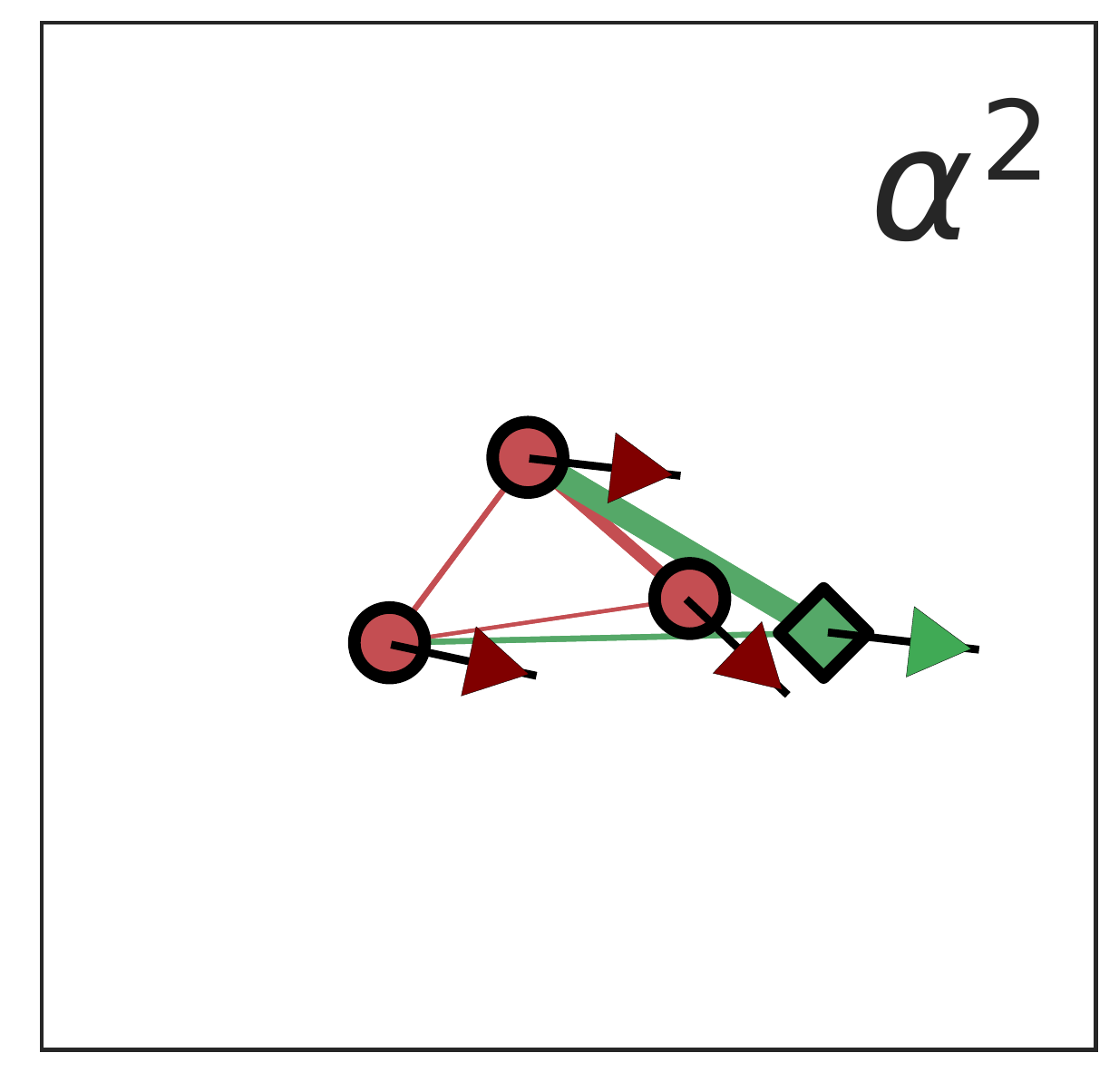}
\end{minipage}
\hspace{0.0cm}
\begin{minipage}[c]{.153\textwidth}
  \centering
  \includegraphics[width=0.85\textwidth]{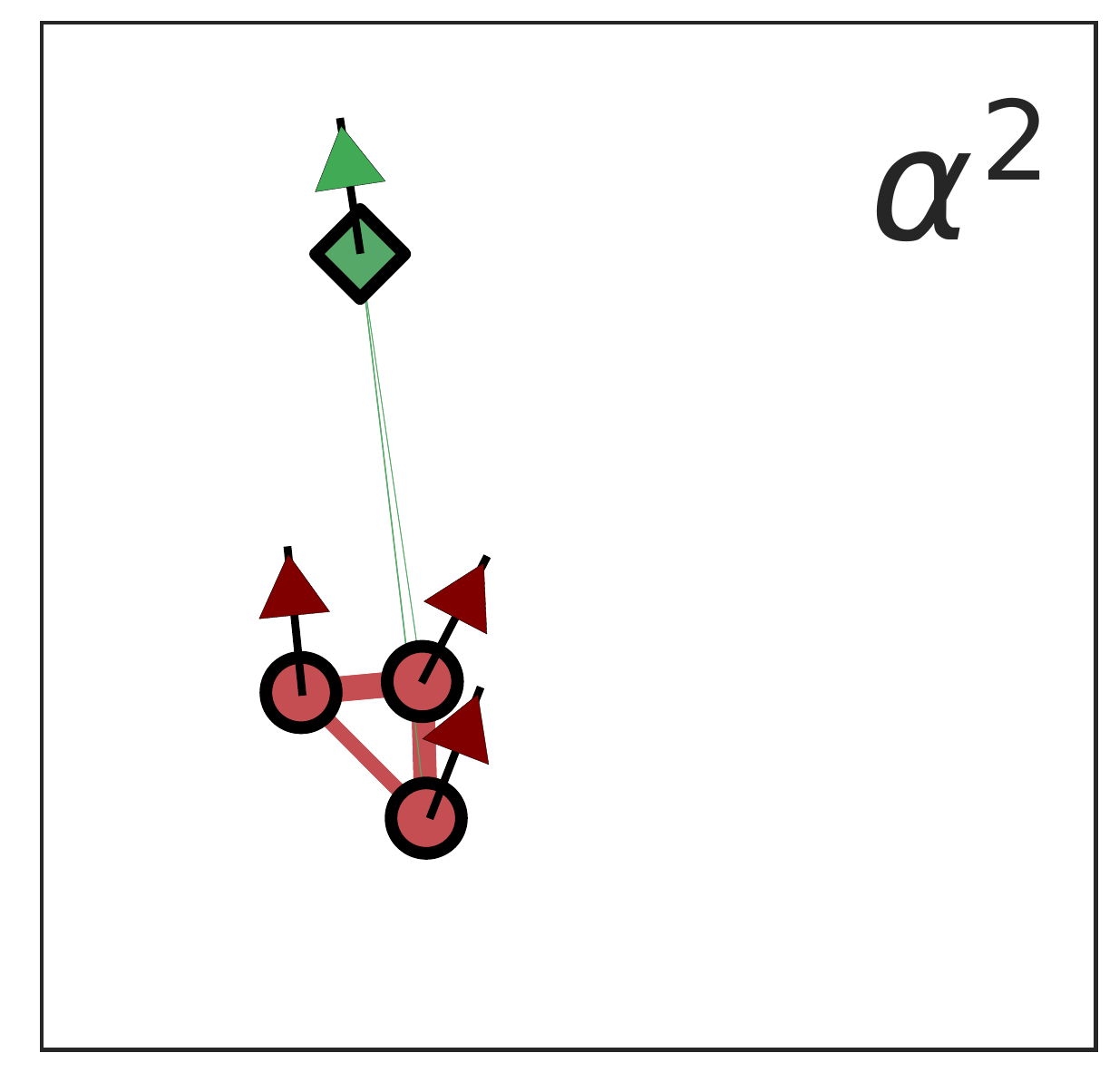}
\end{minipage}
\hspace{0.0cm}
\begin{minipage}[c]{.153\textwidth}
  \centering
  \includegraphics[width=0.85\textwidth]{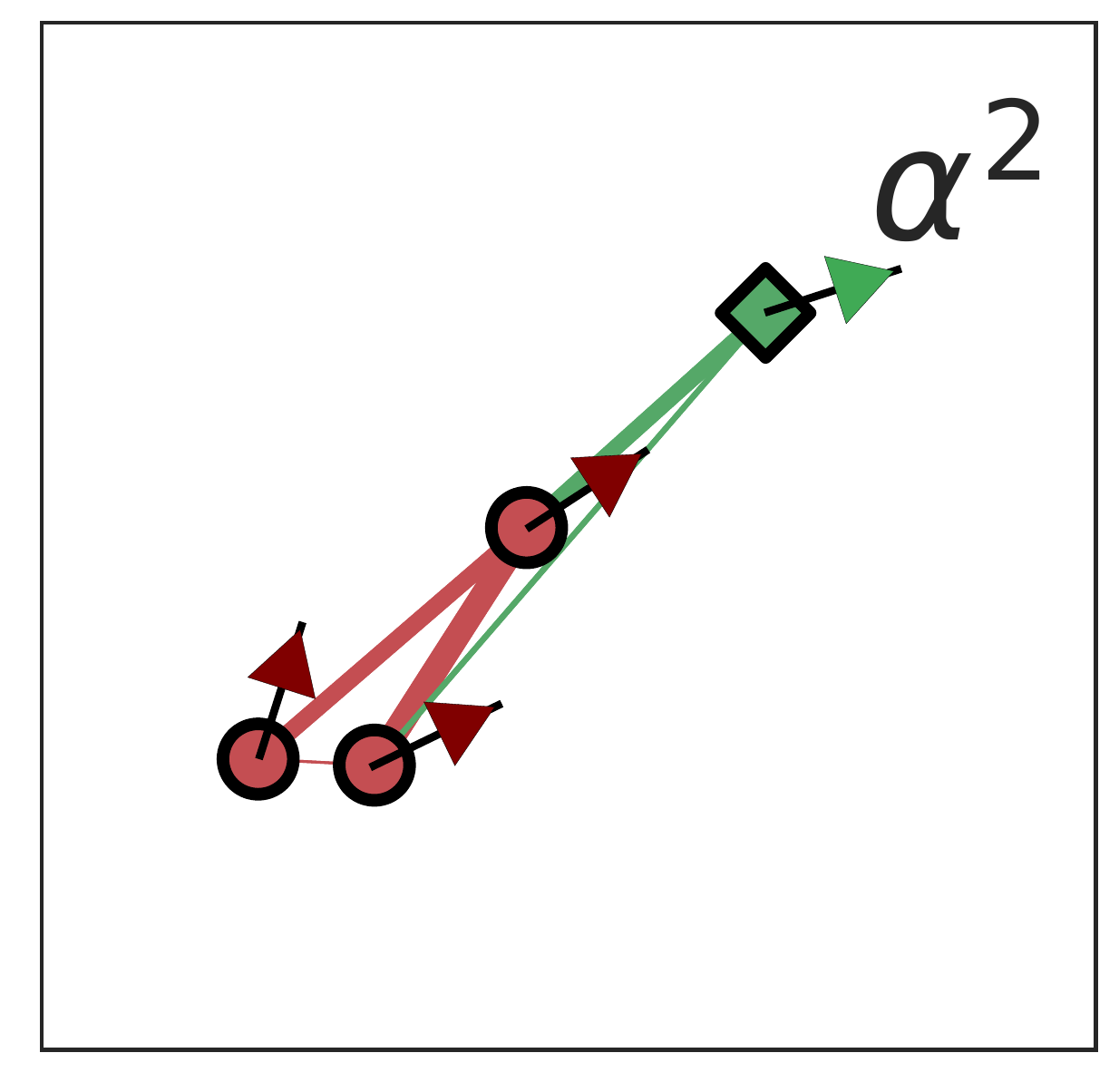}
\end{minipage}
\hspace{0.0cm}
\begin{minipage}[c]{.153\textwidth}
  \centering
  \includegraphics[width=0.85\textwidth]{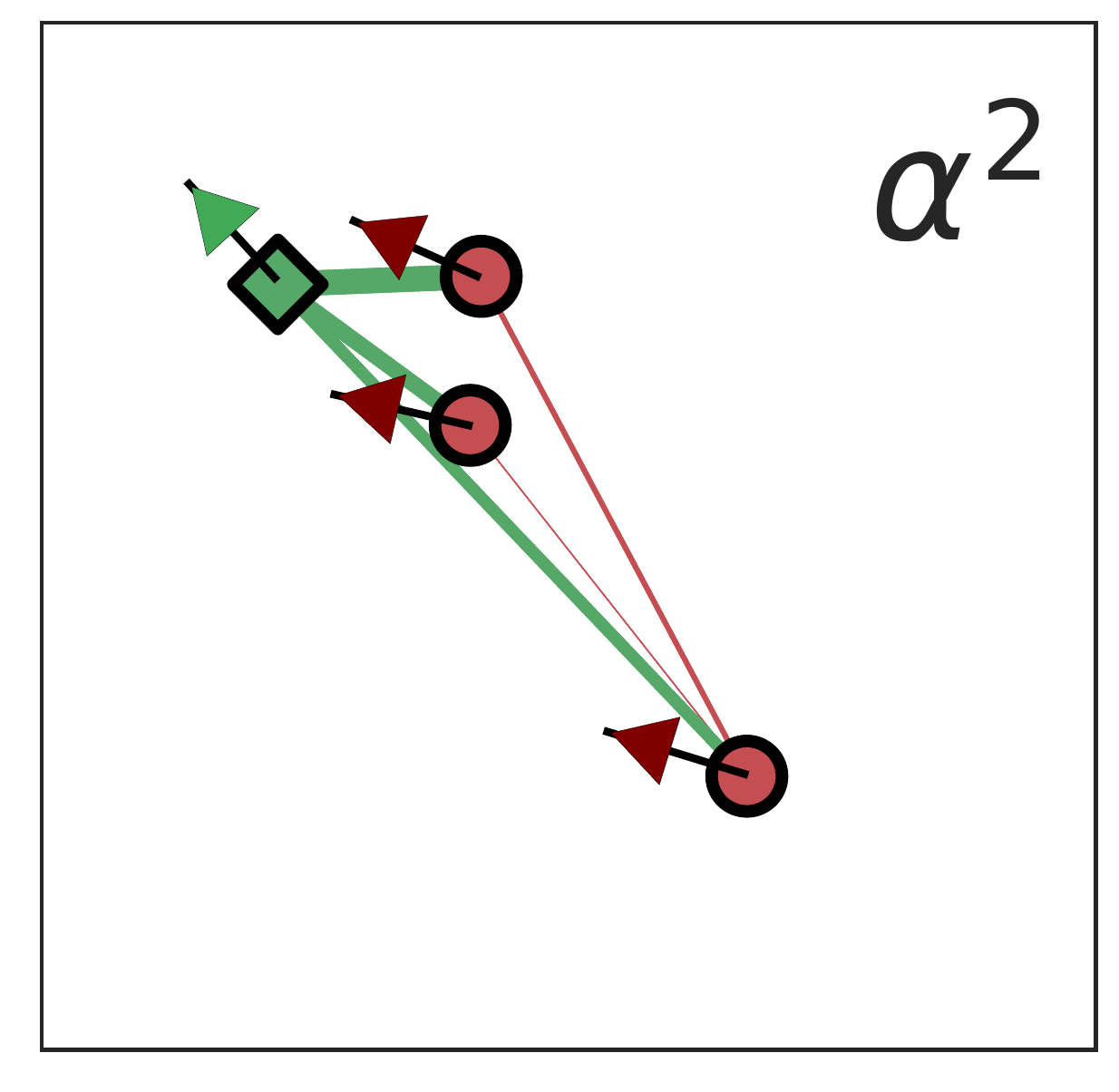}
\end{minipage}
\hspace{0.0cm}
\begin{minipage}[c]{.153\textwidth}
  \centering
  \includegraphics[width=0.85\textwidth]{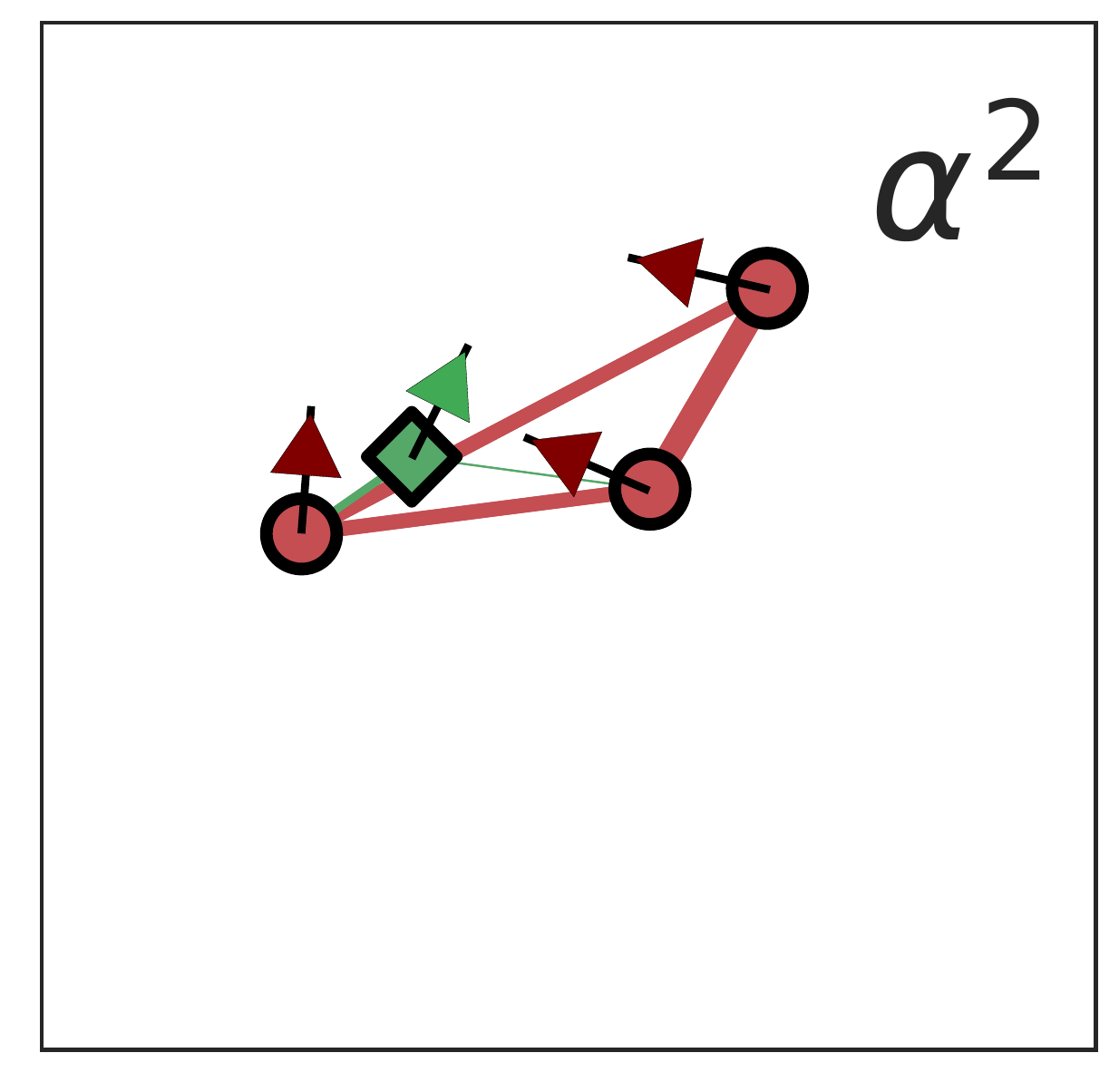}
\end{minipage}
\hspace{0.0cm}
\begin{minipage}[c]{.153\textwidth}
  \centering
  \includegraphics[width=0.85\textwidth]{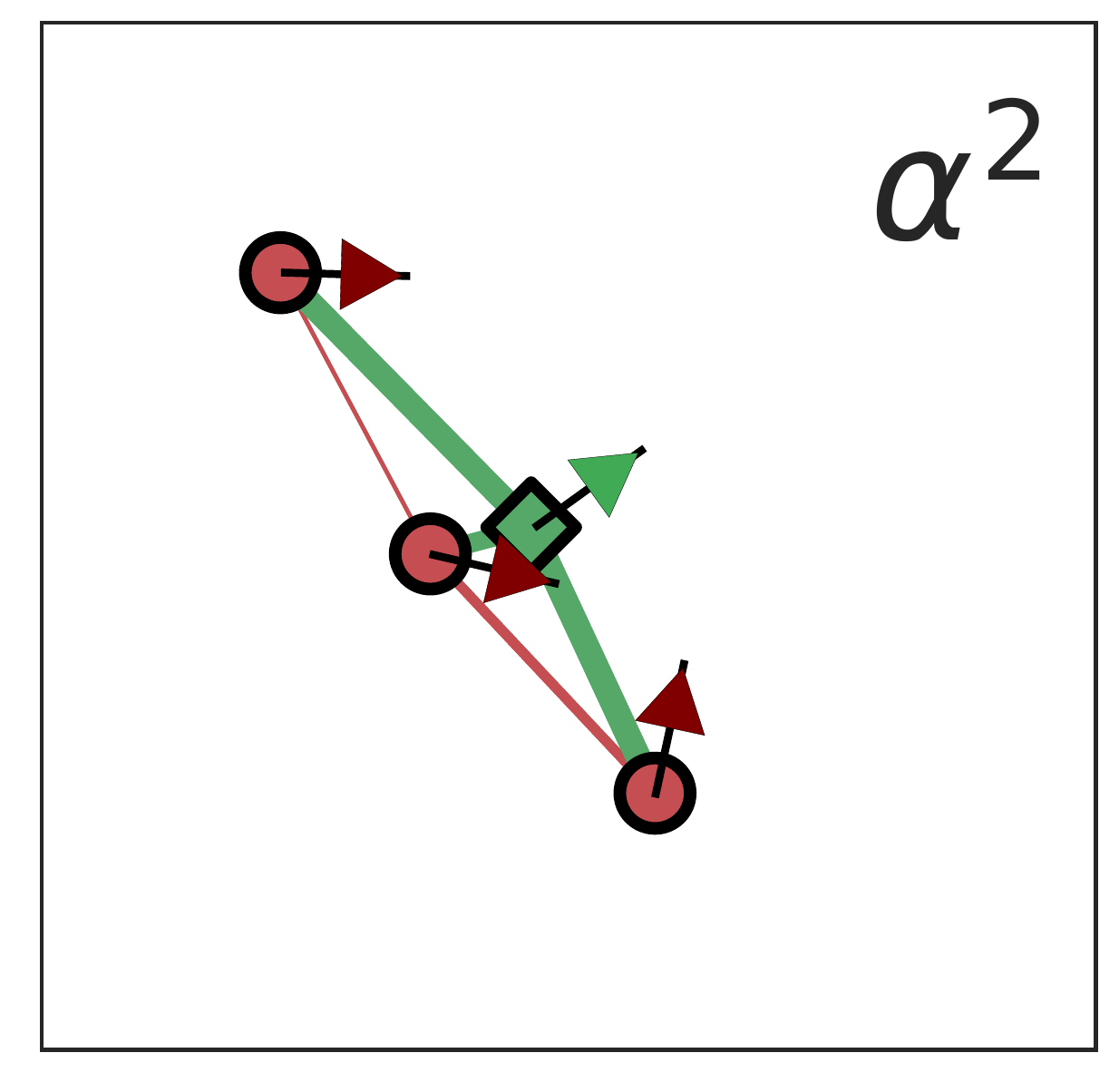}
\end{minipage}

\begin{minipage}[c]{.99\textwidth}
  \centering
  \includegraphics[width=0.92\textwidth]{method_figure_0512_time.pdf}
\end{minipage}
\caption{Multi-head attention of GAT in REMAX in predator-prey.}
\label{fig:fig_pp_attention}
\end{figure*}

\subsubsection{Effectiveness of the REMAX Exploration Score}

Table \ref{table:table_surrogate} presents the number of training episodes according to the types of exploration scores for the surrogate model of REMAX in a cooperative navigation with two agents. In the table, $r$ is the case where the score is replaced with agents' naive rewards corresponding to states. Burrowing and Covering are reward signals designed empirically to minimize and maximize the inverses of state visit counts of agents, which are similar to intrinsic rewards \cite{iqbal2019coordinated}, thus boosting exploitation and exploration, respectively. In the table, Equation \ref{eq:1} requires fewer episodes than the other types of scores. This may be because the naive reward is too sparse to be used as a score, and 
\setcounter{figure}{2}
\begin{figure}[h]
\begin{minipage}[c]{.15\textwidth}
  \centering
  \includegraphics[width=0.9\textwidth]{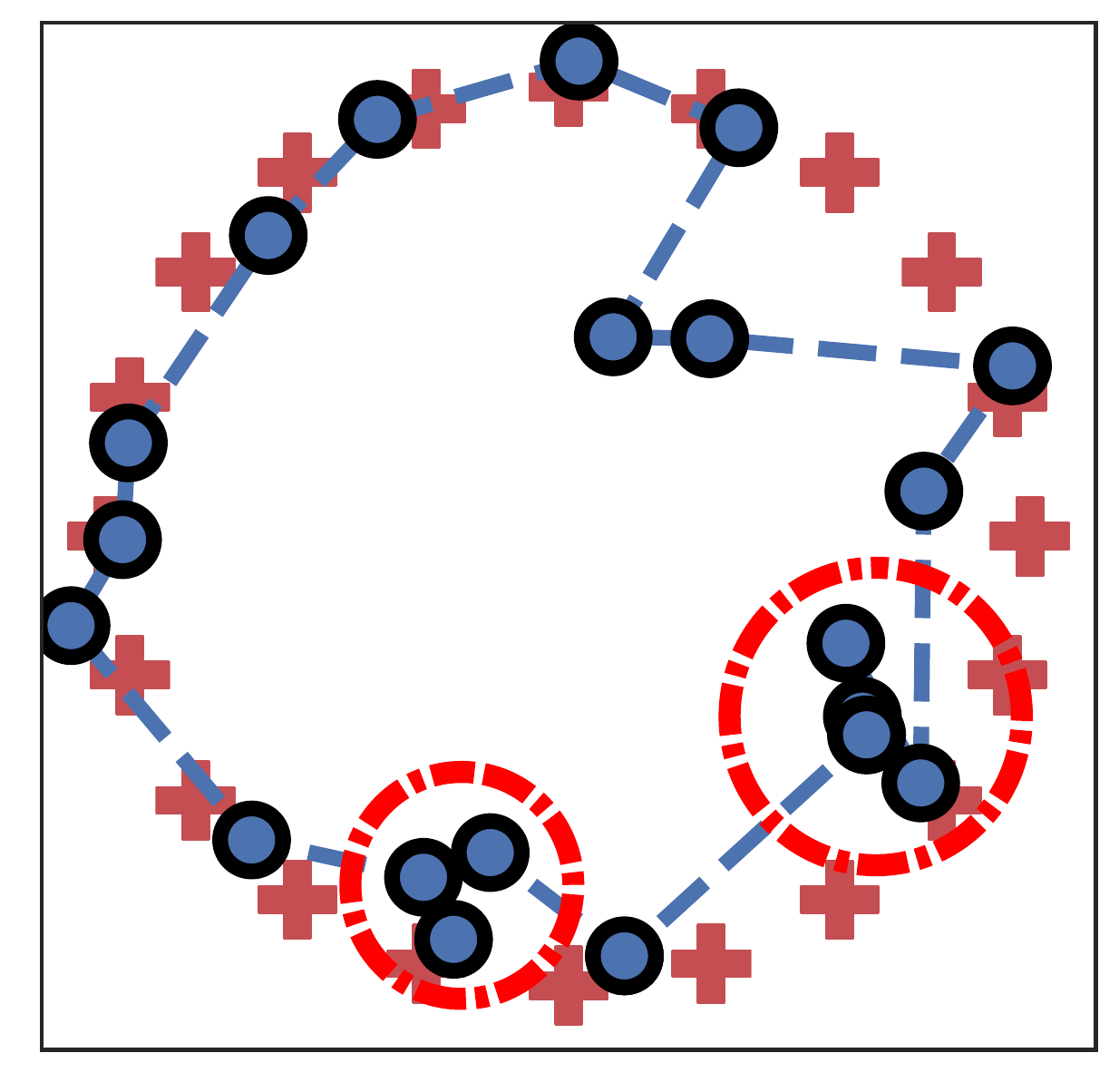}
  \captionsetup{labelformat=empty}
  \caption{a) Random}
\end{minipage}
\begin{minipage}[c]{.15\textwidth}
  \centering
  \includegraphics[width=0.9\textwidth]{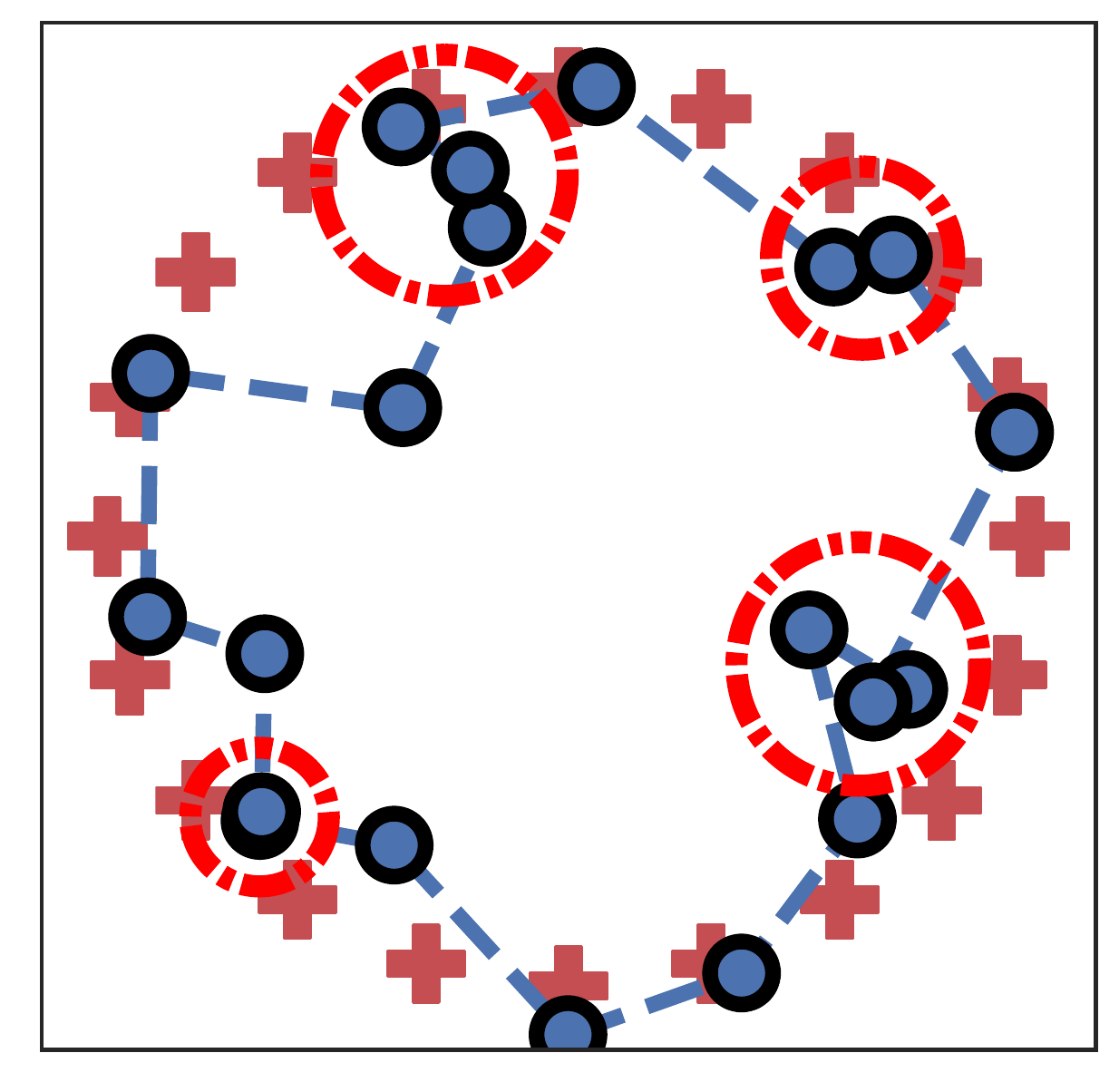}
  \captionsetup{labelformat=empty}
  \caption{b) GENE}
\end{minipage}
\begin{minipage}[c]{.15\textwidth}
  \centering
  \includegraphics[width=0.9\textwidth]{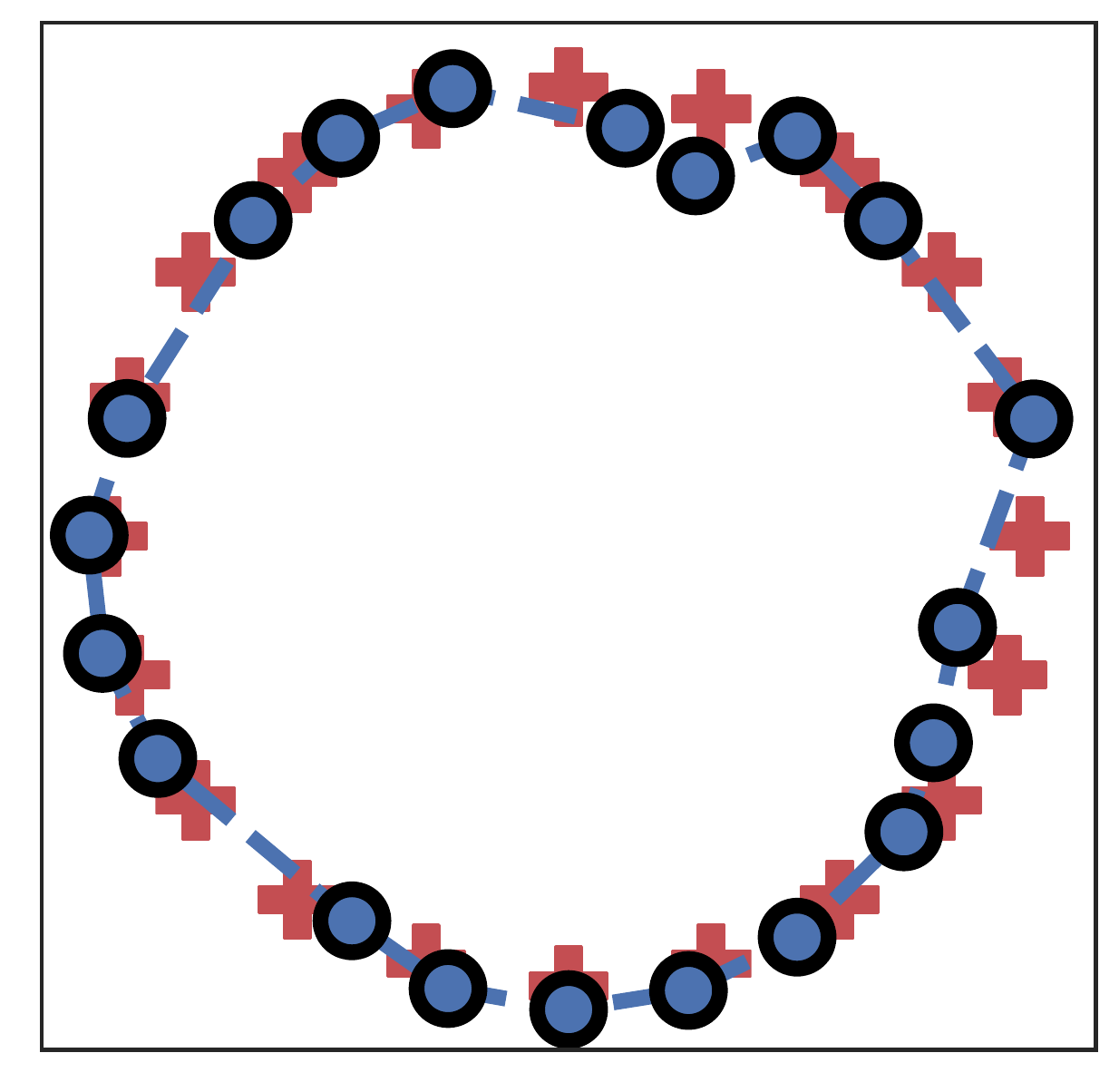}
  \captionsetup{labelformat=empty}
  \caption{c) REMAX}
\end{minipage}
\caption{Occupation in cooperative navigation with 20 agents.}
\label{fig:occ_cn}
\end{figure}
\begin{table}[h]
\centering
  \caption{Occupation rate in cooperative navigation.}
  \label{table:occ_rate_cn}
  \centering
    \begin{tabular}{cccc}
        \toprule
        \multicolumn{1}{c}{} &  
        \multicolumn{1}{c}{Random}&
        \multicolumn{1}{c}{GENE}&
        \multicolumn{1}{c}{REMAX}\\
        \midrule
        \multicolumn{1}{c}{10 agents} & 0.71\scriptsize $\pm$ 0.23 & 
        0.82\scriptsize $\pm$ 0.21 & \textbf{0.96}\scriptsize $\pm$ 0.05\\
        \multicolumn{1}{c}{20 agents} & 0.55\scriptsize $\pm$ 0.19 & 
        0.63\scriptsize $\pm$ 0.23 & \textbf{0.88}\scriptsize $\pm$ 0.12\\
        \multicolumn{1}{c}{30 agents} & 0.42\scriptsize $\pm$ 0.17 & 
        0.51\scriptsize $\pm$ 0.21 & \textbf{0.81}\scriptsize $\pm$ 0.14\\
        \bottomrule
    \end{tabular}
\end{table}
the others are designed only to induce specific behaviors of agents, while Equation \ref{eq:1} balances exploitation and exploration and boosts MARL training. The balance of Equation \ref{eq:1} changes according to the hyper-parameter $\lambda$, and it is discussed in the appendix.

\setcounter{figure}{7}
\begin{figure*}[t]

\begin{minipage}[c]{.153\textwidth}
  \centering
  \includegraphics[width=0.85\textwidth]{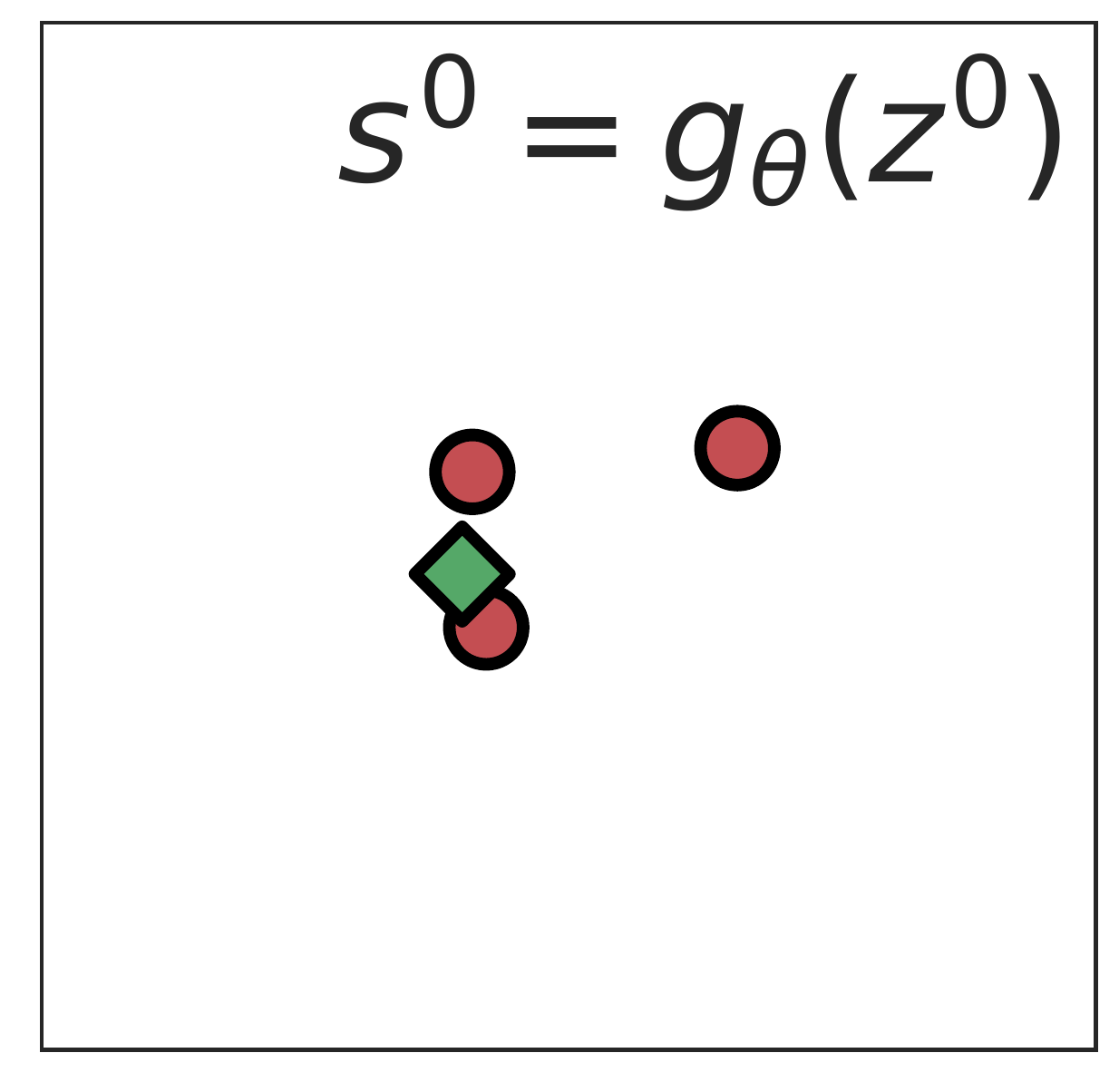}
\end{minipage}
\hspace{0.0cm}
\begin{minipage}[c]{.153\textwidth}
  \centering
  \includegraphics[width=0.85\textwidth]{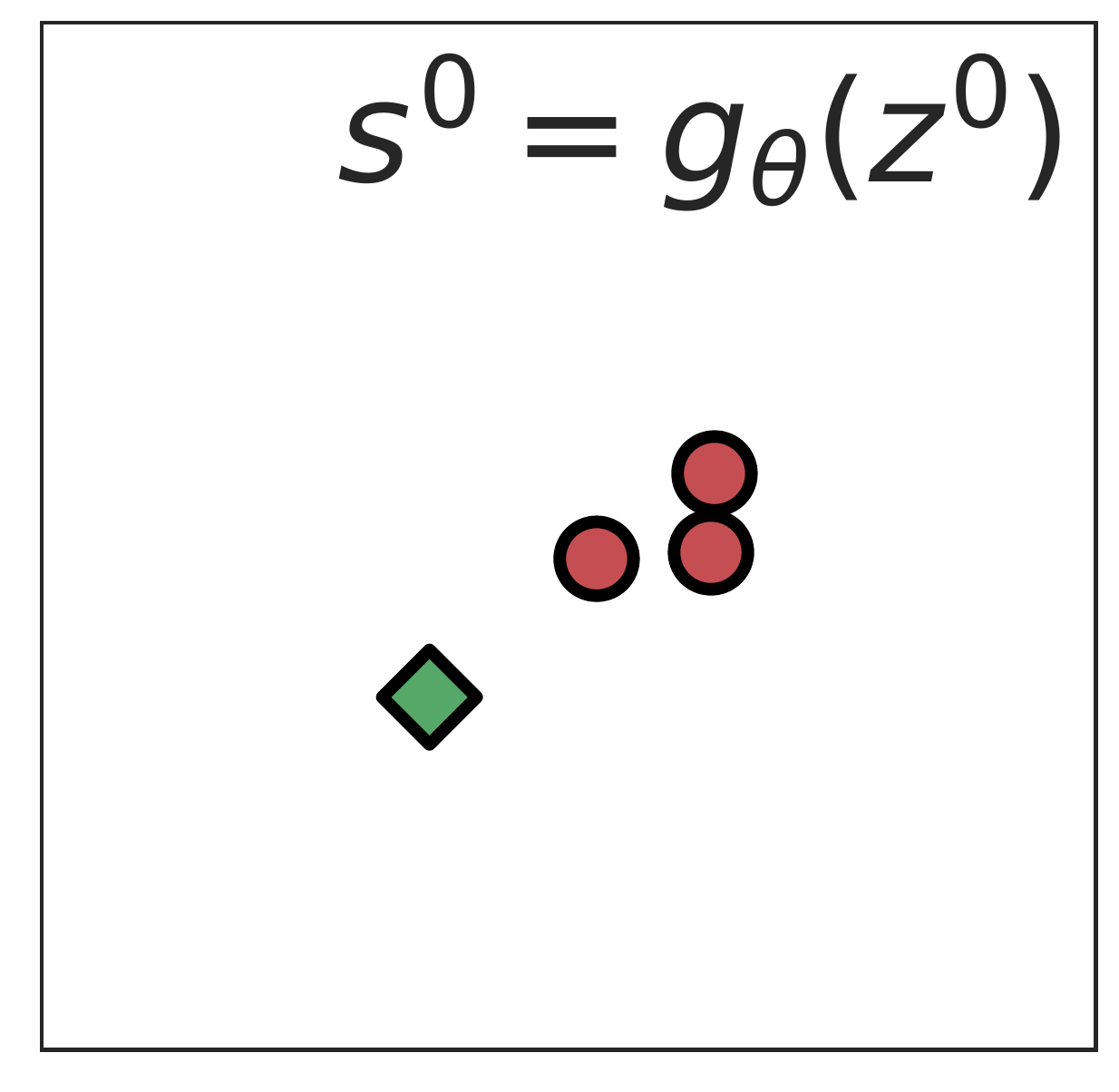}
\end{minipage}
\hspace{0.0cm}
\begin{minipage}[c]{.153\textwidth}
  \centering
  \includegraphics[width=0.85\textwidth]{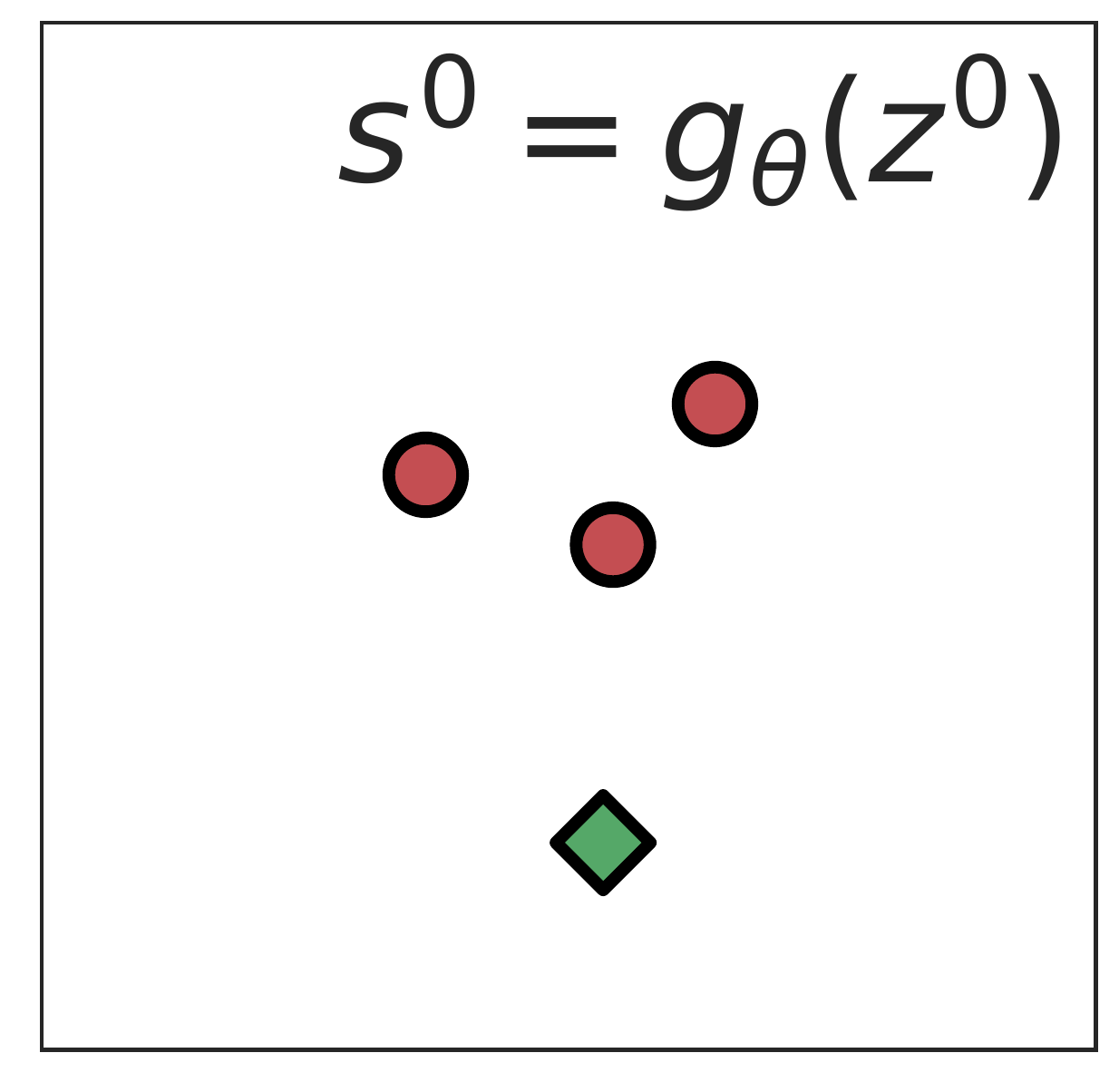}
\end{minipage}
\hspace{0.0cm}
\begin{minipage}[c]{.153\textwidth}
  \centering
  \includegraphics[width=0.85\textwidth]{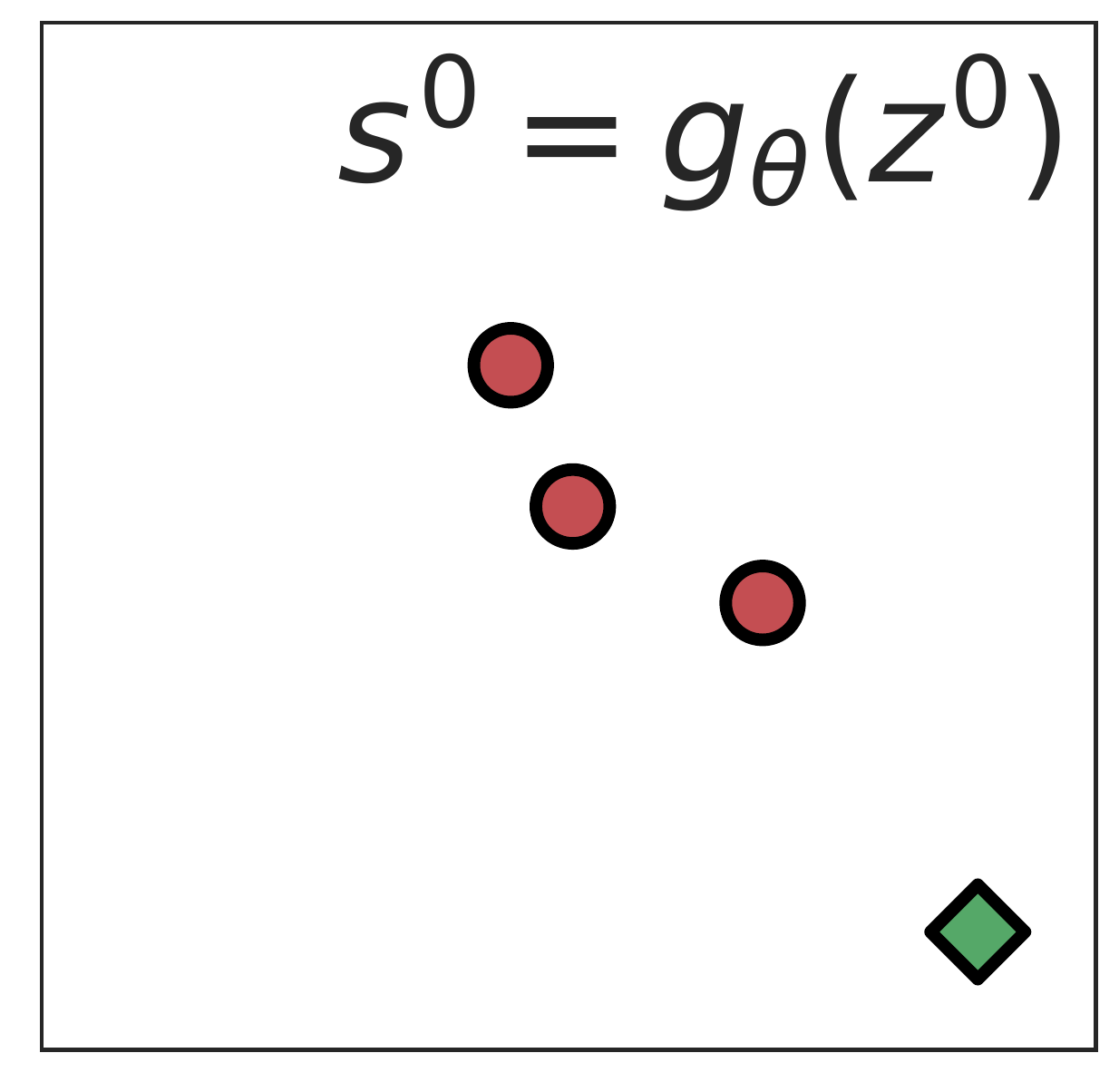}
\end{minipage}
\hspace{0.0cm}
\begin{minipage}[c]{.153\textwidth}
  \centering
  \includegraphics[width=0.85\textwidth]{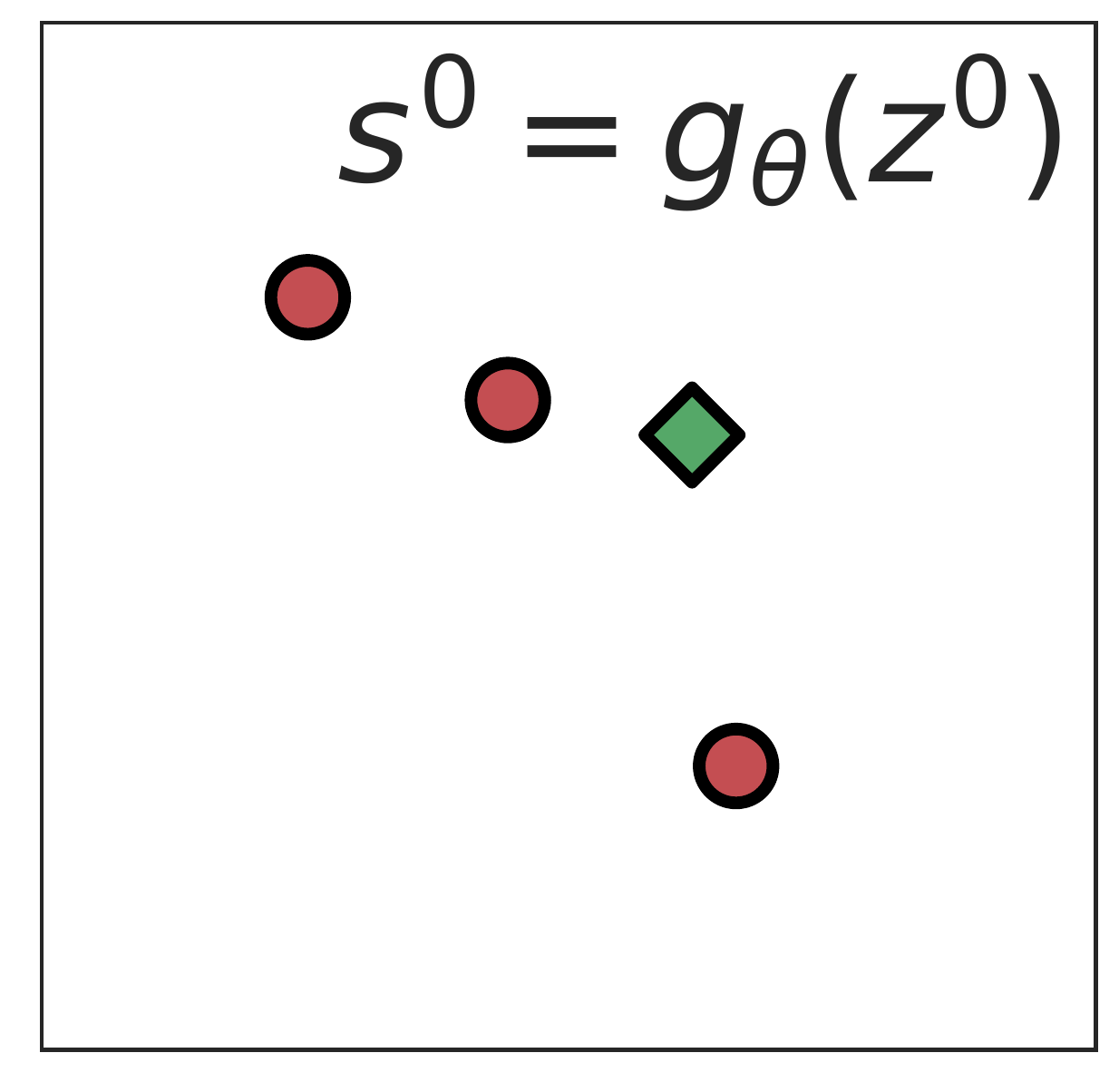}
\end{minipage}
\hspace{0.0cm}
\begin{minipage}[c]{.153\textwidth}
  \centering
  \includegraphics[width=0.85\textwidth]{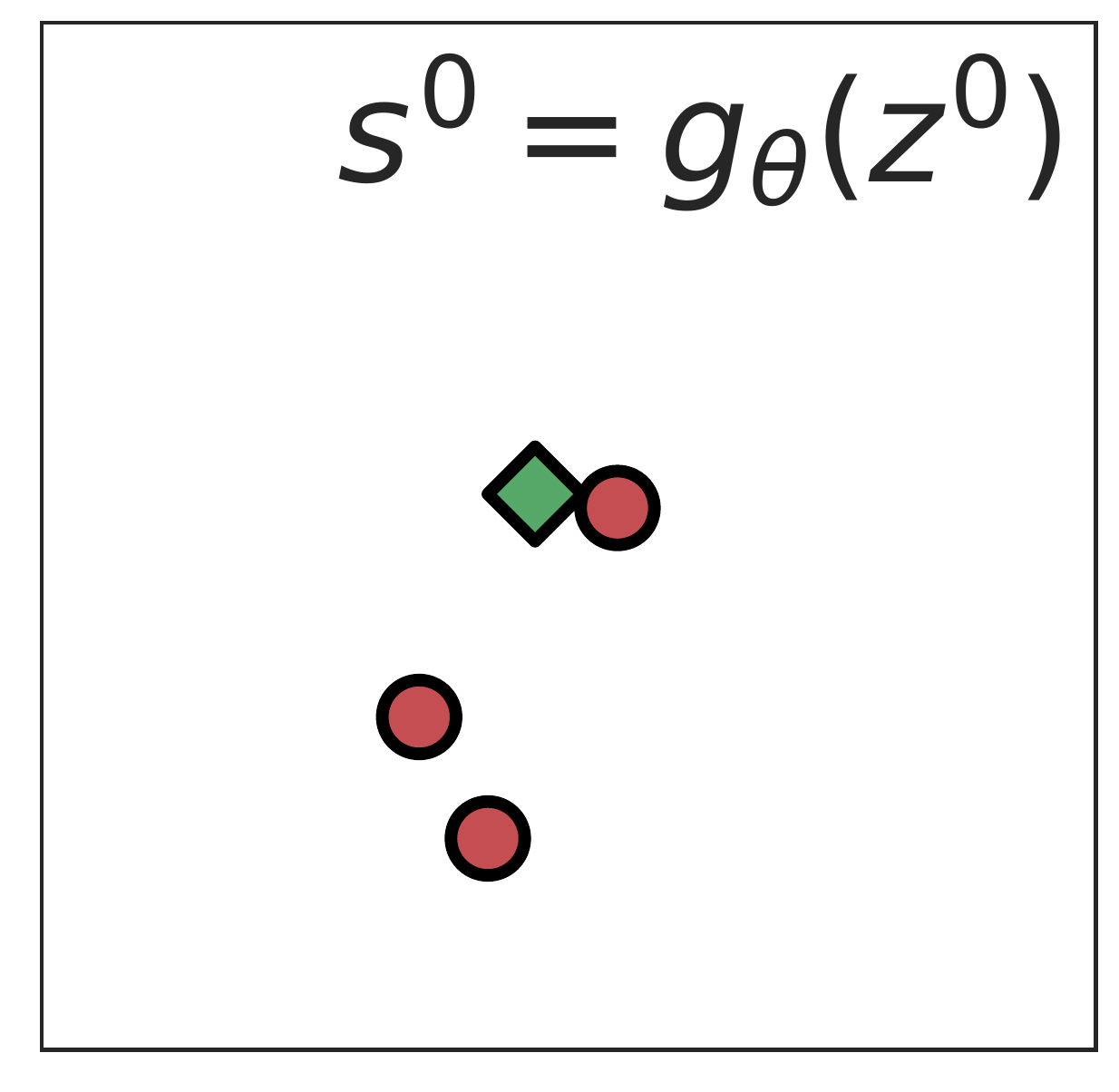}
\end{minipage}

\begin{minipage}[c]{.153\textwidth}
  \centering
  \includegraphics[width=0.85\textwidth]{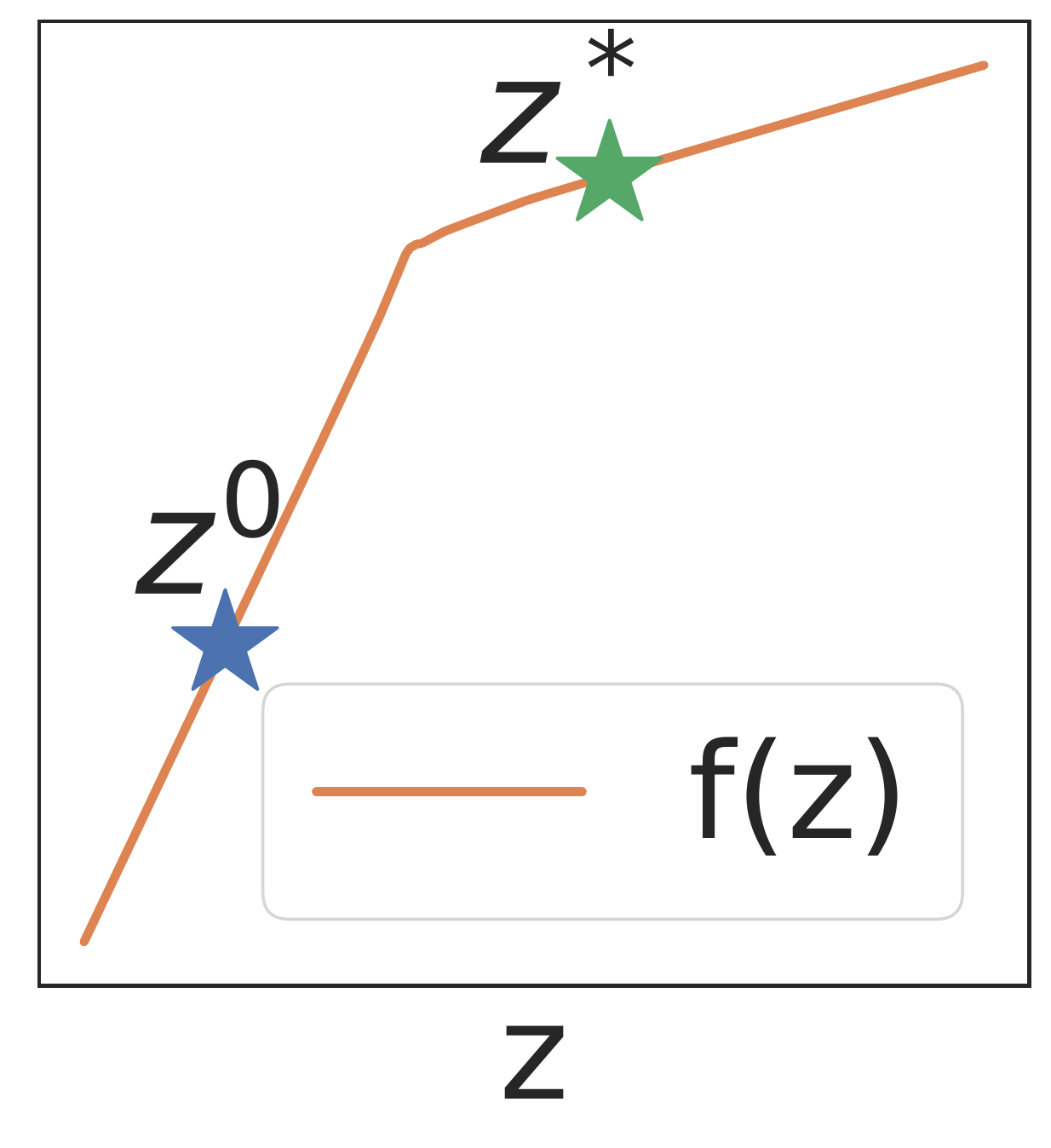}
\end{minipage}
\hspace{0.0cm}
\begin{minipage}[c]{.153\textwidth}
  \centering
  \includegraphics[width=0.85\textwidth]{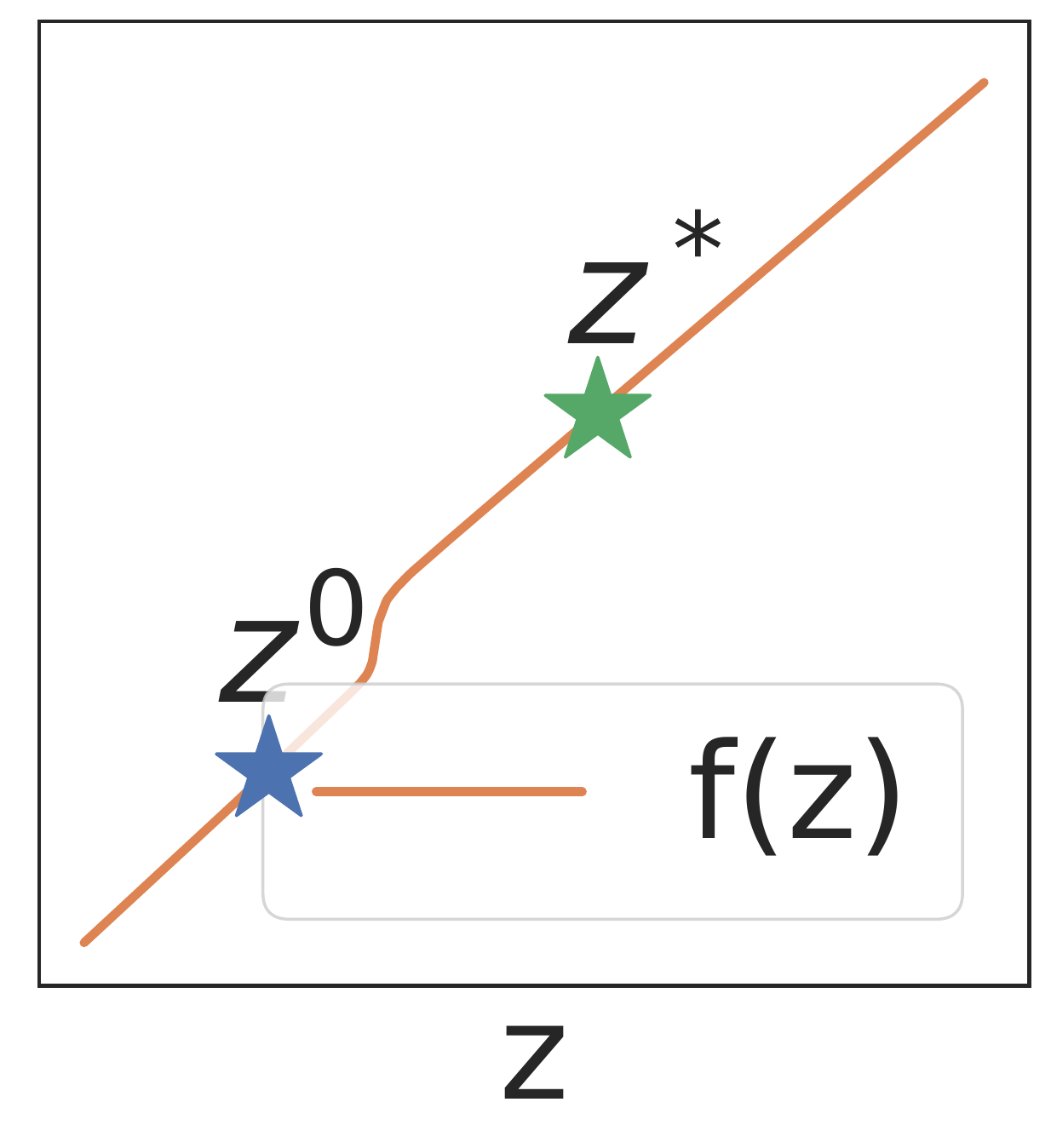}
\end{minipage}
\hspace{0.0cm}
\begin{minipage}[c]{.153\textwidth}
  \centering
  \includegraphics[width=0.85\textwidth]{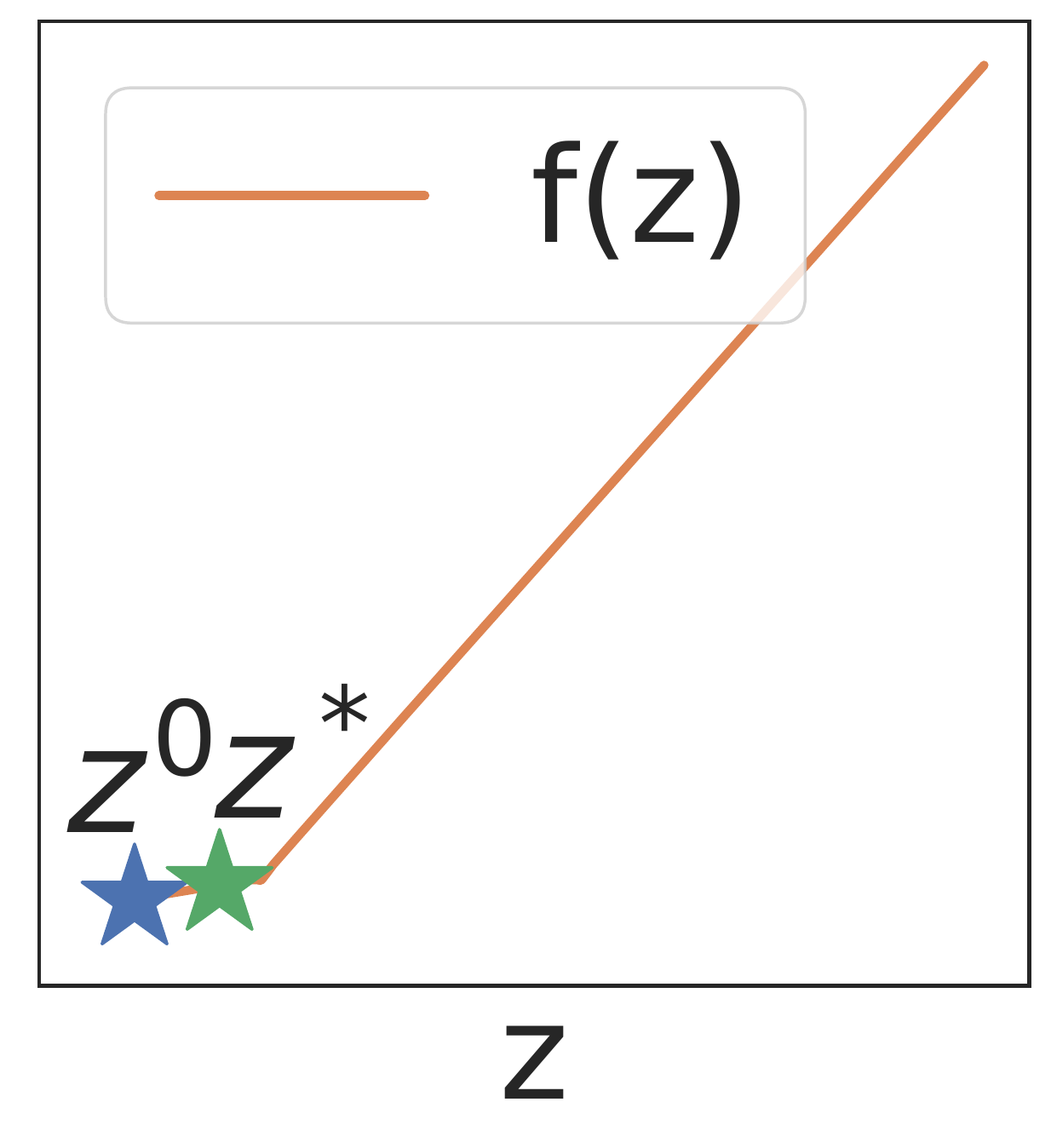}
\end{minipage}
\hspace{0.0cm}
\begin{minipage}[c]{.153\textwidth}
  \centering
  \includegraphics[width=0.85\textwidth]{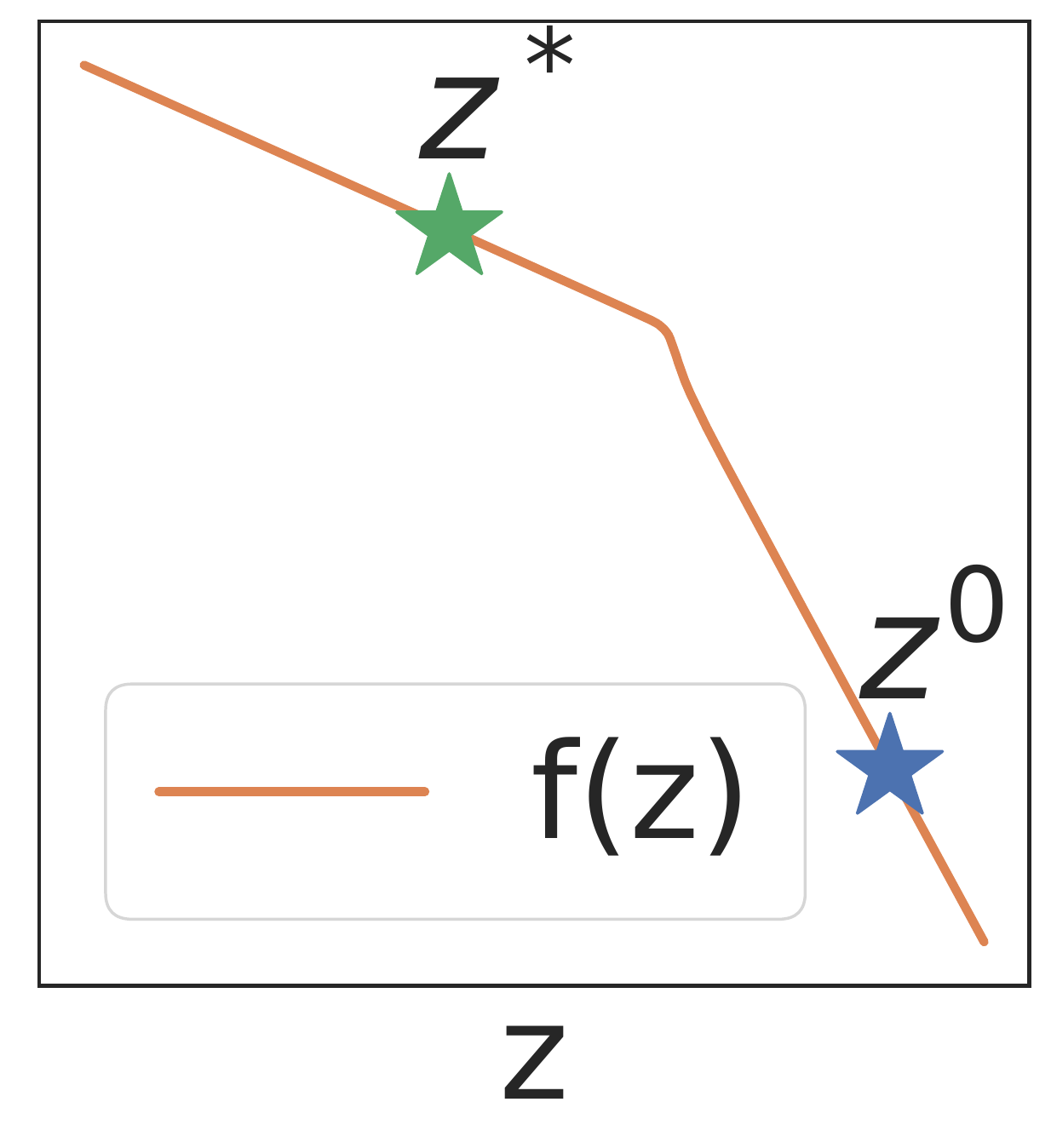}
\end{minipage}
\hspace{0.0cm}
\begin{minipage}[c]{.153\textwidth}
  \centering
  \includegraphics[width=0.85\textwidth]{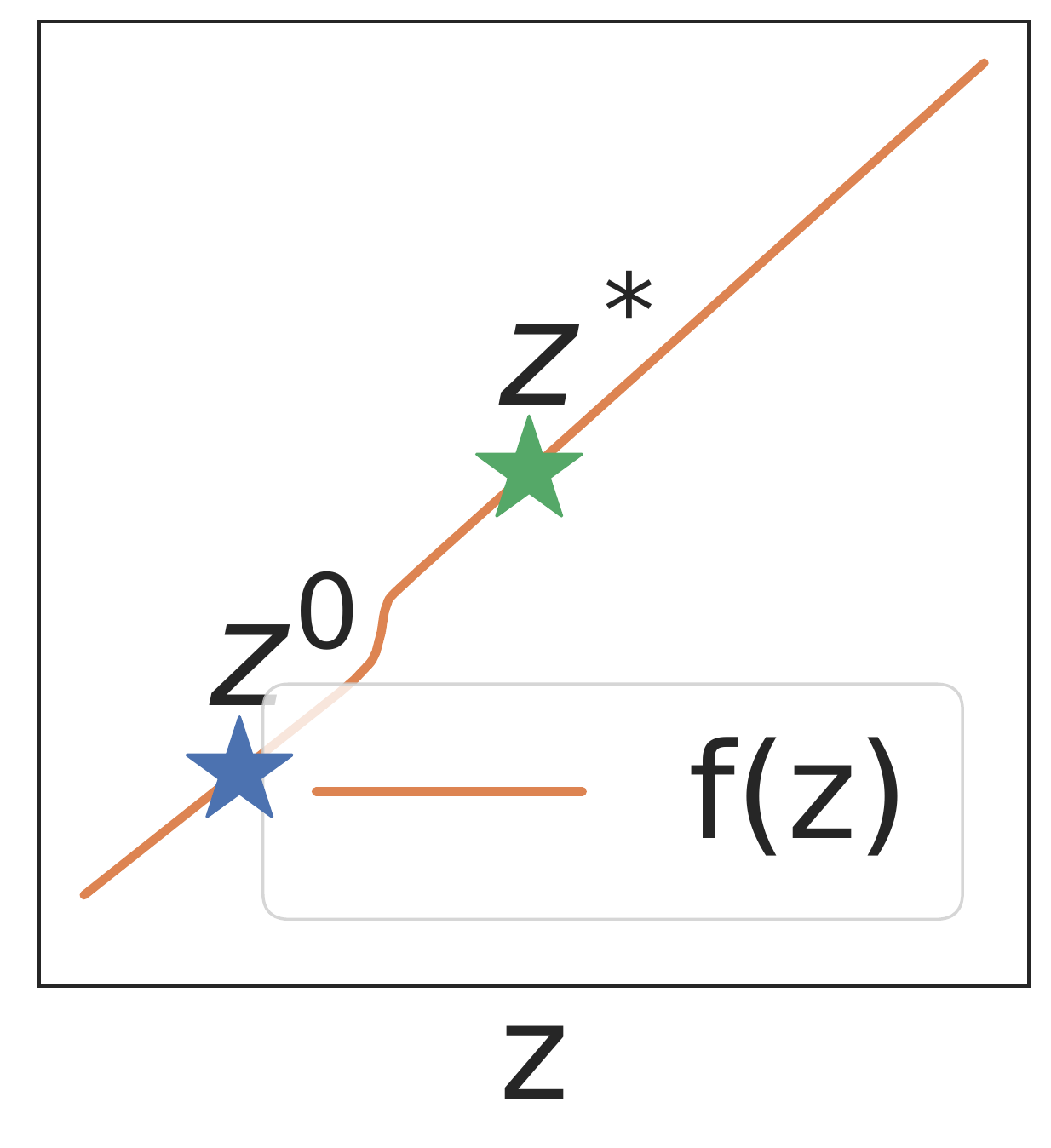}
\end{minipage}
\hspace{0.0cm}
\begin{minipage}[c]{.153\textwidth}
  \centering
  \includegraphics[width=0.85\textwidth]{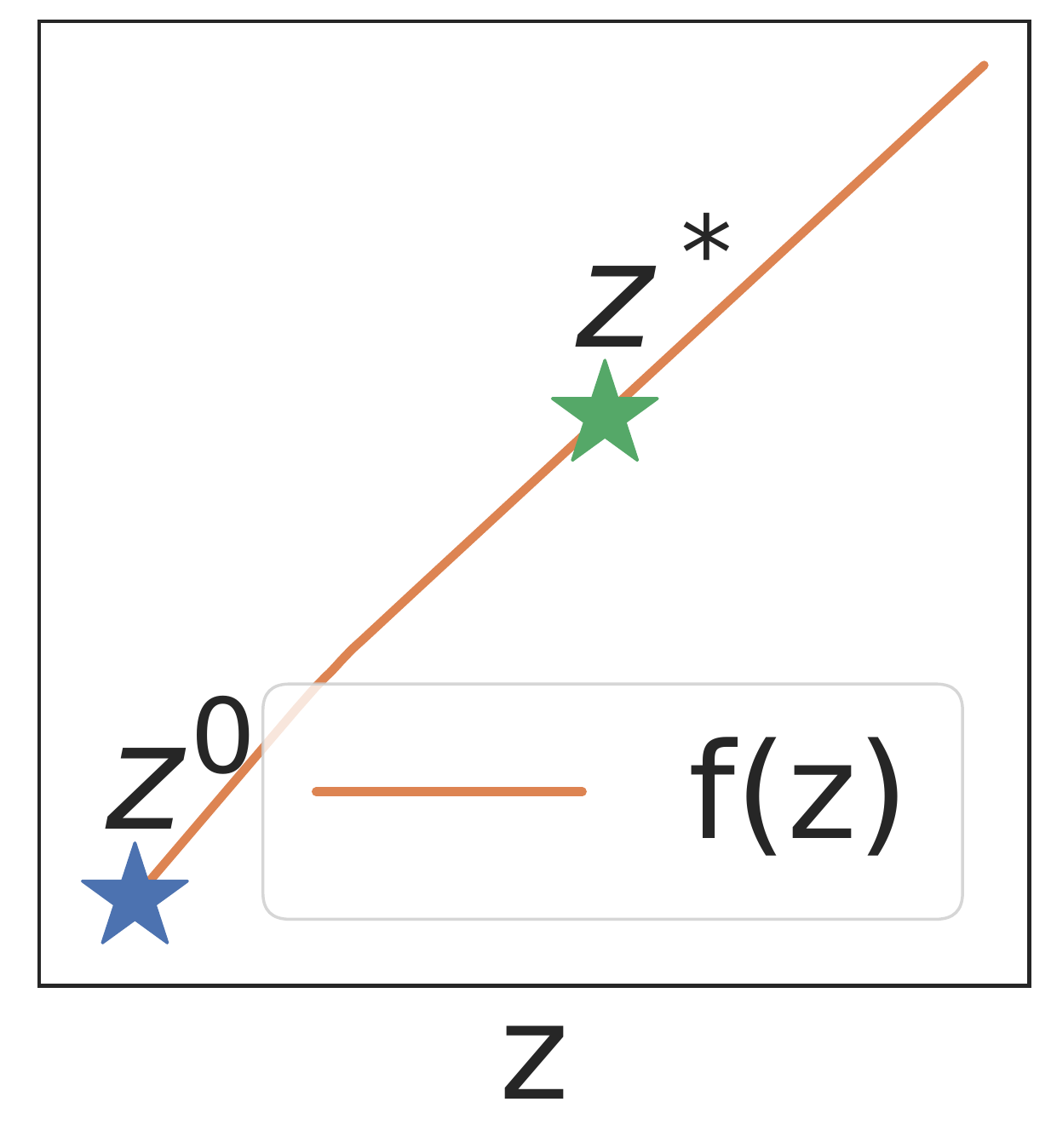}
\end{minipage}

\begin{minipage}[c]{.153\textwidth}
  \centering
  \includegraphics[width=0.85\textwidth]{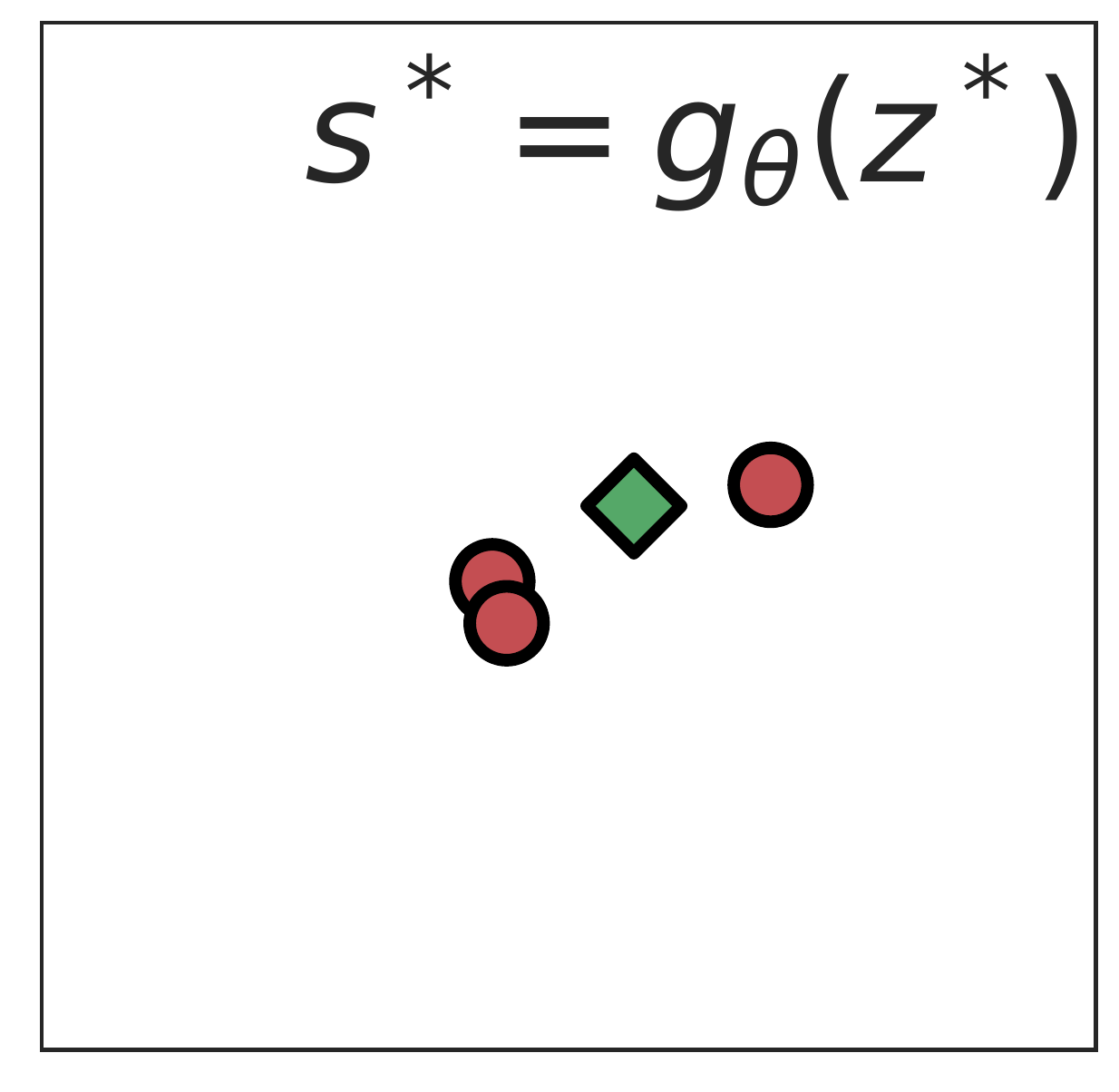}
\end{minipage}
\hspace{0.0cm}
\begin{minipage}[c]{.153\textwidth}
  \centering
  \includegraphics[width=0.85\textwidth]{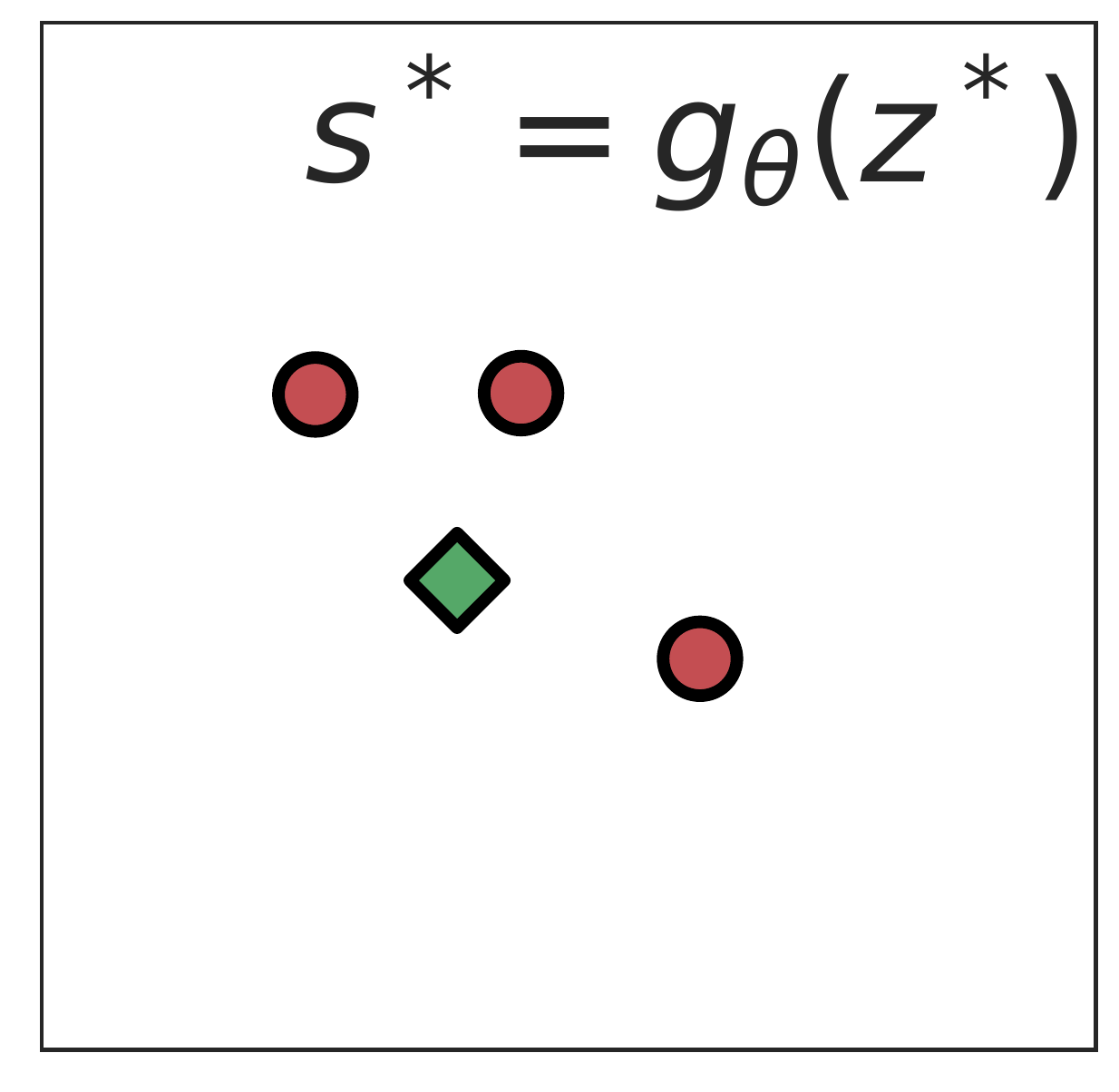}
\end{minipage}
\hspace{0.0cm}
\begin{minipage}[c]{.153\textwidth}
  \centering
  \includegraphics[width=0.85\textwidth]{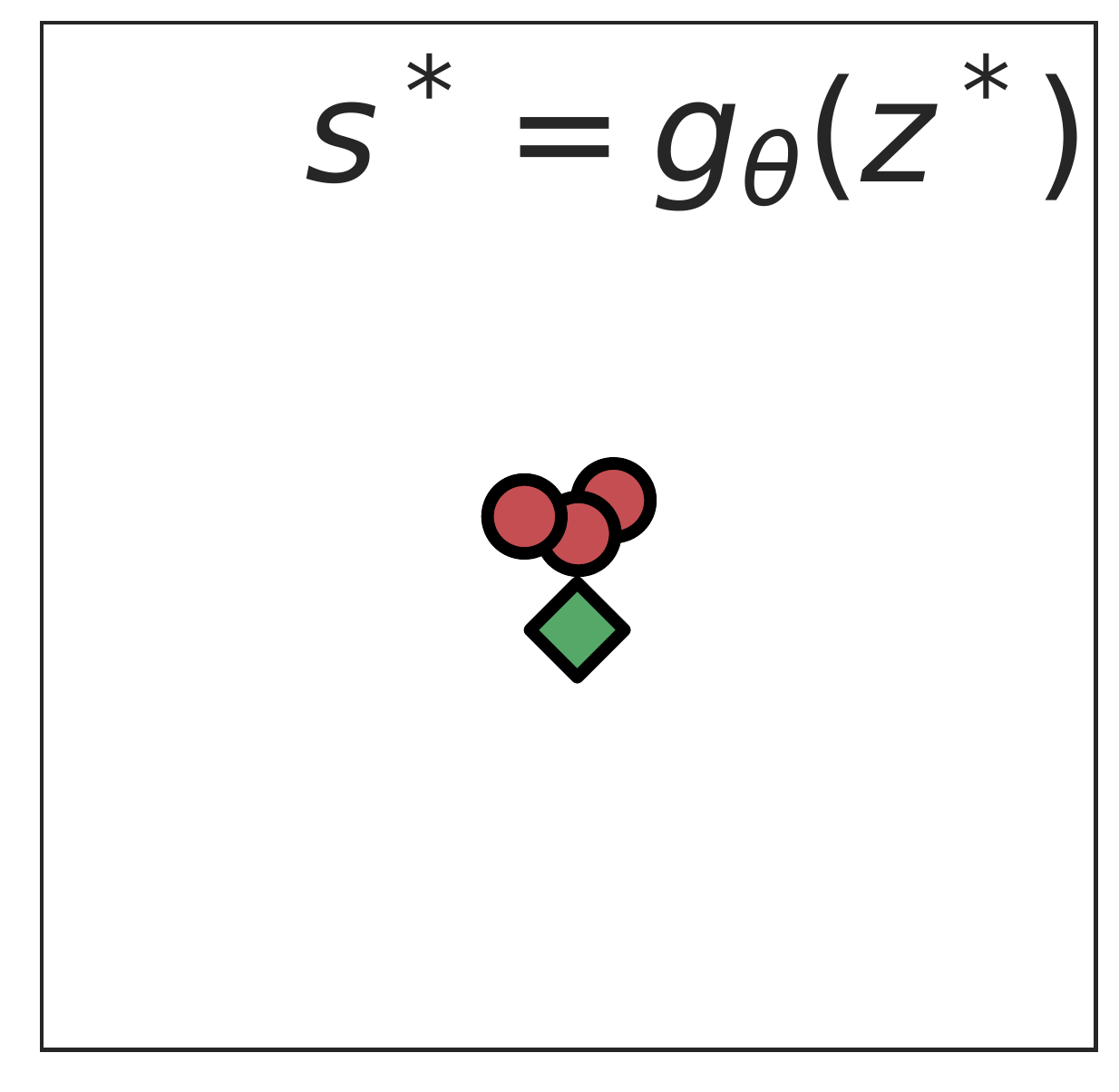}
\end{minipage}
\hspace{0.0cm}
\begin{minipage}[c]{.153\textwidth}
  \centering
  \includegraphics[width=0.85\textwidth]{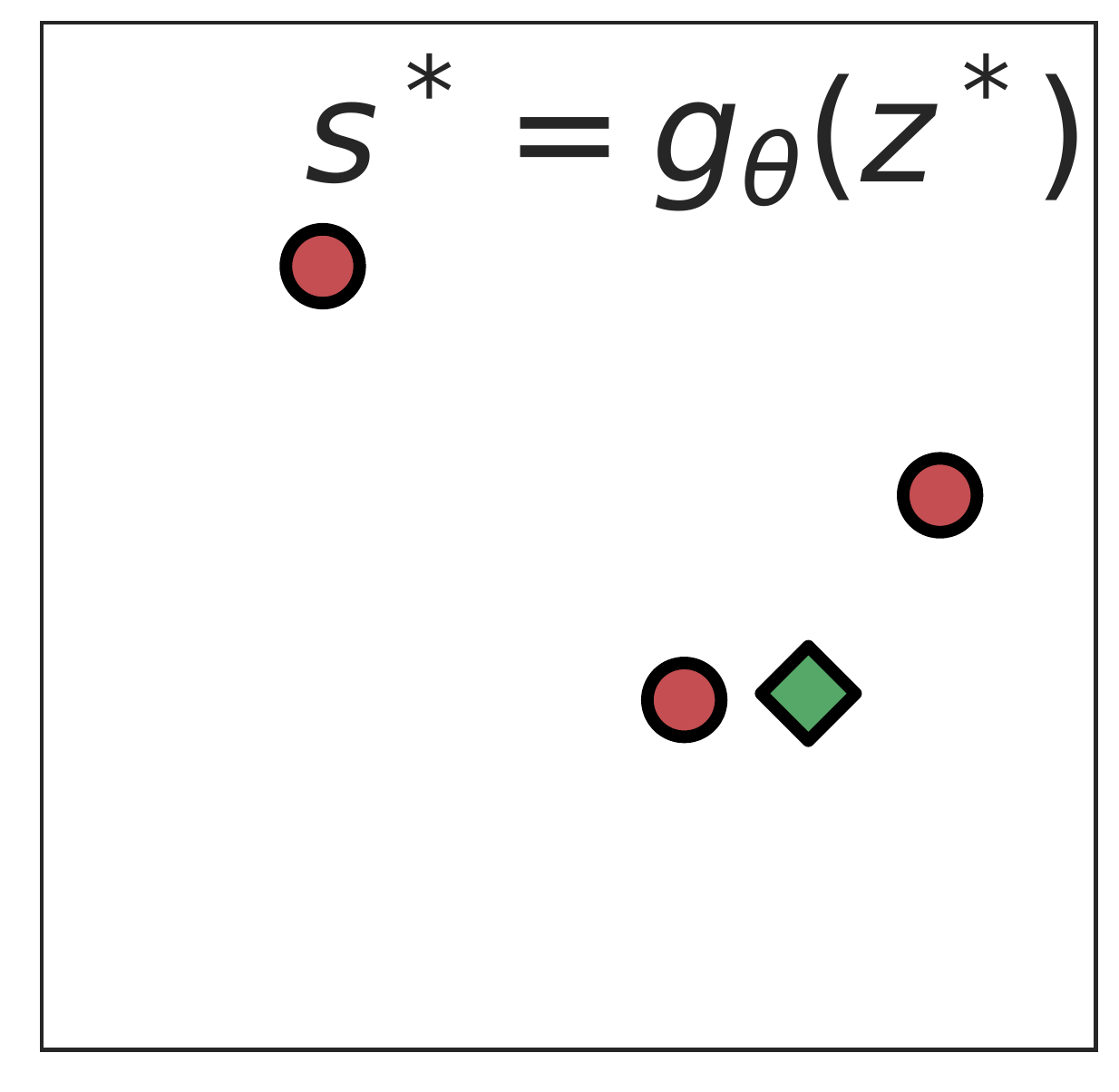}
\end{minipage}
\hspace{0.0cm}
\begin{minipage}[c]{.153\textwidth}
  \centering
  \includegraphics[width=0.85\textwidth]{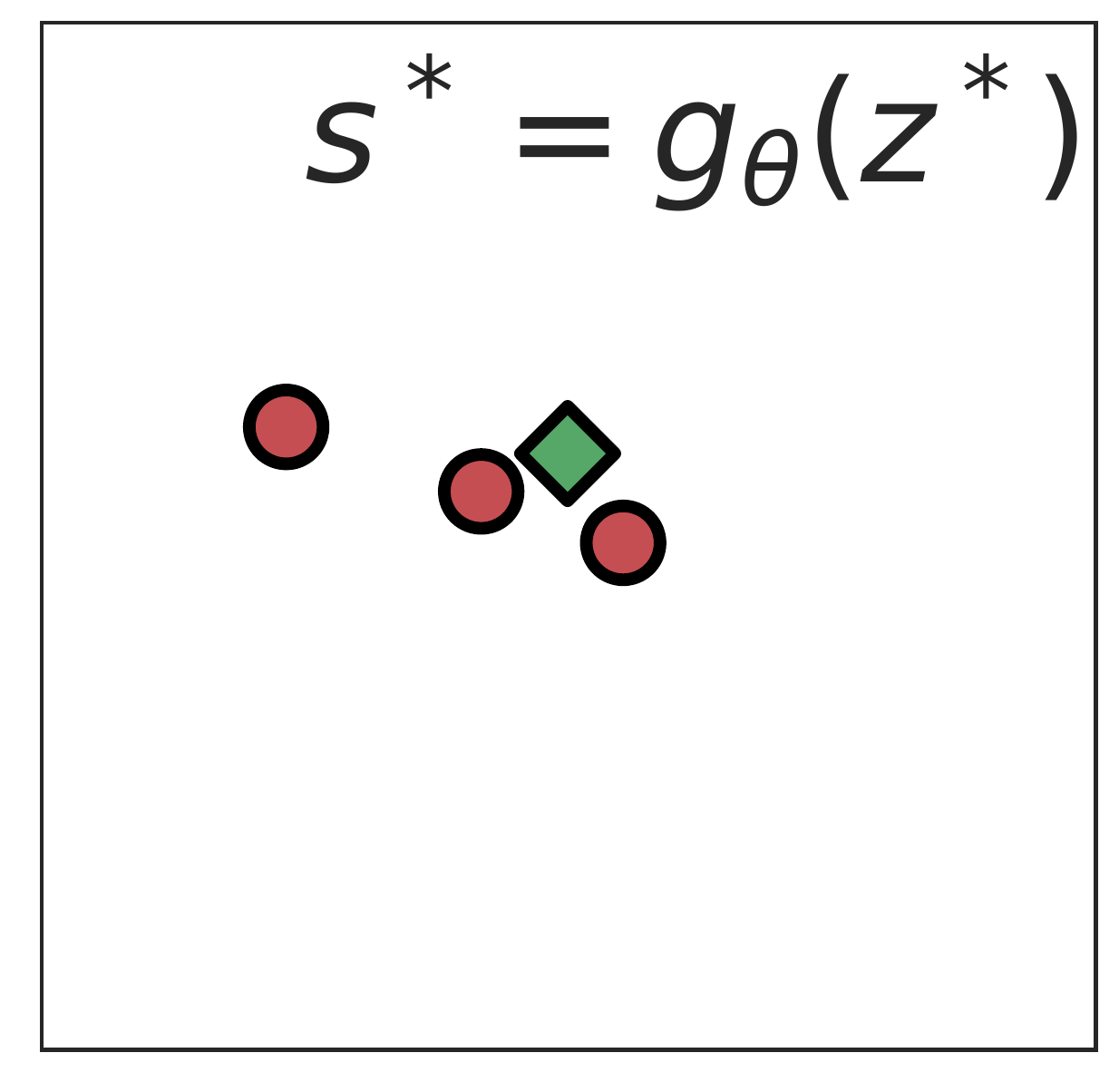}
\end{minipage}
\hspace{0.0cm}
\begin{minipage}[c]{.153\textwidth}
  \centering
  \includegraphics[width=0.85\textwidth]{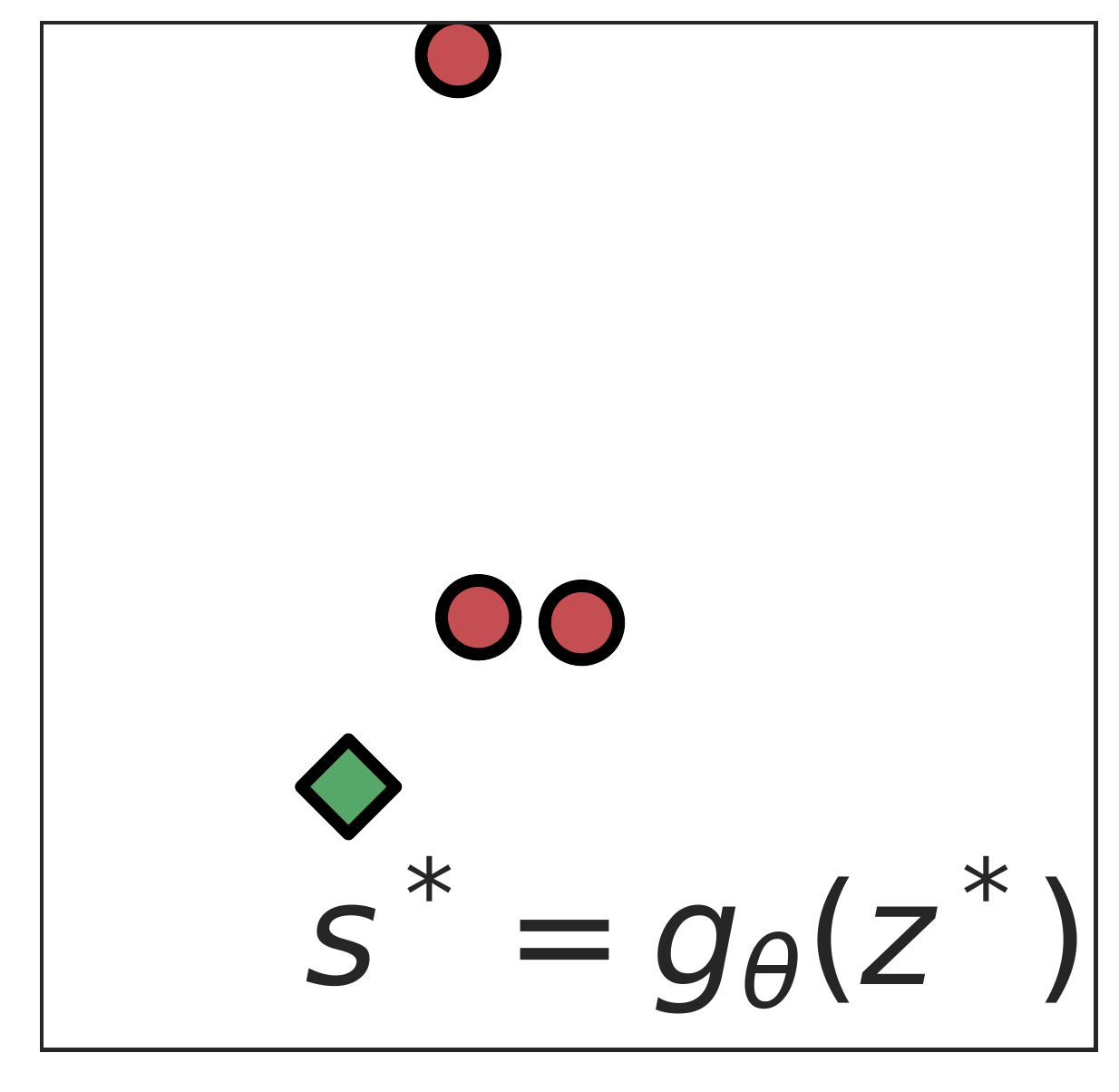}
\end{minipage}

\begin{minipage}[c]{.99\textwidth}
  \centering
  \includegraphics[width=0.92\textwidth]{method_figure_0512_time.pdf}
\end{minipage}
\caption{Surrogate models and generated states in predator-prey.}
\label{fig:fig_pp_1d}
\end{figure*}

\subsubsection{Scalability Test}
Table \ref{table:occ_rate_cn} presents the rate of occupied landmarks in cooperative navigation with more agents. In the table, REMAX has higher occupation rates than the other methods, which implies that the agents trained by REMAX cooperate effectively to occupy distinct landmarks, even when their numbers are large. Figure \ref{fig:occ_cn} shows how 20 agents occupy landmarks positioned in a circular shape after trained by each method. In the figure, the agents in Random and GENE tend to cluster (dotted red circles) at some landmarks, while the agents in REMAX occupy distinct landmarks.



\subsection{Predator-Prey}
As shown in Figure \ref{fig:fig_illustrations} (c), predator-prey is a multi-agent environment where three homogeneous predator agents (red circles) aim to capture one prey agent (green diamond), which is called 3 vs. 1 predator-prey. Because the prey has a faster speed and acceleration than predators, the predators are required to cooperate for capturing the prey. In the 3 vs. 1 game, each predator obtains a reward +10 when two or more predators capture the prey simultaneously, and 0 otherwise. Meanwhile, the prey obtains a reward -10 for getting captured. If there are 6 predators and 2 preys, it is denoted as 6 vs. 2 predator-prey, and the rewards are obtained when three or more predators capture the prey simultaneously. 9 vs. 3 predator-prey is also similar to 6 vs. 2 predator-prey.
In predator-prey, we used the GAT encoder with a $K=2$ multi-head attention for REMAX because there are at least two types of relationships, predator-predator and predator-prey. We exclude HER and RCG as they are not suitable for a non-goal-oriented task.
Instead, we consider EDTI, which uses intrinsic rewards quantifying the influences of one agent's action on other agents' rewards and transitions.

\begin{table}[t]
  \caption{Returns of predators in predator-prey.}
  \label{table:table_pp}
  \centering
    \begin{tabular}{ccccc}
        \toprule
        \multicolumn{1}{c}{} &  
        \multicolumn{1}{c}{Random}&
        \multicolumn{1}{c}{EDTI}&
        \multicolumn{1}{c}{GENE}&
        \multicolumn{1}{c}{REMAX}\\
        \midrule
        \multicolumn{1}{c}{3 vs. 1} & 1.6\scriptsize $\pm$ 0.4 &
        3.7\scriptsize $\pm$ 1.0 &
        14.3\scriptsize $\pm$ 4.9 & \textbf{18.7}\scriptsize $\pm$ 4.6\\
        \multicolumn{1}{c}{6 vs. 2} & 1.8\scriptsize $\pm$ 0.6 &
        2.8\scriptsize $\pm$ 0.8 &
        8.3\scriptsize $\pm$ 1.8 & \textbf{13.5}\scriptsize $\pm$ 4.4\\
        \multicolumn{1}{c}{9 vs. 3} & 2.6\scriptsize $\pm$ 0.7 &
        3.2\scriptsize $\pm$ 1.1 &
        9.1\scriptsize $\pm$ 2.7 & \textbf{14.4}\scriptsize $\pm$ 4.1\\
        \bottomrule
    \end{tabular}
\end{table}

Table \ref{table:table_pp} compares the returns of the predators. After the predators and prey are trained together by each method, the predators are validated for 200 test episodes against the prey trained by a random exploration.
In the table, REMAX has higher returns than the other methods, which implies that the predators trained by REMAX cooperate effectively to capture the prey. Meanwhile, EDTI ends up achieving significantly lower returns than GENE and REMAX. These lower returns may be because EDTI cannot accurately predict the influences of intrinsic rewards. Accurate prediction requires many samples, especially when there are many agents in a continuous state-action space.



\subsubsection{Analysis of Representing States with Multi-Head Attention}

Figure \ref{fig:fig_pp_attention} shows two sets of the normalized attention coefficients, $\alpha^1$ and $\alpha^2$, of the GAT encoder with a $K=2$ multi-head attention in REMAX. The two sets of coefficients can quantify the levels of cooperation among predators and the competition between predators and prey, respectively (upper and lower figures). In the figures, the coefficients (the thickness of the line) increase when two predators with their actions (red arrows) approach the prey with its action (green arrows) simultaneously. For example, in the fourth figure of Figure \ref{fig:fig_pp_attention}, two predators relate to a strong attention (upper figure), and, at the same time, both predators are connected strongly with the prey they are about to capture together (lower figure). We identified that the two predators connected by larger attention coefficients tend to cooperate to capture the prey. 

\subsubsection{Analysis of Generating States}
Figure \ref{fig:fig_pp_1d} compares the two types of states: $s^0$ decoded from an initial latent vector $z^0$ and $s^*$ decoded from an optimized latent vector $z^*$. The second-row figures show $z^0$ and $z^*$ over $f_\psi(z)$. In addition, the first-and third-row figures show the decoded states, $s^0$ and $s^*$, from $z^0$ and $z^*$ using $g_\theta$. As MARL and REMAX are trained with more samples, as shown in the figure, REMAX tends to generate states where the predators are located more closely to the prey or surround the prey. This allows the predators to easily capture the prey and thus receive a reward, which expedites the MARL training.

\subsection{Starcraft Multi-Agent Challenge (SMAC)}
\begin{table}
  \caption{Winning rates of controlled marines in SMAC.}
  \label{table:table_sc}
  \centering
    \begin{tabular}{ccccc}
        \toprule
        \multicolumn{1}{c}{}&
        \multicolumn{1}{c}{Random}&
        \multicolumn{1}{c}{EDTI}&
        \multicolumn{1}{c}{GENE}&
        \multicolumn{1}{c}{REMAX}\\
        \midrule
        \multicolumn{1}{c}{3 vs. 3}&
        0.86\scriptsize $\pm$ 0.05 & 
        0.88\scriptsize $\pm$ 0.06 & 
        0.94\scriptsize $\pm$ 0.08 & \textbf{1.00}\scriptsize $\pm$ 0.00\\
        \multicolumn{1}{c}{5 vs. 6}&
        0.00\scriptsize $\pm$ 0.00 & 
        0.00\scriptsize $\pm$ 0.00 & 
        0.10\scriptsize $\pm$ 0.04 & \textbf{0.43}\scriptsize $\pm$ 0.11\\
        \bottomrule
    \end{tabular}
\end{table}

SMAC \cite{smac} is a more realistic multi-agent environment, as shown in Figure \ref{fig:fig_illustrations} (d), where three marines (agents) aim to fight against equivalent marines of
game AI. SMAC originally has dense reward signals. However, we simply modify the SMAC's rewards to be more sparse at the final timestep of each episode: a victory reward 1, while a defeat -1, with an attack reward 1 to encourage attacking the enemies, rather than running away.

Table \ref{table:table_sc} presents winning rates of controlled marines against game-AI marines in SMAC. 
In the table, 
REMAX outperforms other methods with higher winning rates.


\section{Conclusions}
We proposed REMAX, an exploration method that generates initial states for accelerating the training of a MARL model. Empirically, we demonstrated that REMAX generates states by representing relationships among agents, and the generated states improve the training and performance of the MARL model more than existing exploration methods.

\balance



\bibliographystyle{ACM-Reference-Format} 

\begin{thebibliography}{37}


\ifx \showCODEN    \undefined \def \showCODEN     #1{\unskip}     \fi
\ifx \showDOI      \undefined \def \showDOI       #1{#1}\fi
\ifx \showISBNx    \undefined \def \showISBNx     #1{\unskip}     \fi
\ifx \showISBNxiii \undefined \def \showISBNxiii  #1{\unskip}     \fi
\ifx \showISSN     \undefined \def \showISSN      #1{\unskip}     \fi
\ifx \showLCCN     \undefined \def \showLCCN      #1{\unskip}     \fi
\ifx \shownote     \undefined \def \shownote      #1{#1}          \fi
\ifx \showarticletitle \undefined \def \showarticletitle #1{#1}   \fi
\ifx \showURL      \undefined \def \showURL       {\relax}        \fi
\providecommand\bibfield[2]{#2}
\providecommand\bibinfo[2]{#2}
\providecommand\natexlab[1]{#1}
\providecommand\showeprint[2][]{arXiv:#2}

\bibitem[\protect\citeauthoryear{Andrychowicz, Wolski, Ray, Schneider, Fong,
  Welinder, McGrew, Tobin, Abbeel, and Zaremba}{Andrychowicz
  et~al\mbox{.}}{2017}]%
        {andrychowicz2017hindsight}
\bibfield{author}{\bibinfo{person}{Marcin Andrychowicz}, \bibinfo{person}{Filip
  Wolski}, \bibinfo{person}{Alex Ray}, \bibinfo{person}{Jonas Schneider},
  \bibinfo{person}{Rachel Fong}, \bibinfo{person}{Peter Welinder},
  \bibinfo{person}{Bob McGrew}, \bibinfo{person}{Josh Tobin},
  \bibinfo{person}{OpenAI~Pieter Abbeel}, {and} \bibinfo{person}{Wojciech
  Zaremba}.} \bibinfo{year}{2017}\natexlab{}.
\newblock \showarticletitle{Hindsight experience replay}. In
  \bibinfo{booktitle}{\emph{Advances in neural information processing
  systems}}. \bibinfo{pages}{5048--5058}.
\newblock


\bibitem[\protect\citeauthoryear{Bellemare, Srinivasan, Ostrovski, Schaul,
  Saxton, and Munos}{Bellemare et~al\mbox{.}}{2016}]%
        {bellemare2016unifying}
\bibfield{author}{\bibinfo{person}{Marc Bellemare}, \bibinfo{person}{Sriram
  Srinivasan}, \bibinfo{person}{Georg Ostrovski}, \bibinfo{person}{Tom Schaul},
  \bibinfo{person}{David Saxton}, {and} \bibinfo{person}{Remi Munos}.}
  \bibinfo{year}{2016}\natexlab{}.
\newblock \showarticletitle{Unifying count-based exploration and intrinsic
  motivation}. In \bibinfo{booktitle}{\emph{Advances in neural information
  processing systems}}. \bibinfo{pages}{1471--1479}.
\newblock


\bibitem[\protect\citeauthoryear{B{\"o}hmer, Rashid, and Whiteson}{B{\"o}hmer
  et~al\mbox{.}}{2019}]%
        {bohmer2019exploration}
\bibfield{author}{\bibinfo{person}{Wendelin B{\"o}hmer},
  \bibinfo{person}{Tabish Rashid}, {and} \bibinfo{person}{Shimon Whiteson}.}
  \bibinfo{year}{2019}\natexlab{}.
\newblock \showarticletitle{Exploration with unreliable intrinsic reward in
  multi-agent reinforcement learning}.
\newblock \bibinfo{journal}{\emph{arXiv preprint arXiv:1906.02138}}
  (\bibinfo{year}{2019}).
\newblock


\bibitem[\protect\citeauthoryear{Burda, Edwards, Pathak, Storkey, Darrell, and
  Efros}{Burda et~al\mbox{.}}{2019a}]%
        {burda2018large}
\bibfield{author}{\bibinfo{person}{Yuri Burda}, \bibinfo{person}{Harri
  Edwards}, \bibinfo{person}{Deepak Pathak}, \bibinfo{person}{Amos Storkey},
  \bibinfo{person}{Trevor Darrell}, {and} \bibinfo{person}{Alexei~A. Efros}.}
  \bibinfo{year}{2019}\natexlab{a}.
\newblock \showarticletitle{Large-Scale Study of Curiosity-Driven Learning}.
\newblock \bibinfo{journal}{\emph{International Conference on Learning
  Representations}} (\bibinfo{year}{2019}).
\newblock


\bibitem[\protect\citeauthoryear{Burda, Edwards, Storkey, and Klimov}{Burda
  et~al\mbox{.}}{2019b}]%
        {burda2018exploration}
\bibfield{author}{\bibinfo{person}{Yuri Burda}, \bibinfo{person}{Harrison
  Edwards}, \bibinfo{person}{Amos Storkey}, {and} \bibinfo{person}{Oleg
  Klimov}.} \bibinfo{year}{2019}\natexlab{b}.
\newblock \showarticletitle{Exploration by random network distillation}.
\newblock \bibinfo{journal}{\emph{International Conference on Learning
  Representations}} (\bibinfo{year}{2019}).
\newblock


\bibitem[\protect\citeauthoryear{Cao, Yu, Ren, and Chen}{Cao
  et~al\mbox{.}}{2013}]%
        {cao2013overview}
\bibfield{author}{\bibinfo{person}{Yongcan Cao}, \bibinfo{person}{Wenwu Yu},
  \bibinfo{person}{Wei Ren}, {and} \bibinfo{person}{Guanrong Chen}.}
  \bibinfo{year}{2013}\natexlab{}.
\newblock \showarticletitle{An overview of recent progress in the study of
  distributed multi-agent coordination}.
\newblock \bibinfo{journal}{\emph{IEEE Transactions on Industrial informatics}}
  \bibinfo{volume}{9}, \bibinfo{number}{1} (\bibinfo{year}{2013}),
  \bibinfo{pages}{427--438}.
\newblock


\bibitem[\protect\citeauthoryear{Chen, Sidor, Abbeel, and Schulman}{Chen
  et~al\mbox{.}}{2017}]%
        {chen2017ucb}
\bibfield{author}{\bibinfo{person}{Richard~Y Chen}, \bibinfo{person}{Szymon
  Sidor}, \bibinfo{person}{Pieter Abbeel}, {and} \bibinfo{person}{John
  Schulman}.} \bibinfo{year}{2017}\natexlab{}.
\newblock \showarticletitle{{UCB} exploration via {Q}-ensembles}.
\newblock \bibinfo{journal}{\emph{arXiv preprint arXiv:1706.01502}}
  (\bibinfo{year}{2017}).
\newblock


\bibitem[\protect\citeauthoryear{Corke, Peterson, and Rus}{Corke
  et~al\mbox{.}}{2005}]%
        {corke2005networked}
\bibfield{author}{\bibinfo{person}{Peter Corke}, \bibinfo{person}{Ron
  Peterson}, {and} \bibinfo{person}{Daniela Rus}.}
  \bibinfo{year}{2005}\natexlab{}.
\newblock \showarticletitle{Networked robots: Flying robot navigation using a
  sensor net}. In \bibinfo{booktitle}{\emph{Robotics research. The eleventh
  international symposium}}. Springer, \bibinfo{pages}{234--243}.
\newblock


\bibitem[\protect\citeauthoryear{Dall'Anese, Zhu, and Giannakis}{Dall'Anese
  et~al\mbox{.}}{2013}]%
        {dall2013distributed}
\bibfield{author}{\bibinfo{person}{Emiliano Dall'Anese}, \bibinfo{person}{Hao
  Zhu}, {and} \bibinfo{person}{Georgios~B Giannakis}.}
  \bibinfo{year}{2013}\natexlab{}.
\newblock \showarticletitle{Distributed Optimal Power Flow for Smart
  Microgrids.}
\newblock \bibinfo{journal}{\emph{IEEE Trans. Smart Grid}} \bibinfo{volume}{4},
  \bibinfo{number}{3} (\bibinfo{year}{2013}), \bibinfo{pages}{1464--1475}.
\newblock


\bibitem[\protect\citeauthoryear{Davis, Lii, and Politis}{Davis
  et~al\mbox{.}}{2011}]%
        {davis2011remarks}
\bibfield{author}{\bibinfo{person}{Richard~A Davis}, \bibinfo{person}{Keh-Shin
  Lii}, {and} \bibinfo{person}{Dimitris~N Politis}.}
  \bibinfo{year}{2011}\natexlab{}.
\newblock \showarticletitle{Remarks on some nonparametric estimates of a
  density function}.
\newblock In \bibinfo{booktitle}{\emph{Selected Works of Murray Rosenblatt}}.
  \bibinfo{publisher}{Springer}, \bibinfo{pages}{95--100}.
\newblock


\bibitem[\protect\citeauthoryear{Dhariwal, Hesse, Klimov, Nichol, Plappert,
  Radford, Schulman, Sidor, Wu, and Zhokhov}{Dhariwal et~al\mbox{.}}{2017}]%
        {baselines}
\bibfield{author}{\bibinfo{person}{Prafulla Dhariwal},
  \bibinfo{person}{Christopher Hesse}, \bibinfo{person}{Oleg Klimov},
  \bibinfo{person}{Alex Nichol}, \bibinfo{person}{Matthias Plappert},
  \bibinfo{person}{Alec Radford}, \bibinfo{person}{John Schulman},
  \bibinfo{person}{Szymon Sidor}, \bibinfo{person}{Yuhuai Wu}, {and}
  \bibinfo{person}{Peter Zhokhov}.} \bibinfo{year}{2017}\natexlab{}.
\newblock \bibinfo{title}{OpenAI Baselines}.
\newblock \bibinfo{howpublished}{\url{https://github.com/openai/baselines}}.
\newblock


\bibitem[\protect\citeauthoryear{Fax and Murray}{Fax and Murray}{2004}]%
        {fax2004information}
\bibfield{author}{\bibinfo{person}{J~Alexander Fax} {and}
  \bibinfo{person}{Richard~M Murray}.} \bibinfo{year}{2004}\natexlab{}.
\newblock \showarticletitle{Information flow and cooperative control of vehicle
  formations}.
\newblock \bibinfo{journal}{\emph{IEEE transactions on automatic control}}
  \bibinfo{volume}{49}, \bibinfo{number}{9} (\bibinfo{year}{2004}),
  \bibinfo{pages}{1465--1476}.
\newblock


\bibitem[\protect\citeauthoryear{Florensa, Held, Geng, and Abbeel}{Florensa
  et~al\mbox{.}}{2018}]%
        {florensa2017automatic}
\bibfield{author}{\bibinfo{person}{Carlos Florensa}, \bibinfo{person}{David
  Held}, \bibinfo{person}{Xinyang Geng}, {and} \bibinfo{person}{Pieter
  Abbeel}.} \bibinfo{year}{2018}\natexlab{}.
\newblock \showarticletitle{Automatic Goal Generation for Reinforcement
  Learning Agents}. In \bibinfo{booktitle}{\emph{Proceedings of the 35th
  International Conference on Machine Learning}}
  \emph{(\bibinfo{series}{Proceedings of Machine Learning Research},
  Vol.~\bibinfo{volume}{80})}. \bibinfo{publisher}{PMLR},
  \bibinfo{pages}{1515--1528}.
\newblock


\bibitem[\protect\citeauthoryear{Florensa, Held, Wulfmeier, Zhang, and
  Abbeel}{Florensa et~al\mbox{.}}{2017}]%
        {florensa2017reverse}
\bibfield{author}{\bibinfo{person}{Carlos Florensa}, \bibinfo{person}{David
  Held}, \bibinfo{person}{Markus Wulfmeier}, \bibinfo{person}{Michael Zhang},
  {and} \bibinfo{person}{Pieter Abbeel}.} \bibinfo{year}{2017}\natexlab{}.
\newblock \showarticletitle{Reverse Curriculum Generation for Reinforcement
  Learning}. In \bibinfo{booktitle}{\emph{Proceedings of the 1st Annual
  Conference on Robot Learning}} \emph{(\bibinfo{series}{Proceedings of Machine
  Learning Research}, Vol.~\bibinfo{volume}{78})}. \bibinfo{publisher}{PMLR},
  \bibinfo{pages}{482--495}.
\newblock


\bibitem[\protect\citeauthoryear{Goyal, Brakel, Fedus, Singhal, Lillicrap,
  Levine, Larochelle, and Bengio}{Goyal et~al\mbox{.}}{2019}]%
        {goyal2018recall}
\bibfield{author}{\bibinfo{person}{Anirudh Goyal}, \bibinfo{person}{Philemon
  Brakel}, \bibinfo{person}{William Fedus}, \bibinfo{person}{Soumye Singhal},
  \bibinfo{person}{Timothy Lillicrap}, \bibinfo{person}{Sergey Levine},
  \bibinfo{person}{Hugo Larochelle}, {and} \bibinfo{person}{Yoshua Bengio}.}
  \bibinfo{year}{2019}\natexlab{}.
\newblock \showarticletitle{Recall traces: Backtracking models for efficient
  reinforcement learning}.
\newblock \bibinfo{journal}{\emph{International Conference on Learning
  Representations}} (\bibinfo{year}{2019}).
\newblock


\bibitem[\protect\citeauthoryear{Iqbal and Sha}{Iqbal and Sha}{2019}]%
        {iqbal2019coordinated}
\bibfield{author}{\bibinfo{person}{Shariq Iqbal} {and} \bibinfo{person}{Fei
  Sha}.} \bibinfo{year}{2019}\natexlab{}.
\newblock \showarticletitle{Coordinated Exploration via Intrinsic Rewards for
  Multi-Agent Reinforcement Learning}.
\newblock \bibinfo{journal}{\emph{arXiv preprint arXiv:1905.12127}}
  (\bibinfo{year}{2019}).
\newblock


\bibitem[\protect\citeauthoryear{Jiang and Lu}{Jiang and Lu}{2020}]%
        {jiang2019generative}
\bibfield{author}{\bibinfo{person}{Jiechuan Jiang} {and}
  \bibinfo{person}{Zongqing Lu}.} \bibinfo{year}{2020}\natexlab{}.
\newblock \showarticletitle{Generative Exploration and Exploitation}. In
  \bibinfo{booktitle}{\emph{Thirty-Fourth AAAI conference on artificial
  intelligence}}.
\newblock


\bibitem[\protect\citeauthoryear{Kingma and Welling}{Kingma and
  Welling}{2014}]%
        {kingma2013auto}
\bibfield{author}{\bibinfo{person}{Diederik~P Kingma} {and}
  \bibinfo{person}{Max Welling}.} \bibinfo{year}{2014}\natexlab{}.
\newblock \showarticletitle{Auto-encoding variational bayes}.
\newblock \bibinfo{journal}{\emph{International Conference on Learning
  Representations}} (\bibinfo{year}{2014}).
\newblock


\bibitem[\protect\citeauthoryear{Kipf and Welling}{Kipf and Welling}{2016}]%
        {kipf2016variational}
\bibfield{author}{\bibinfo{person}{Thomas~N Kipf} {and} \bibinfo{person}{Max
  Welling}.} \bibinfo{year}{2016}\natexlab{}.
\newblock \showarticletitle{Variational Graph Auto-Encoders}.
\newblock \bibinfo{journal}{\emph{NIPS Workshop on Bayesian Deep Learning}}
  (\bibinfo{year}{2016}).
\newblock


\bibitem[\protect\citeauthoryear{Kipf and Welling}{Kipf and Welling}{2017}]%
        {kipf2016semi}
\bibfield{author}{\bibinfo{person}{Thomas~N Kipf} {and} \bibinfo{person}{Max
  Welling}.} \bibinfo{year}{2017}\natexlab{}.
\newblock \showarticletitle{Semi-supervised classification with graph
  convolutional networks}.
\newblock \bibinfo{journal}{\emph{International Conference on Learning
  Representations}} (\bibinfo{year}{2017}).
\newblock


\bibitem[\protect\citeauthoryear{Littman}{Littman}{1994}]%
        {littman1994markov}
\bibfield{author}{\bibinfo{person}{Michael~L Littman}.}
  \bibinfo{year}{1994}\natexlab{}.
\newblock \showarticletitle{Markov games as a framework for multi-agent
  reinforcement learning}.
\newblock In \bibinfo{booktitle}{\emph{Machine learning proceedings 1994}}.
  \bibinfo{publisher}{Elsevier}, \bibinfo{pages}{157--163}.
\newblock


\bibitem[\protect\citeauthoryear{Lowe, Wu, Tamar, Harb, Abbeel, and
  Mordatch}{Lowe et~al\mbox{.}}{2017}]%
        {lowe2017multi}
\bibfield{author}{\bibinfo{person}{Ryan Lowe}, \bibinfo{person}{Yi~I Wu},
  \bibinfo{person}{Aviv Tamar}, \bibinfo{person}{Jean Harb},
  \bibinfo{person}{OpenAI~Pieter Abbeel}, {and} \bibinfo{person}{Igor
  Mordatch}.} \bibinfo{year}{2017}\natexlab{}.
\newblock \showarticletitle{Multi-agent actor-critic for mixed
  cooperative-competitive environments}. In \bibinfo{booktitle}{\emph{Advances
  in neural information processing systems}}. \bibinfo{pages}{6379--6390}.
\newblock


\bibitem[\protect\citeauthoryear{Matignon, Jeanpierre, Mouaddib,
  et~al\mbox{.}}{Matignon et~al\mbox{.}}{2012}]%
        {matignon2012coordinated}
\bibfield{author}{\bibinfo{person}{La{\"e}titia Matignon},
  \bibinfo{person}{Laurent Jeanpierre}, \bibinfo{person}{Abdel-Illah Mouaddib},
  {et~al\mbox{.}}} \bibinfo{year}{2012}\natexlab{}.
\newblock \showarticletitle{Coordinated Multi-Robot Exploration Under
  Communication Constraints Using Decentralized Markov Decision Processes.}. In
  \bibinfo{booktitle}{\emph{AAAI}}.
\newblock


\bibitem[\protect\citeauthoryear{Mnih, Kavukcuoglu, Silver, Rusu, Veness,
  Bellemare, Graves, Riedmiller, Fidjeland, Ostrovski, et~al\mbox{.}}{Mnih
  et~al\mbox{.}}{2015}]%
        {mnih2015human}
\bibfield{author}{\bibinfo{person}{Volodymyr Mnih}, \bibinfo{person}{Koray
  Kavukcuoglu}, \bibinfo{person}{David Silver}, \bibinfo{person}{Andrei~A
  Rusu}, \bibinfo{person}{Joel Veness}, \bibinfo{person}{Marc~G Bellemare},
  \bibinfo{person}{Alex Graves}, \bibinfo{person}{Martin Riedmiller},
  \bibinfo{person}{Andreas~K Fidjeland}, \bibinfo{person}{Georg Ostrovski},
  {et~al\mbox{.}}} \bibinfo{year}{2015}\natexlab{}.
\newblock \showarticletitle{Human-level control through deep reinforcement
  learning}.
\newblock \bibinfo{journal}{\emph{Nature}} \bibinfo{volume}{518},
  \bibinfo{number}{7540} (\bibinfo{year}{2015}), \bibinfo{pages}{529}.
\newblock


\bibitem[\protect\citeauthoryear{Nair, McGrew, Andrychowicz, Zaremba, and
  Abbeel}{Nair et~al\mbox{.}}{2018}]%
        {nair2018overcoming}
\bibfield{author}{\bibinfo{person}{Ashvin Nair}, \bibinfo{person}{Bob McGrew},
  \bibinfo{person}{Marcin Andrychowicz}, \bibinfo{person}{Wojciech Zaremba},
  {and} \bibinfo{person}{Pieter Abbeel}.} \bibinfo{year}{2018}\natexlab{}.
\newblock \showarticletitle{Overcoming exploration in reinforcement learning
  with demonstrations}. In \bibinfo{booktitle}{\emph{2018 IEEE International
  Conference on Robotics and Automation (ICRA)}}. IEEE,
  \bibinfo{pages}{6292--6299}.
\newblock


\bibitem[\protect\citeauthoryear{Ostrovski, Bellemare, van~den Oord, and
  Munos}{Ostrovski et~al\mbox{.}}{2017}]%
        {ostrovski2017count}
\bibfield{author}{\bibinfo{person}{Georg Ostrovski}, \bibinfo{person}{Marc~G
  Bellemare}, \bibinfo{person}{A{\"a}ron van~den Oord}, {and}
  \bibinfo{person}{R{\'e}mi Munos}.} \bibinfo{year}{2017}\natexlab{}.
\newblock \showarticletitle{Count-based exploration with neural density
  models}. In \bibinfo{booktitle}{\emph{Proceedings of the 34th International
  Conference on Machine Learning-Volume 70}}. JMLR. org,
  \bibinfo{pages}{2721--2730}.
\newblock


\bibitem[\protect\citeauthoryear{Pathak, Agrawal, Efros, and Darrell}{Pathak
  et~al\mbox{.}}{2017}]%
        {pathak2017curiosity}
\bibfield{author}{\bibinfo{person}{Deepak Pathak}, \bibinfo{person}{Pulkit
  Agrawal}, \bibinfo{person}{Alexei~A Efros}, {and} \bibinfo{person}{Trevor
  Darrell}.} \bibinfo{year}{2017}\natexlab{}.
\newblock \showarticletitle{Curiosity-driven exploration by self-supervised
  prediction}. In \bibinfo{booktitle}{\emph{Proceedings of the IEEE Conference
  on Computer Vision and Pattern Recognition Workshops}}.
  \bibinfo{pages}{16--17}.
\newblock


\bibitem[\protect\citeauthoryear{Resnick, Raileanu, Kapoor, Peysakhovich, Cho,
  and Bruna}{Resnick et~al\mbox{.}}{2018}]%
        {resnick2018backplay}
\bibfield{author}{\bibinfo{person}{Cinjon Resnick}, \bibinfo{person}{Roberta
  Raileanu}, \bibinfo{person}{Sanyam Kapoor}, \bibinfo{person}{Alexander
  Peysakhovich}, \bibinfo{person}{Kyunghyun Cho}, {and} \bibinfo{person}{Joan
  Bruna}.} \bibinfo{year}{2018}\natexlab{}.
\newblock \showarticletitle{Backplay:" Man muss immer umkehren"}.
\newblock \bibinfo{journal}{\emph{arXiv preprint arXiv:1807.06919}}
  (\bibinfo{year}{2018}).
\newblock


\bibitem[\protect\citeauthoryear{Samvelyan, Rashid, Schroeder~de Witt,
  Farquhar, Nardelli, Rudner, Hung, Torr, Foerster, and Whiteson}{Samvelyan
  et~al\mbox{.}}{2019}]%
        {smac}
\bibfield{author}{\bibinfo{person}{Mikayel Samvelyan}, \bibinfo{person}{Tabish
  Rashid}, \bibinfo{person}{Christian Schroeder~de Witt},
  \bibinfo{person}{Gregory Farquhar}, \bibinfo{person}{Nantas Nardelli},
  \bibinfo{person}{Tim G.~J. Rudner}, \bibinfo{person}{Chia-Man Hung},
  \bibinfo{person}{Philip H.~S. Torr}, \bibinfo{person}{Jakob Foerster}, {and}
  \bibinfo{person}{Shimon Whiteson}.} \bibinfo{year}{2019}\natexlab{}.
\newblock \showarticletitle{The StarCraft Multi-Agent Challenge}. In
  \bibinfo{booktitle}{\emph{Proceedings of the 18th International Conference on
  Autonomous Agents and MultiAgent Systems}} \emph{(\bibinfo{series}{AAMAS
  '19})}. \bibinfo{publisher}{International Foundation for Autonomous Agents
  and Multiagent Systems}, \bibinfo{pages}{2186–2188}.
\newblock


\bibitem[\protect\citeauthoryear{Schaul, Quan, Antonoglou, and Silver}{Schaul
  et~al\mbox{.}}{2016}]%
        {schaul2015prioritized}
\bibfield{author}{\bibinfo{person}{Tom Schaul}, \bibinfo{person}{John Quan},
  \bibinfo{person}{Ioannis Antonoglou}, {and} \bibinfo{person}{David Silver}.}
  \bibinfo{year}{2016}\natexlab{}.
\newblock \showarticletitle{Prioritized experience replay}.
\newblock \bibinfo{journal}{\emph{International Conference on Learning
  Representations}} (\bibinfo{year}{2016}).
\newblock


\bibitem[\protect\citeauthoryear{Silver, Huang, Maddison, Guez, Sifre, Van
  Den~Driessche, Schrittwieser, Antonoglou, Panneershelvam, Lanctot,
  et~al\mbox{.}}{Silver et~al\mbox{.}}{2016}]%
        {silver2016mastering}
\bibfield{author}{\bibinfo{person}{David Silver}, \bibinfo{person}{Aja Huang},
  \bibinfo{person}{Chris~J Maddison}, \bibinfo{person}{Arthur Guez},
  \bibinfo{person}{Laurent Sifre}, \bibinfo{person}{George Van Den~Driessche},
  \bibinfo{person}{Julian Schrittwieser}, \bibinfo{person}{Ioannis Antonoglou},
  \bibinfo{person}{Veda Panneershelvam}, \bibinfo{person}{Marc Lanctot},
  {et~al\mbox{.}}} \bibinfo{year}{2016}\natexlab{}.
\newblock \showarticletitle{Mastering the game of Go with deep neural networks
  and tree search}.
\newblock \bibinfo{journal}{\emph{Nature}} \bibinfo{volume}{529},
  \bibinfo{number}{7587} (\bibinfo{year}{2016}), \bibinfo{pages}{484}.
\newblock


\bibitem[\protect\citeauthoryear{Silver, Schrittwieser, Simonyan, Antonoglou,
  Huang, Guez, Hubert, Baker, Lai, Bolton, et~al\mbox{.}}{Silver
  et~al\mbox{.}}{2017}]%
        {silver2017mastering}
\bibfield{author}{\bibinfo{person}{David Silver}, \bibinfo{person}{Julian
  Schrittwieser}, \bibinfo{person}{Karen Simonyan}, \bibinfo{person}{Ioannis
  Antonoglou}, \bibinfo{person}{Aja Huang}, \bibinfo{person}{Arthur Guez},
  \bibinfo{person}{Thomas Hubert}, \bibinfo{person}{Lucas Baker},
  \bibinfo{person}{Matthew Lai}, \bibinfo{person}{Adrian Bolton},
  {et~al\mbox{.}}} \bibinfo{year}{2017}\natexlab{}.
\newblock \showarticletitle{Mastering the game of go without human knowledge}.
\newblock \bibinfo{journal}{\emph{Nature}} \bibinfo{volume}{550},
  \bibinfo{number}{7676} (\bibinfo{year}{2017}), \bibinfo{pages}{354}.
\newblock


\bibitem[\protect\citeauthoryear{Tang, Houthooft, Foote, Stooke, Chen, Duan,
  Schulman, DeTurck, and Abbeel}{Tang et~al\mbox{.}}{2017}]%
        {tang2017exploration}
\bibfield{author}{\bibinfo{person}{Haoran Tang}, \bibinfo{person}{Rein
  Houthooft}, \bibinfo{person}{Davis Foote}, \bibinfo{person}{Adam Stooke},
  \bibinfo{person}{OpenAI~Xi Chen}, \bibinfo{person}{Yan Duan},
  \bibinfo{person}{John Schulman}, \bibinfo{person}{Filip DeTurck}, {and}
  \bibinfo{person}{Pieter Abbeel}.} \bibinfo{year}{2017}\natexlab{}.
\newblock \showarticletitle{\#{Exploration}: A study of count-based exploration
  for deep reinforcement learning}. In \bibinfo{booktitle}{\emph{Advances in
  neural information processing systems}}. \bibinfo{pages}{2753--2762}.
\newblock


\bibitem[\protect\citeauthoryear{Vaswani, Shazeer, Parmar, Uszkoreit, Jones,
  Gomez, Kaiser, and Polosukhin}{Vaswani et~al\mbox{.}}{2017}]%
        {vaswani2017attention}
\bibfield{author}{\bibinfo{person}{Ashish Vaswani}, \bibinfo{person}{Noam
  Shazeer}, \bibinfo{person}{Niki Parmar}, \bibinfo{person}{Jakob Uszkoreit},
  \bibinfo{person}{Llion Jones}, \bibinfo{person}{Aidan~N Gomez},
  \bibinfo{person}{{\L}ukasz Kaiser}, {and} \bibinfo{person}{Illia
  Polosukhin}.} \bibinfo{year}{2017}\natexlab{}.
\newblock \showarticletitle{Attention is all you need}. In
  \bibinfo{booktitle}{\emph{Advances in neural information processing
  systems}}. \bibinfo{pages}{5998--6008}.
\newblock


\bibitem[\protect\citeauthoryear{Veli{\v{c}}kovi{\'{c}}, Cucurull, Casanova,
  Romero, Li{\`{o}}, and Bengio}{Veli{\v{c}}kovi{\'{c}} et~al\mbox{.}}{2018}]%
        {velivckovic2017graph}
\bibfield{author}{\bibinfo{person}{Petar Veli{\v{c}}kovi{\'{c}}},
  \bibinfo{person}{Guillem Cucurull}, \bibinfo{person}{Arantxa Casanova},
  \bibinfo{person}{Adriana Romero}, \bibinfo{person}{Pietro Li{\`{o}}}, {and}
  \bibinfo{person}{Yoshua Bengio}.} \bibinfo{year}{2018}\natexlab{}.
\newblock \showarticletitle{Graph Attention Networks}.
\newblock \bibinfo{journal}{\emph{International Conference on Learning
  Representations}} (\bibinfo{year}{2018}).
\newblock


\bibitem[\protect\citeauthoryear{Wang, Wang, Wu, and Zhang}{Wang
  et~al\mbox{.}}{2020}]%
        {wang2019influence}
\bibfield{author}{\bibinfo{person}{Tonghan Wang}, \bibinfo{person}{Jianhao
  Wang}, \bibinfo{person}{Yi Wu}, {and} \bibinfo{person}{Chongjie Zhang}.}
  \bibinfo{year}{2020}\natexlab{}.
\newblock \showarticletitle{Influence-Based Multi-Agent Exploration}.
\newblock \bibinfo{journal}{\emph{International Conference on Learning
  Representations}} (\bibinfo{year}{2020}).
\newblock


\bibitem[\protect\citeauthoryear{Ying and Dayong}{Ying and Dayong}{2005}]%
        {ying2005multi}
\bibfield{author}{\bibinfo{person}{Wang Ying} {and} \bibinfo{person}{Sang
  Dayong}.} \bibinfo{year}{2005}\natexlab{}.
\newblock \showarticletitle{Multi-agent framework for third party logistics in
  E-commerce}.
\newblock \bibinfo{journal}{\emph{Expert Systems with Applications}}
  \bibinfo{volume}{29}, \bibinfo{number}{2} (\bibinfo{year}{2005}),
  \bibinfo{pages}{431--436}.
\newblock


\end{thebibliography}

\newpage


\section*{Relational Representation for Multi-Agent Exploration Algorithm}
For completeness, we provide the REMAX algorithm below.
\begin{algorithm}[h!]
\caption{Relational Representation for Multi-Agent Exploration}
\begin{algorithmic}
\STATE Initialize an MARL model (\emph{e.g.,} MADDPG) for $N$ agents
\STATE Initialize state buffer $\mathcal{D}$
\FOR {episode $= 1$ to $M$}
\STATE Update the MARL model
\STATE Store states in $\mathcal{D}$
\IF {episode$\%N_s=0$}
\STATE Initialize VGAE and a surrogate model
\STATE Train VGAE and the surrogate model together 
\STATE using $\mathcal{D}$ (using Equation (1) and (2))  
\STATE Find $\mathbf{z^*}$ maximizing the surrogate model (using
\STATE Equation (3))
\STATE Reconstruct states $\mathbf{s^*}$ from $\mathbf{z^*}$ using VGAE
\STATE Clear $\mathcal{D}$  
\ENDIF
\STATE Episode starts from $\mathbf{s^*}$ in a certain probability
\ENDFOR
\end{algorithmic}
\end{algorithm}

\section*{Cooperative Navigation}

\subsection*{Learning Curves}

\begin{figure}[h!]
    \centering
    \includegraphics[scale = 0.36]{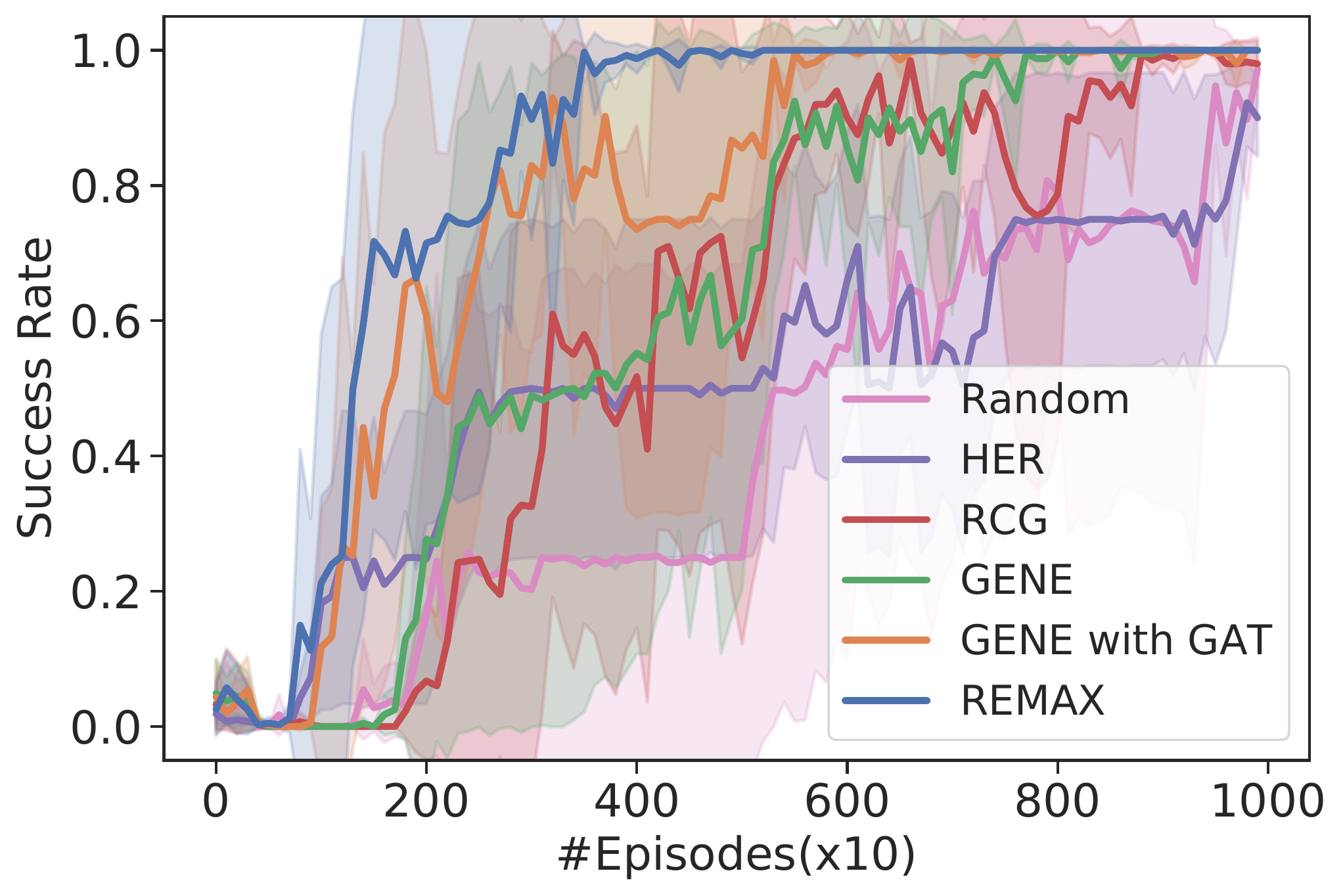}
    \caption{Success rates during training in cooperative navigation with 2 agents.}
    \label{fig:learning_curve}
\end{figure}

Figure \ref{fig:learning_curve} shows the improvement in success rates when training the MARL model with the proposed and baseline exploration methods in a cooperative navigation with two agents. The success rate is defined as the rate of episodes in which ten consecutive successes are achieved, starting from the initial states provided by the environment. The figure shows that all the exploration methods improve the success rates with the training's progress. Among them, the proposed REMAX induces the fastest convergence to a success rate of 1. The figure also shows that the GAT module added to GENE improves the performance, and using both GAT and the surrogate model (REMAX) further enhances the performance.

\subsection*{Choosing the Hyper-Parameter $\lambda$}

\begin{figure}[h!]
    \centering
    \includegraphics[scale = 0.23]{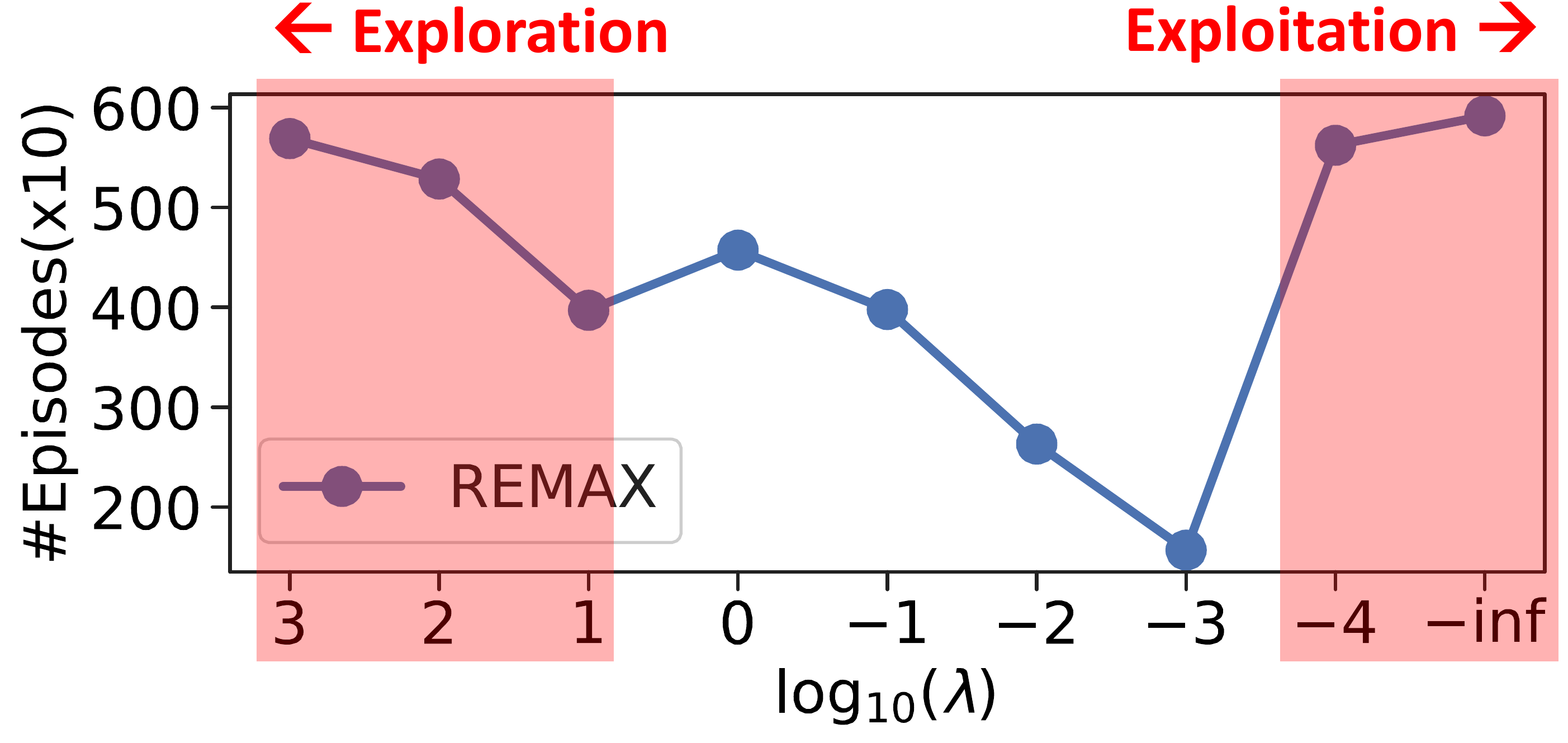}
    \caption{Performance changes according to $\lambda$ values.}
    \label{fig:lambda}
\end{figure}

Figure \ref{fig:lambda} shows how the value of parameter $\lambda$, which is used to define the surrogate output in Equation 1, affects the performance (number of training episodes). When $\lambda$ is small (the left part of Figure 9), TD error contributes more to the surrogate output, thus boosting exploration. On the other hand, when $\lambda$ is large (the right part of Figure 9), the state action value $Q$ contributes more to the surrogate output, thus encouraging exploitation. As shown in Figure 9, when the exploration and exploitation are well balanced by the properly selected $\lambda$ value, the proposed method achieves the best performance (i.e., the lowest number of episodes required). We choose to use $\lambda=10^{-3}$, as it requires fewer training episodes than other values in the figure.  It might be possible to improve the performance by optimally scheduling the variation of $\lambda$.

\begin{figure*}[t]
\begin{minipage}[c]{.153\textwidth}
  \centering
  \includegraphics[width=0.9\textwidth]{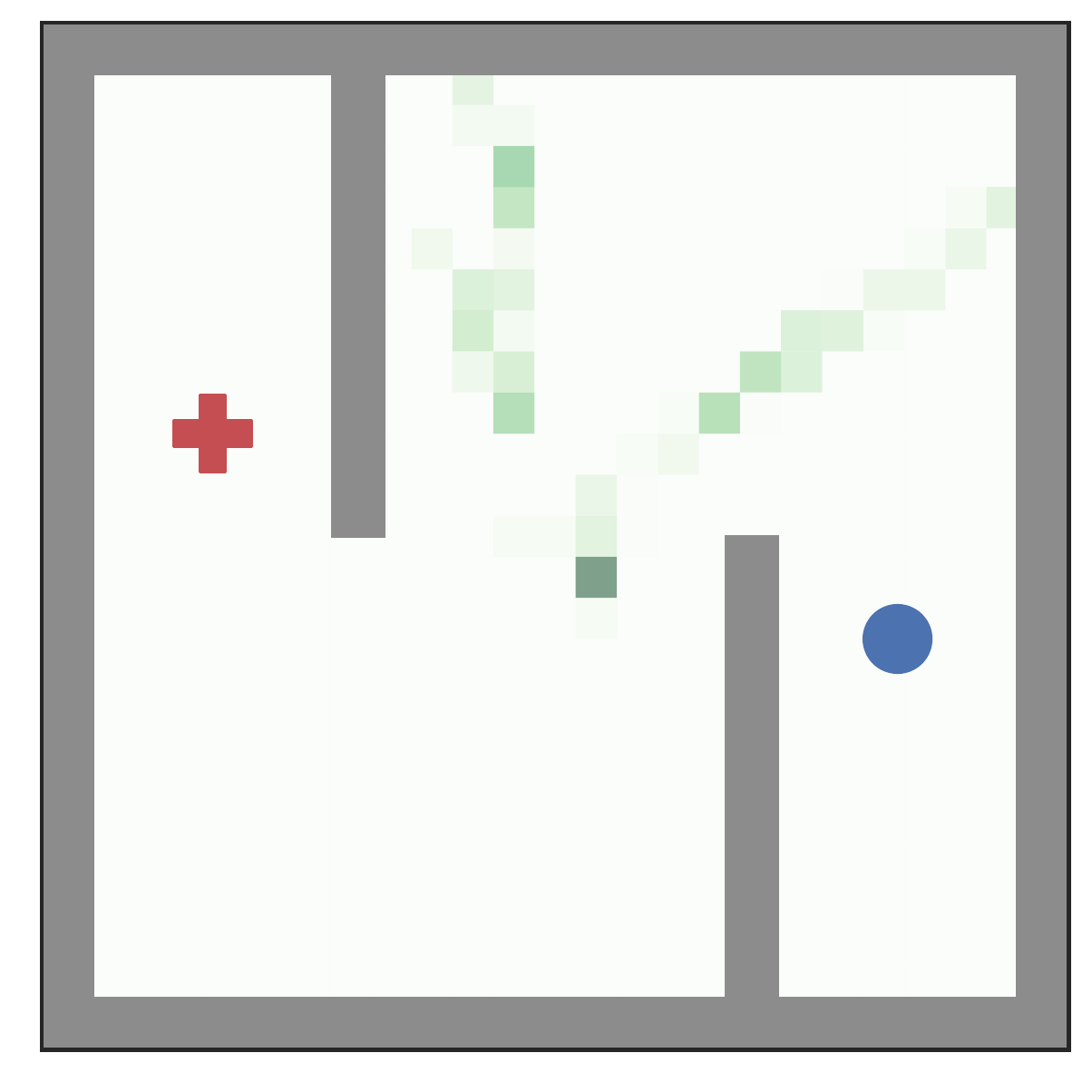}
\end{minipage}
\hspace{0.0cm}
\begin{minipage}[c]{.153\textwidth}
  \centering
  \includegraphics[width=0.9\textwidth]{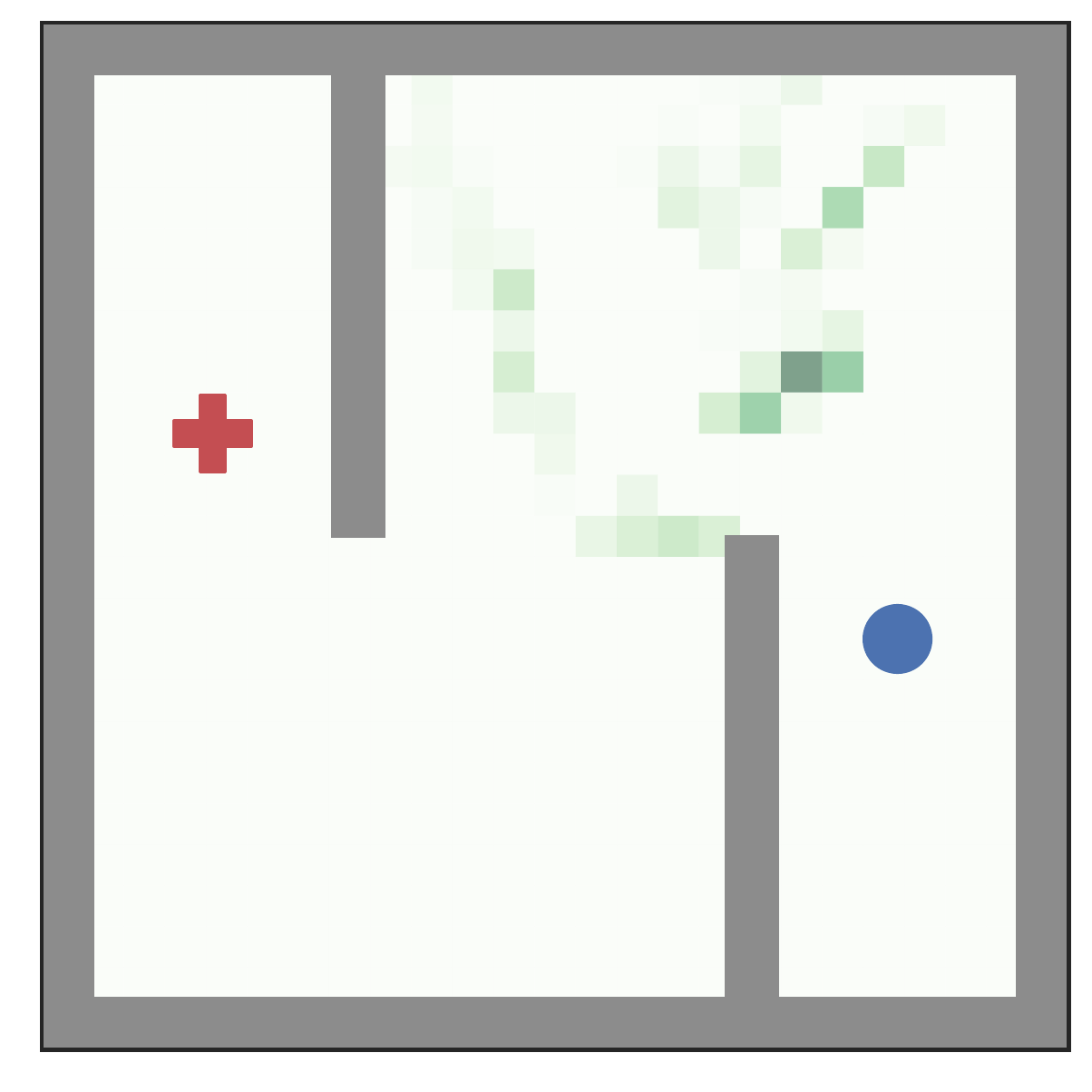}
\end{minipage}
\hspace{0.0cm}
\begin{minipage}[c]{.153\textwidth}
  \centering
  \includegraphics[width=0.9\textwidth]{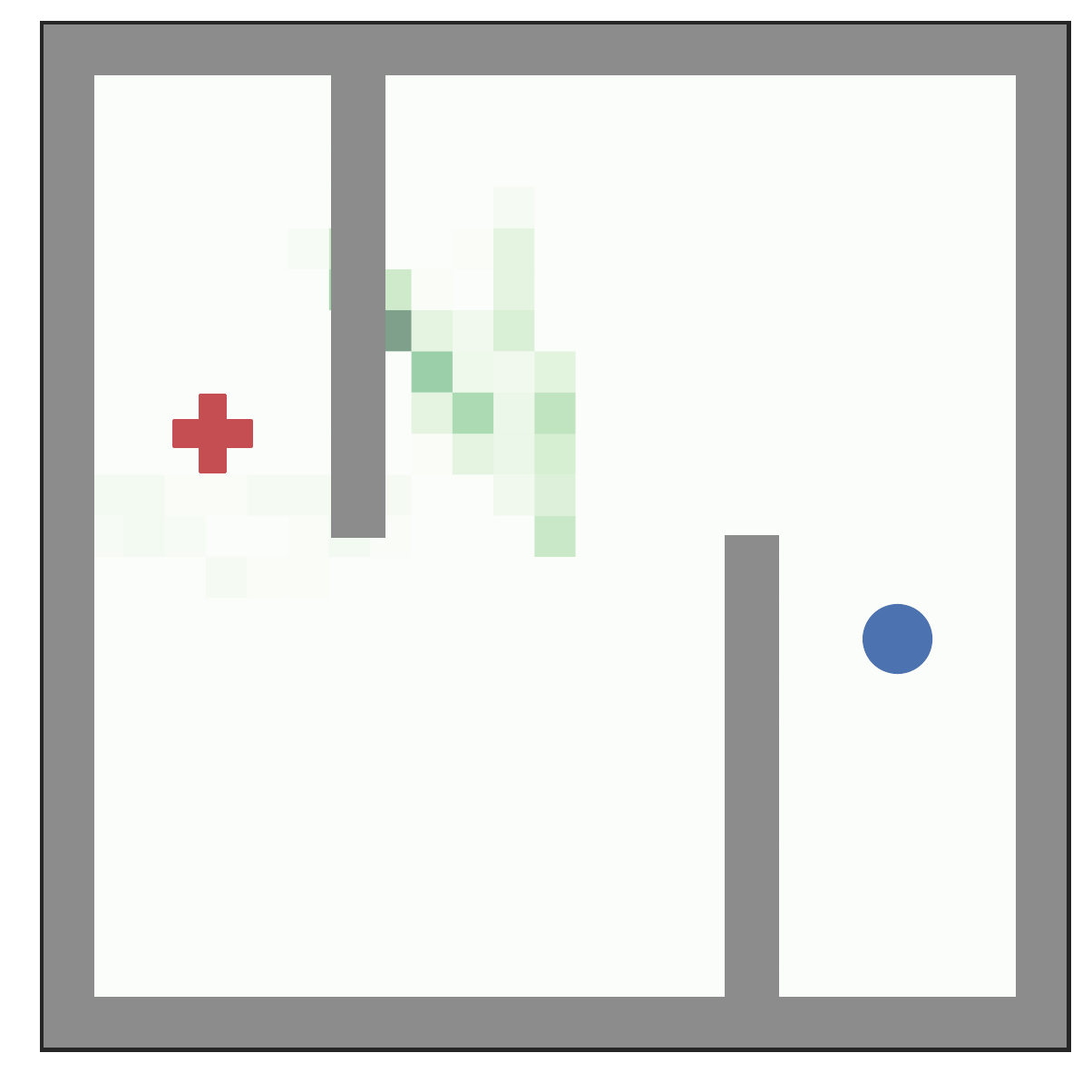}
\end{minipage}
\hspace{0.0cm}
\begin{minipage}[c]{.153\textwidth}
  \centering
  \includegraphics[width=0.9\textwidth]{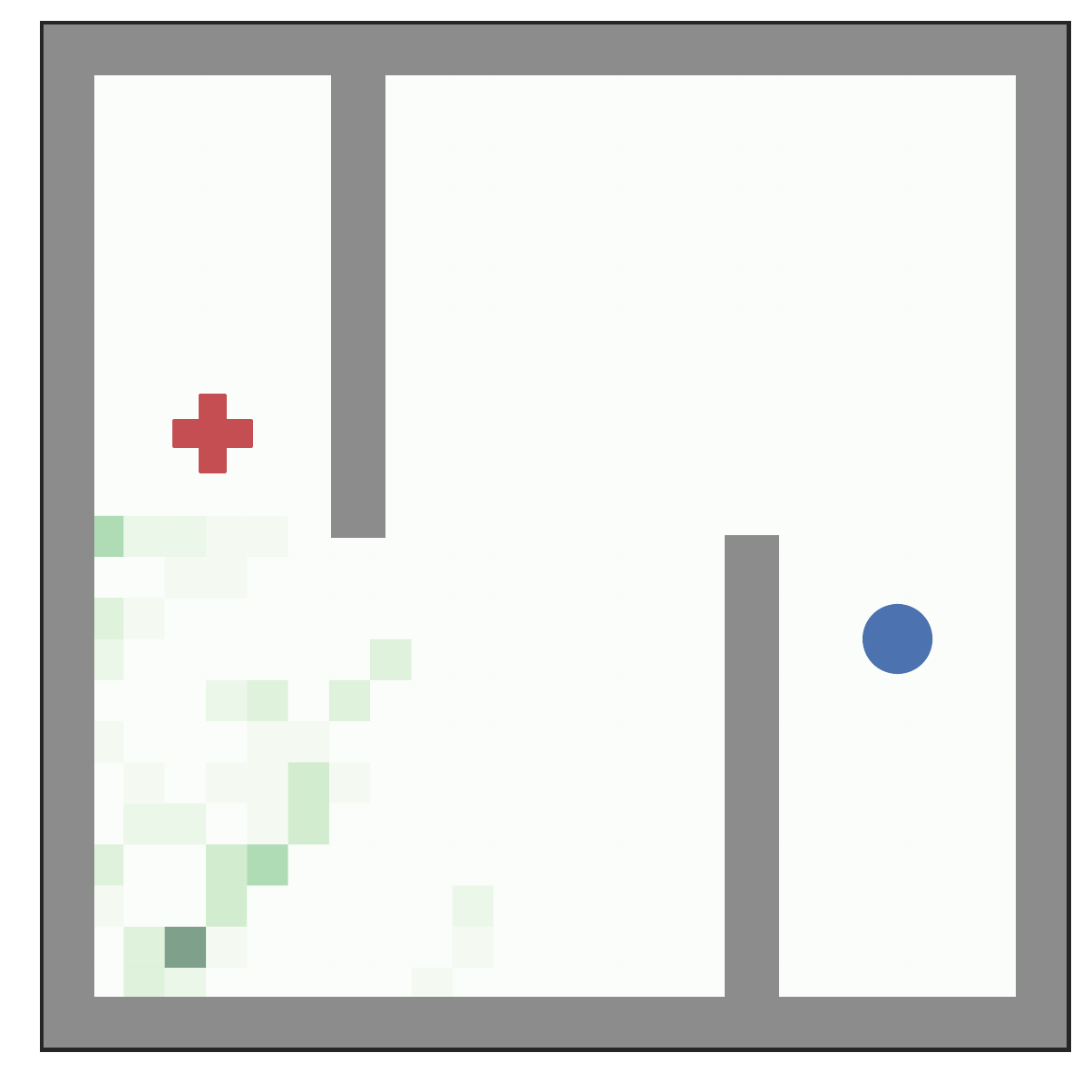}
\end{minipage}
\hspace{0.0cm}
\begin{minipage}[c]{.153\textwidth}
  \centering
  \includegraphics[width=0.9\textwidth]{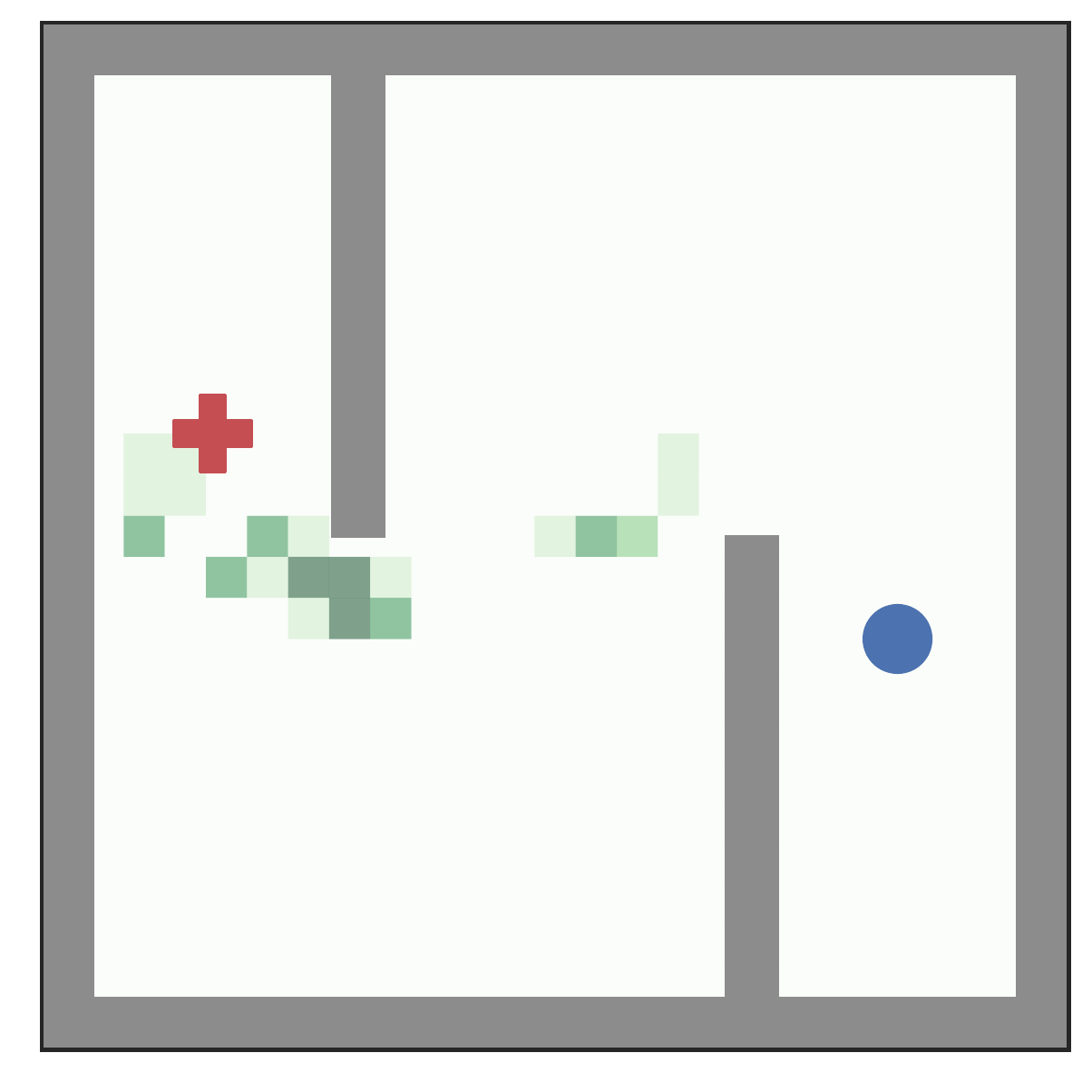}
\end{minipage}
\hspace{0.0cm}
\begin{minipage}[c]{.153\textwidth}
  \centering
  \includegraphics[width=0.9\textwidth]{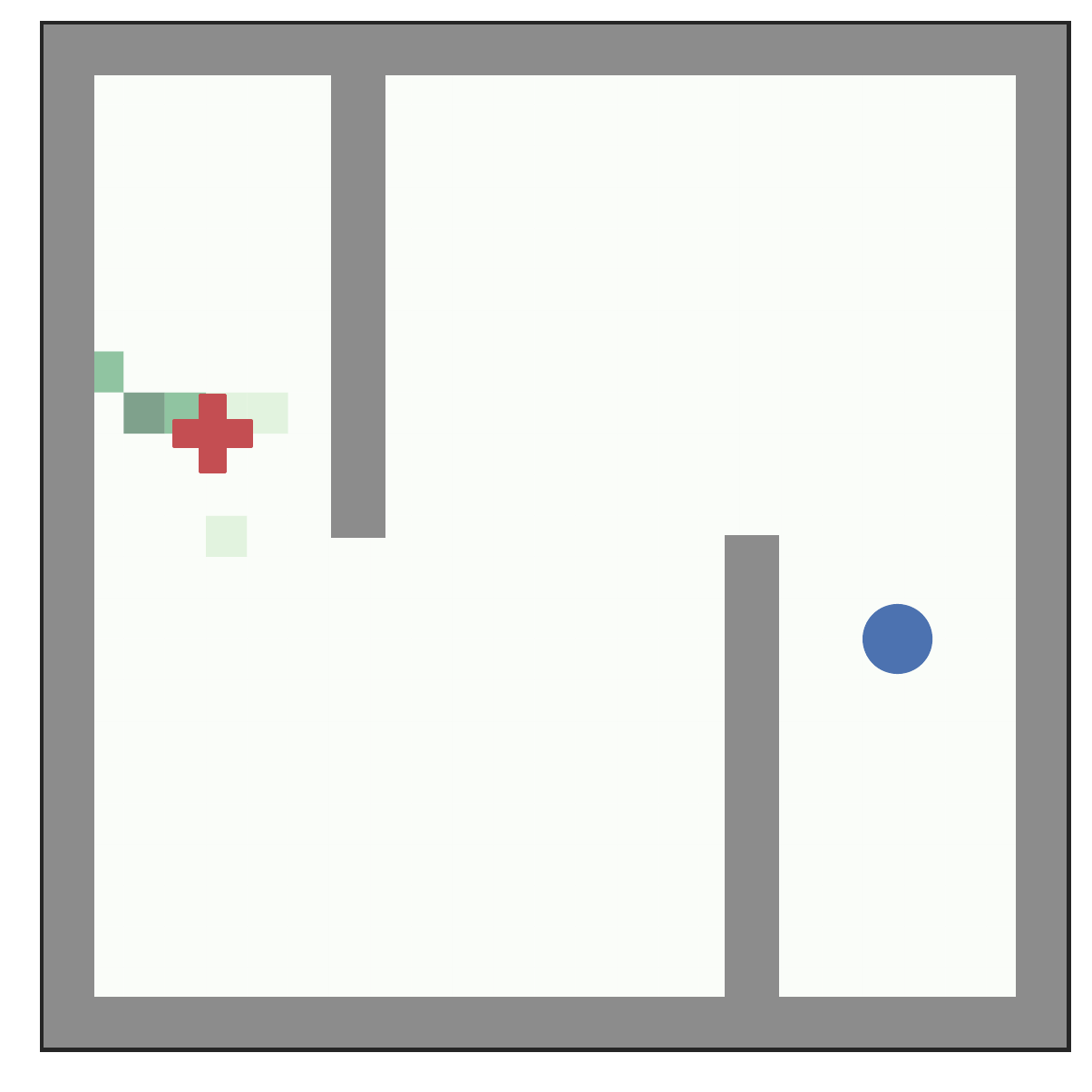}
\end{minipage}

\begin{minipage}[c]{.153\textwidth}
  \centering
  \includegraphics[width=0.9\textwidth]{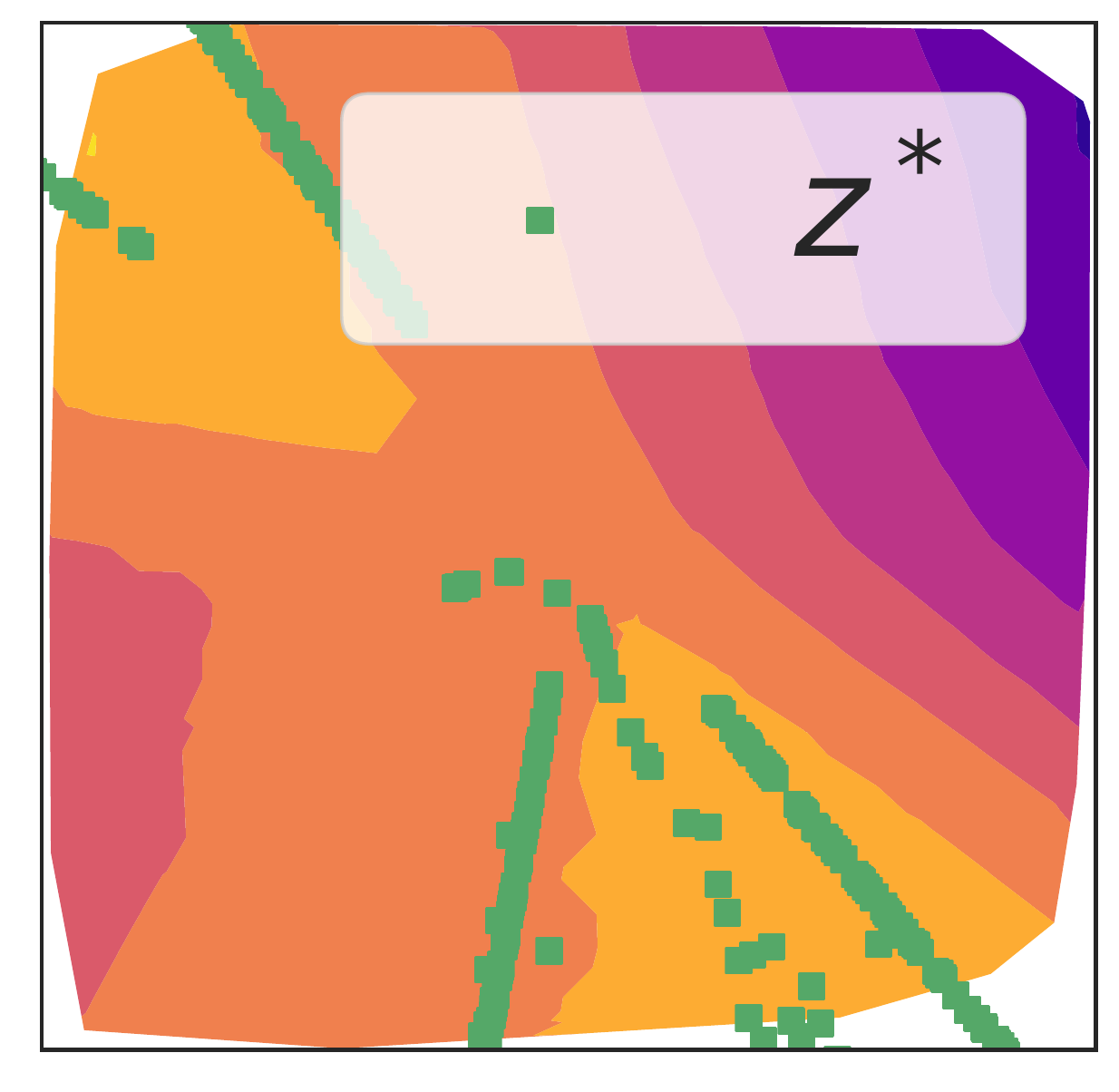}
\end{minipage}
\hspace{0.0cm}
\begin{minipage}[c]{.153\textwidth}
  \centering
  \includegraphics[width=0.9\textwidth]{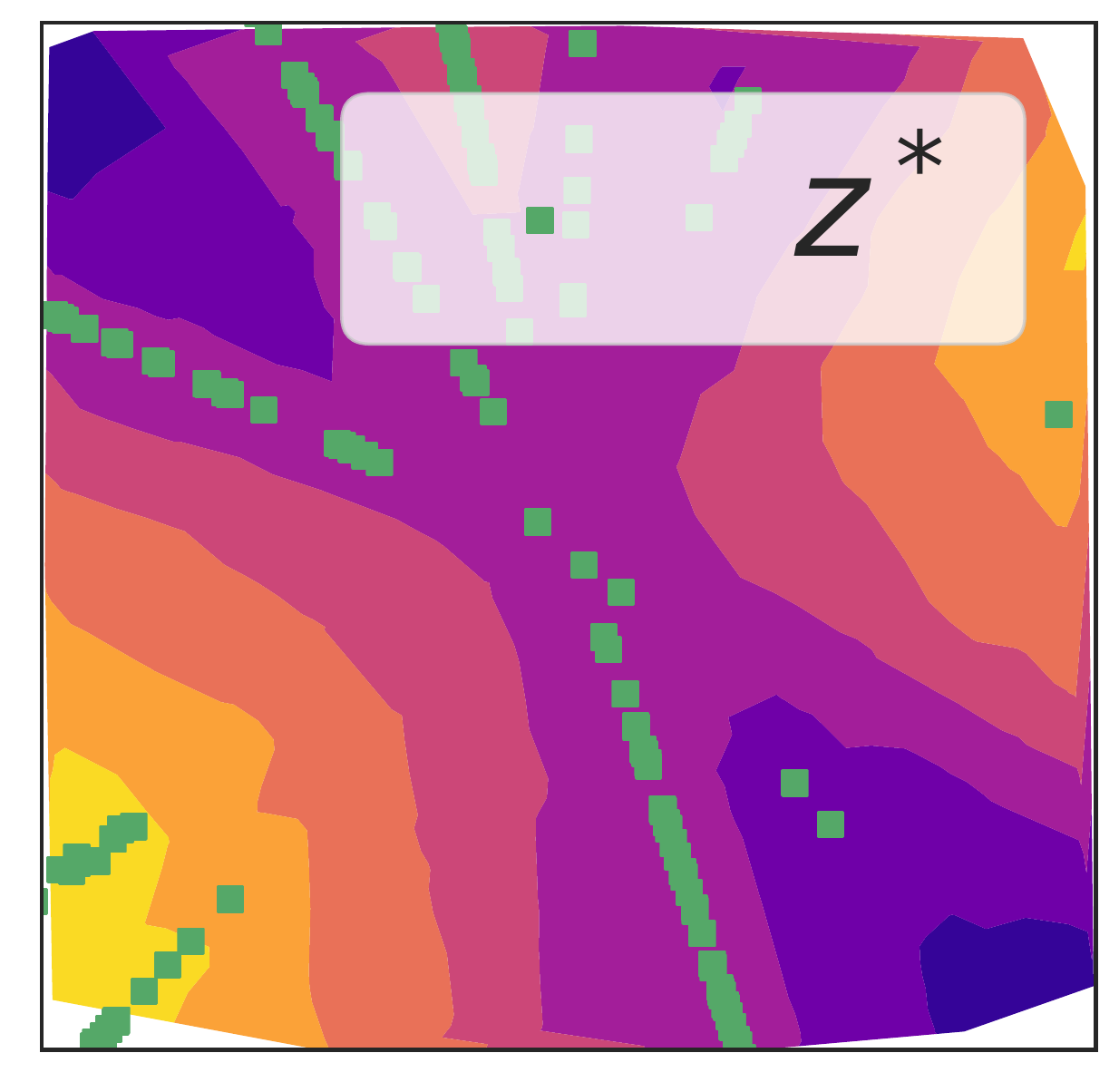}
\end{minipage}
\hspace{0.0cm}
\begin{minipage}[c]{.153\textwidth}
  \centering
  \includegraphics[width=0.9\textwidth]{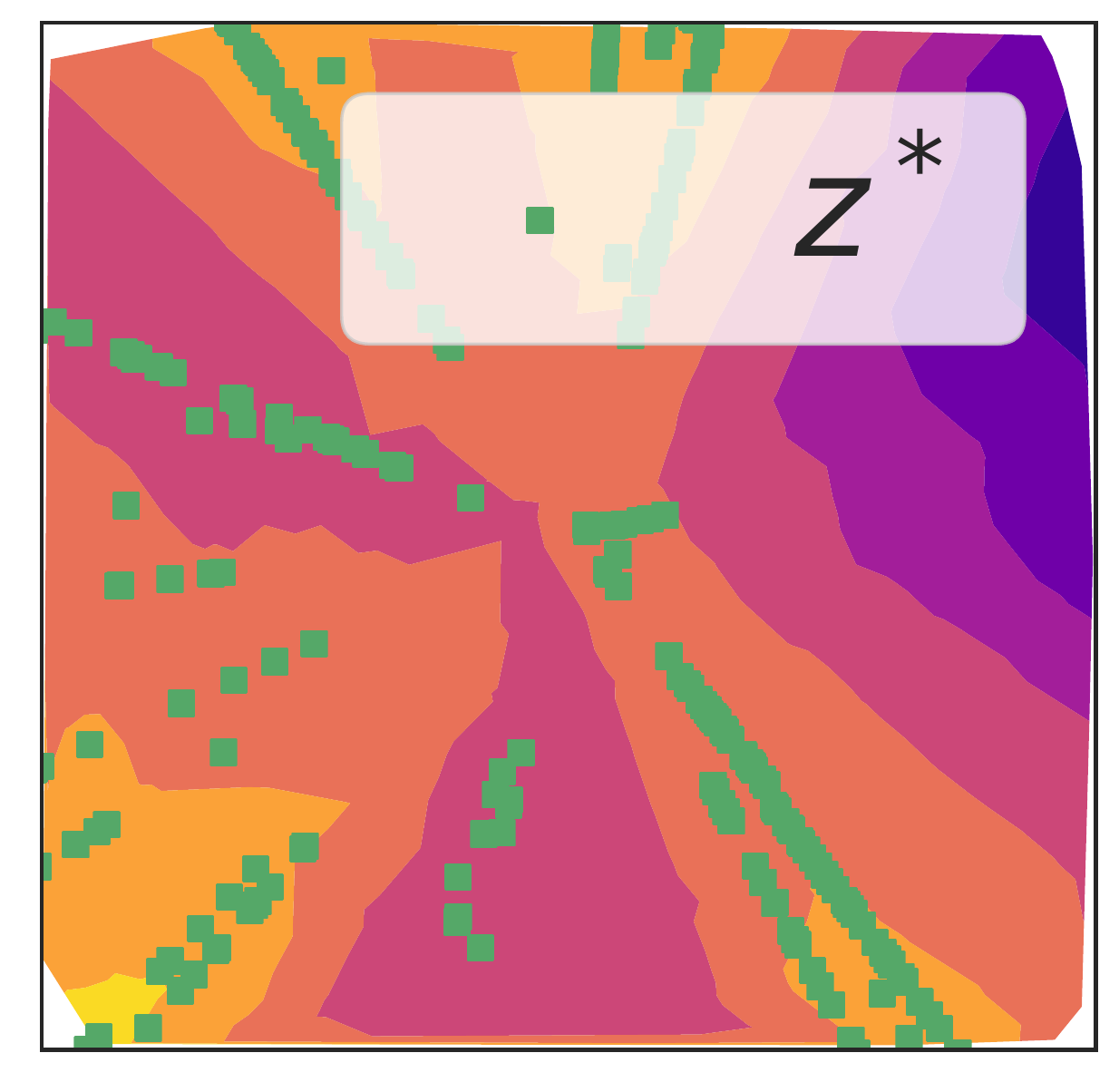}
\end{minipage}
\hspace{0.0cm}
\begin{minipage}[c]{.153\textwidth}
  \centering
  \includegraphics[width=0.9\textwidth]{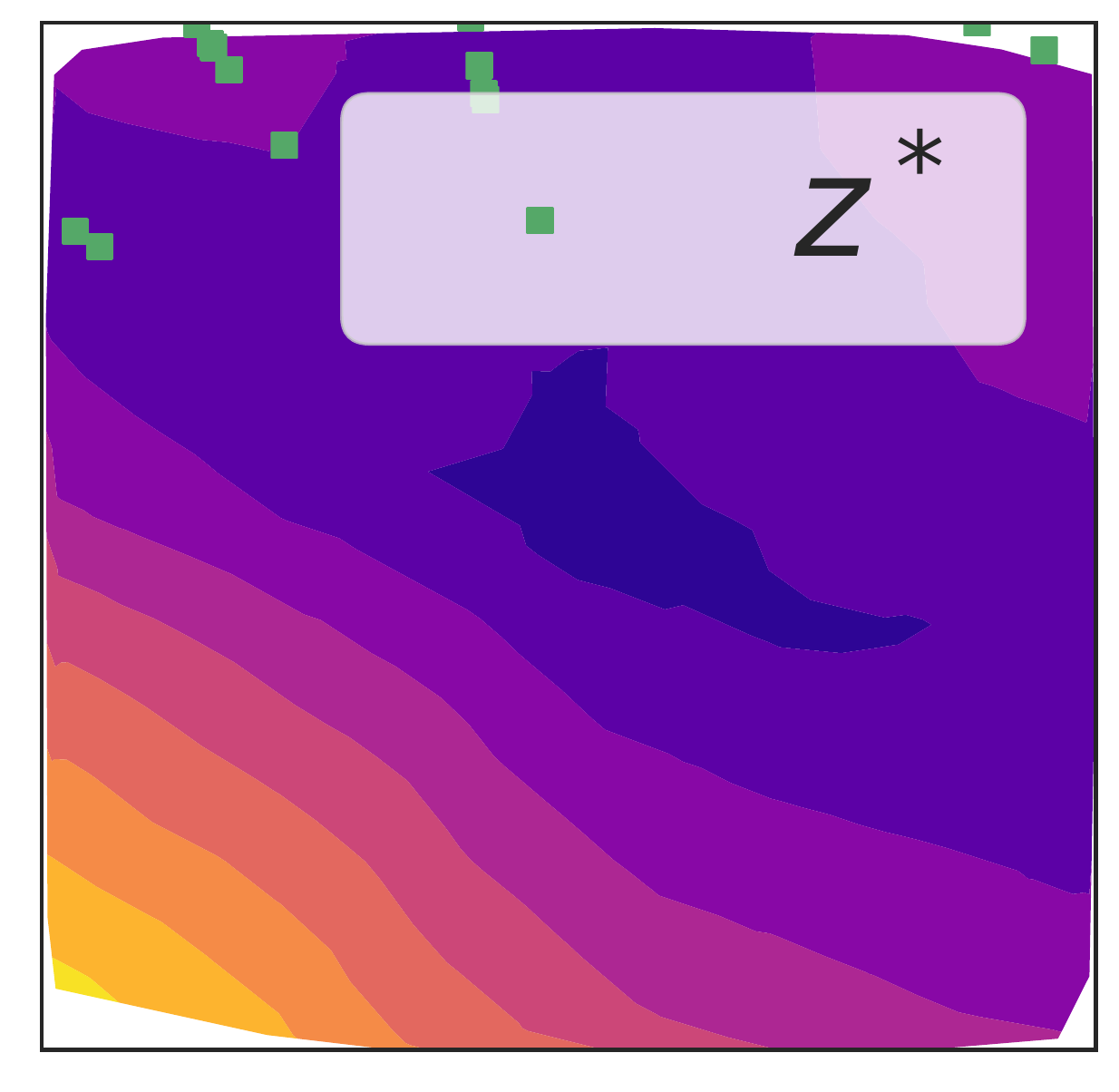}
\end{minipage}
\hspace{0.0cm}
\begin{minipage}[c]{.153\textwidth}
  \centering
  \includegraphics[width=0.9\textwidth]{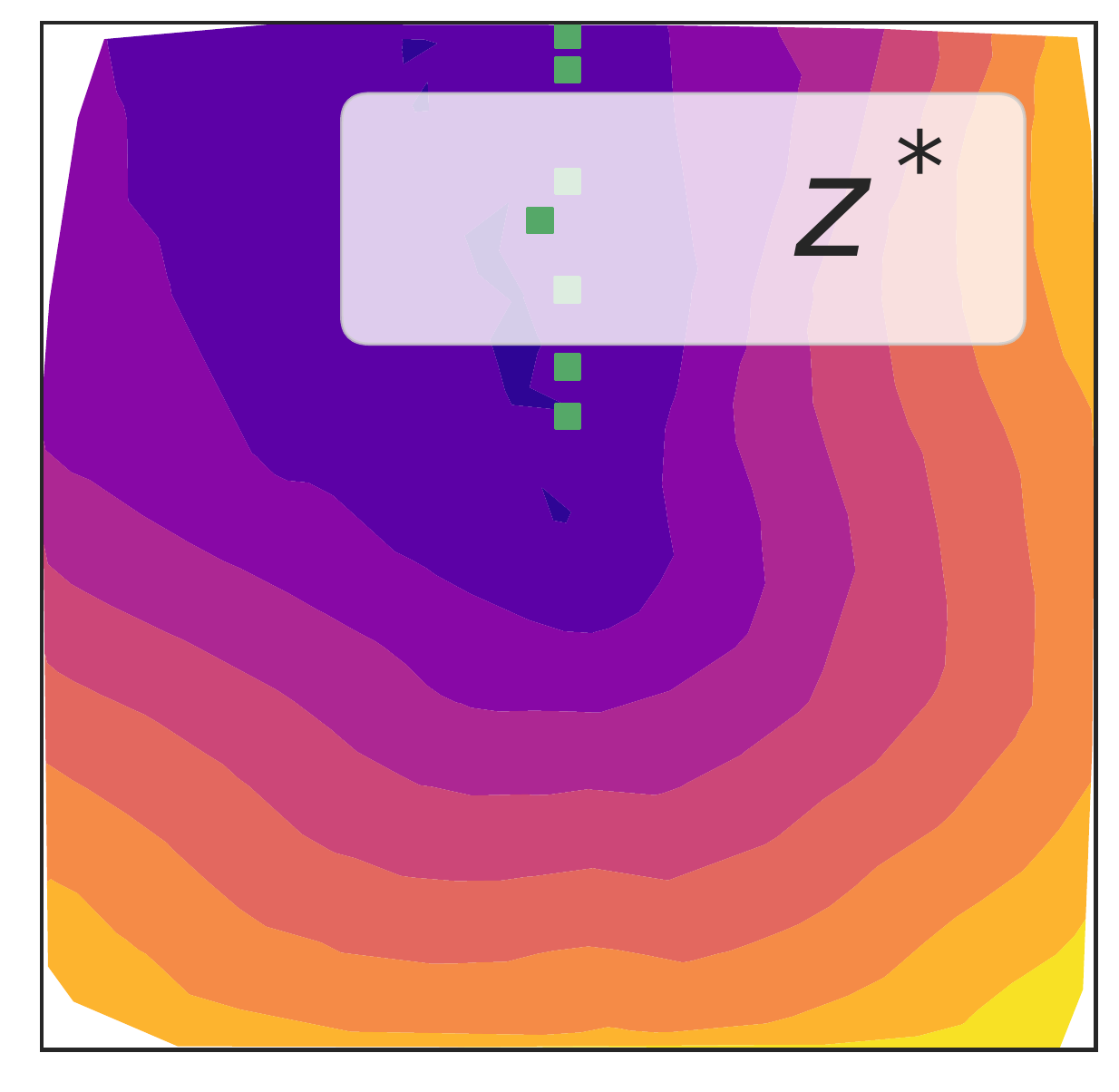}
\end{minipage}
\hspace{0.0cm}
\begin{minipage}[c]{.153\textwidth}
  \centering
  \includegraphics[width=0.9\textwidth]{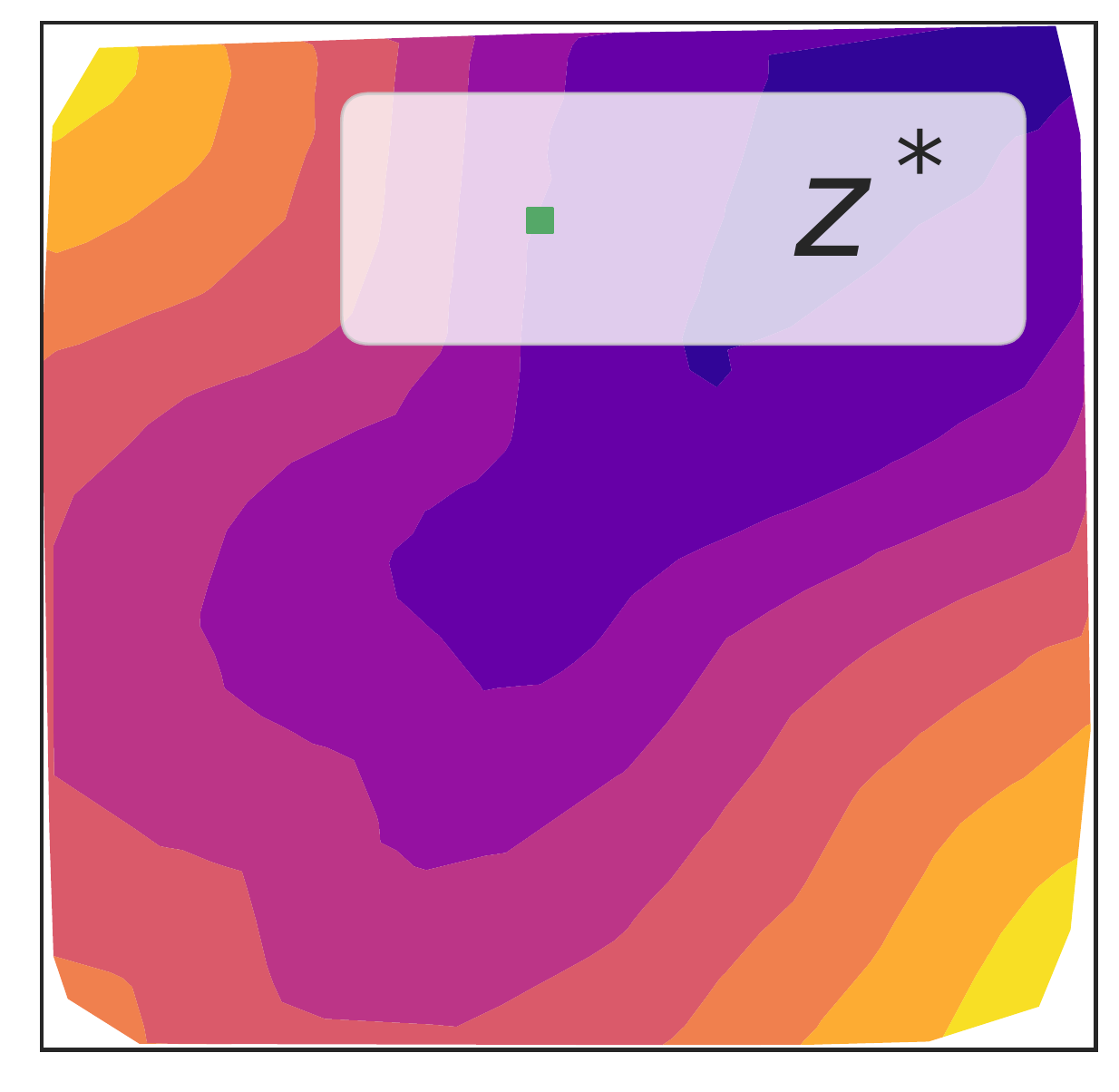}
\end{minipage}

\begin{minipage}[c]{.99\textwidth}
  \centering
  \includegraphics[width=0.92\textwidth]{method_figure_0512_time.pdf}
\end{minipage}
\caption{Generated states and surrogate models with two-dimensional latent space in maze.}
\label{fig:fig_maze_2d}
\end{figure*}

\section*{Maze}

\subsection*{Analysis of Generating States with 2-D Surrogate Models}

The first-row plots in Figure \ref{fig:fig_maze_2d} represent the distribution of generated states $\mathbf{s^*}$ as green dots, and the second-row plots show how the surrogate model $f_\psi(z)$ and the optimized latent vectors $\mathbf{z^*}$ change as the training proceeds. These states $\mathbf{s^*}$ are generated by decoding 
$\mathbf{z^*}$. 
Figure \ref{fig:fig_maze_2d} shows the case where a two-dimensional (2-D) latent space is used for VGAE along with the associated 2-D surrogate model. 
As MARL and REMAX are trained with more samples, as shown in the figures, REMAX tends to generate states near the landmark because it learns that these states are easily rewardable and helpful for MARL training.

\section*{Details about Environments}
We assume that the agents in the environments used in our study have the positions and velocities (health in SMAC, instead of velocities)  of all agents as a state whose dimension is $4N$, which consists of $2N$ for x-y positions and $2N$ for x-y velocities of $N$ agents. In other words, $F$ in the main paper is $4$. All the environments impose a penalty on an agent when the agent goes beyond the boundary of the environment. If $P_a$ and $P_l$ denote the positions of an agent and a landmark, the landmark is occupied when $\|P_a-P_l\|<0.1$ in the environment. 

\textbf{Maze.}
The maze has an agent ($N=1$) within $[0,1.5]^2$. $P_a$ and $P_l$ in the maze are $(1.25, 0.6)$ and $(0.25, 0.9)$, respectively. 

\textbf{Cooperative Navigation.}
The cooperative navigation has the agents ($N=2,3,4,6,8$) within $[0,1]^2$. The agents and landmarks in the cooperative navigation are at the center and at the four corners, respectively. If there are more than four landmarks, additional landmarks are at the middle of the sides, which is the middle of the landmarks located at the corners.

\textbf{Predator-Prey.}
The predator-prey has the agents ($N=4,8$) within $[0,1]^2$. The agents in the predator-prey are randomly positioned within $[0,1]^2$.

\textbf{SMAC.}
The SMAC in this study contained the agents (marines) ($N=3$). The three marines aim to fight against an equivalent number of marines of the game AI with very hard difficulties. SMAC originally has dense reward signals. However, we simply modify its rewards to be sparser at the final timestep of each episode: a victory reward of 1, a defeat reward of -1, and an attack reward of 1 to encourage attacking the enemies rather than running away.

\section*{Hyper-Parameters for Experiments}
We used 20,000 training episodes with 50 timesteps (total 1 million timesteps) for training the models in the cooperative navigation, 50,000 training episodes with 50 timesteps (total 2.5 million timesteps) in the predator-prey, and 50,000 training episodes with 60 timesteps (total 3 million timesteps) in SMAC. We trained the models on an NVIDIA GEFORCE RTX 2080 Ti GPU with five different random seeds. To summarize, all results in this study were averaged across five different seeds. 
All codes used in the experiments will be released.

\textbf{Hyper-Parameters of MADDPG.}
The hyper-parameters of MADDPG used in the experiments are summarized in Table \ref{table:table_hp_maddpg}. The output layer of the policy network of MADDPG for the environments provides the action as a five-sized tensor for hold, right, left, up, and down (in SMAC, nine-sized tensor for already-dead, hold, up, down, right, left, and attack-targets for each enemy). 

\begin{table}[h!]
  \caption{Hyper-parameters of MADDPG.}
  \label{table:table_hp_maddpg}
  \centering
    \begin{tabular}{cc}
        \toprule
        \multicolumn{1}{c}{MADDPG Hyper-Parameter} &  \\
        \midrule
        \multicolumn{1}{c}{\# Policy network MLP units} &(64, 64) \\
        \multicolumn{1}{c}{\# $Q$-network MLP units} &(64, 64) \\
        \multicolumn{1}{c}{Network parameter initialization} &Xavier uniform \\
        \multicolumn{1}{c}{Nonlinear activation} &ReLU \\
        \multicolumn{1}{c}{Policy network learning rate} &$10^{-2}$ \\
        \multicolumn{1}{c}{$Q$-network learning rate} &$10^{-2}$ \\
        \multicolumn{1}{c}{$\tau$ for updating target networks} &$10^{-2}$ \\
        \multicolumn{1}{c}{$\gamma$} &0.95 \\
        \multicolumn{1}{c}{Replay buffer size} &$10^{6}$ \\
        \multicolumn{1}{c}{Mini-batch size} &1024 \\
        \multicolumn{1}{c}{Optimizer} &Adam \\
        \bottomrule
    \end{tabular}
\end{table}

\textbf{Hyper-Parameters of REMAX.}
The hyper-parameters of REMAX used in the experiments are summarized in Table \ref{table:table_hp_remax}. The number of heads $K$ for multi-head attention in REMAX is $1$ for cooperative navigation and $2$ for predator-prey. 

\begin{table}[t]
  \caption{Hyper-parameters of REMAX.}
  \label{table:table_hp_remax}
  \centering
    \begin{tabular}{cc}
        \toprule
        \multicolumn{1}{c}{REMAX Hyper-Parameter} &    \\
        \midrule
        \multicolumn{1}{c}{$N_s$} &400\\
        \multicolumn{1}{c}{$F'$} &16 \\
        \multicolumn{1}{c}{$\sigma$} &ReLU \\
        \multicolumn{1}{c}{\# VGAE decoder network MLP units} &(32, 32) \\
        \multicolumn{1}{c}{\# surrogate model network MLP units} &(64, 64) \\
        \multicolumn{1}{c}{Network parameter initialization} &Xavier uniform \\
        \multicolumn{1}{c}{Nonlinear activation} &ReLU \\
        \multicolumn{1}{c}{Training epochs} &$3$ \\
        \multicolumn{1}{c}{Learning rate} &$10^{-4}$ \\
        \multicolumn{1}{c}{$\lambda$} &$10^{-3}$ \\
        \multicolumn{1}{c}{$\beta$} &1 \\
        \multicolumn{1}{c}{Mini-batch size} &1024 \\
        \multicolumn{1}{c}{Optimizer} &Adam \\
         \multicolumn{1}{c}{\# Loops $L$ for optimizing $f_\psi(z)$ w.r.t. $z$} &400 \\
        \multicolumn{1}{c}{$\delta$} &$10^{-1}$ \\
         \multicolumn{1}{c}{$\eta$} &$\mathcal{N}(0,\frac{1}{t})$ \\
        \bottomrule
    \end{tabular}
\end{table}

\textbf{Hyper-Parameters of GENE.}
The hyper-parameters of GENE used in the experiments are summarized in Table \ref{table:table_hp_gene}.

\begin{table}[t]
  \caption{Hyper-parameters of GENE.}
  \label{table:table_hp_gene}
  \centering
    \begin{tabular}{cc}
        \toprule
        \multicolumn{1}{c}{GENE Hyper-Parameter} &    \\
        \midrule
        \multicolumn{1}{c}{$N_s$} &400\\
        \multicolumn{1}{c}{VAE encoding space dimension} &1 \\
        \multicolumn{1}{c}{\# VAE encoder MLP units} &(32, 32) \\
        \multicolumn{1}{c}{\# VAE decoder MLP units} &(32, 32) \\
        \multicolumn{1}{c}{Network parameter initialization} &Xavier uniform \\
        \multicolumn{1}{c}{Nonlinear activation} &ReLU \\
        \multicolumn{1}{c}{Training epochs} &$3$ \\
        \multicolumn{1}{c}{Learning rate} &$10^{-4}$ \\
        \multicolumn{1}{c}{KDE bandwidth} &$0.05$ \\
        \multicolumn{1}{c}{KDE kernel} &Gaussian \\
        \multicolumn{1}{c}{Mini-batch size} &1024 \\
        \multicolumn{1}{c}{Optimizer} &Adam \\
        \bottomrule
    \end{tabular}
\end{table}

\textbf{Hyper-Parameters of RCG.}
For RCG used in the experiments, $R_\mathrm{min}=0.1$ and $R_\mathrm{max}=0.9$.

\textbf{Hyper-Parameters of EDTI.}
EDTI is based on the OpenAI implementation of PPO2 \cite{baselines,wang2019influence}, and its
default parameters are used here. The hyper-parameters, apart from the default parameters of EDTI used in the experiments, are summarized in Table \ref{table:table_hp_edti}.

\begin{table}[b!]
  \caption{Hyper-parameters of EDTI.}
  \label{table:table_hp_edti}
  \centering
    \begin{tabular}{cc}
        \toprule
        \multicolumn{1}{c}{EDTI Hyper-Parameter} &    \\
        \midrule
        \multicolumn{1}{c}{$\eta$} &1\\
        \multicolumn{1}{c}{$\beta_\text{T}$} &10 \\
        \multicolumn{1}{c}{$\beta_\text{int}$} &1 \\
        \multicolumn{1}{c}{$\beta_\text{ext}$} &0.1\\
        \multicolumn{1}{c}{$\beta_\text{r}$} &0.1 \\
        \multicolumn{1}{c}{$\beta_\text{int}^\text{plusV}$} &0.1 \\
        \multicolumn{1}{c}{$\beta_\text{ext}^\text{plusV}$} &0.01 \\
        \bottomrule
    \end{tabular}
\end{table}


\end{document}